%% file: main.tex
\documentclass{article} 
\usepackage{iclr2023_conference,times}
\iclrfinalcopy

\usepackage[utf8]{inputenc} 
\usepackage[T1]{fontenc}    
\usepackage{hyperref}       
\usepackage{url}            
\usepackage{booktabs}       
\usepackage{amsfonts}       
\usepackage{nicefrac}       
\usepackage{microtype}      
\usepackage{xcolor}         

\input{math_commands.tex}
\input{mycommand}

\title{Mitigating Dataset Bias by Using Per-sample Gradient}

\author{%
  Sumyeong Ahn\thanks{Equal contribution}\\
  KAIST AI\\
  \texttt{sumyeongahn@kaist.ac.kr}\\
  \And
  Seongyoon Kim$^{*}$\\
  KAIST ISysE\\
  \texttt{curisam@kaist.ac.kr}\\
  \And
  Se-young Yun\\
  KAIST AI\\
  \texttt{yunseyoung@kaist.ac.kr}\\
}

\begin{document}

\maketitle

\begin{abstract}
The performance of deep neural networks is strongly influenced by the training dataset setup. In particular, when attributes \revision{with} a strong correlation with the target attribute are present, the trained model can provide unintended prejudgments and show significant inference errors (\emph{i.e., the dataset bias problem}). Various methods have been proposed to mitigate dataset bias, and their emphasis is on weakly correlated samples, called \emph{bias-conflicting samples}. These methods are based on explicit bias labels provided by human\revision{s}. However, such methods require human costs. Recently, several studies have \revision{sought} to reduce human intervention by utilizing the output space values of neural networks, such as feature space, logits, loss, or accuracy. However, these output space values may be insufficient for the model to understand the bias attributes well. In this study, we propose a debiasing algorithm leveraging gradient called \textbf{P}er-sample \textbf{G}radient-based \textbf{D}ebiasing \revision{(PGD)}. PGD \revision{is comprised of} three steps:  (1) training a model on uniform batch sampling, (2) setting the importance of each sample in proportion to the norm of the sample gradient, and (3) training the model using importance-batch sampling, whose probability is obtained in step (2). Compared with existing baselines for various datasets, the proposed method showed state-of-the-art accuracy for the classification task. Furthermore, we describe theoretical understandings of how PGD can mitigate dataset bias. Code is available at \href{https://github.com/sumyeongahn/PGD}{Link}
\end{abstract}

\input{intro_iclr}
\input{problem_iclr}

\input{method_iclr}

\input{exp_iclr}

\input{analysis_iclr}

\input{related_iclr}

\input{conclusion_iclr}

\section*{Acknowledgement}
This work was supported by Institute of Information \& communications Technology Planning \& Evaluation (IITP) grant funded by the Korea government(MSIT) [No.2019-0-00075, Artificial Intelligence Graduate School Program(KAIST), 10\%] and Institute of Information \& communications Technology Planning \& Evaluation (IITP) grant funded by the Korea government(MSIT) [No.2022-0-00641, XVoice: Multi-Modal Voice Meta Learning, 90\%]

\bibliography{ref}
\bibliographystyle{iclr2023_conference}

\newpage
\appendix
\input{appendix}

\end{document}

%% file: math_commands.tex
\usepackage{amsmath,amsfonts,bm}

\def\eqref#1{equation~\ref{#1}}

\def\1{\bm{1}}

\DeclareMathAlphabet{\mathsfit}{\encodingdefault}{\sfdefault}{m}{sl}
\SetMathAlphabet{\mathsfit}{bold}{\encodingdefault}{\sfdefault}{bx}{n}

\DeclareMathOperator{\Tr}{Tr}

\DeclareMathOperator*{\argmax}{arg\,max}
\DeclareMathOperator*{\argmin}{arg\,min}

\newcommand{\trace}{\Tr}

%% file: mycommand.tex
\usepackage{microtype}
\usepackage{subfig}
\usepackage{booktabs} 
\usepackage{bbm}
\usepackage{mathtools}
\usepackage{amsthm}
\usepackage{nccmath}

\usepackage{algorithmic}
\usepackage{algorithm}

\usepackage[T1]{fontenc}
\usepackage{wrapfig,lipsum,booktabs}
\usepackage{natbib}
\usepackage{soul}
\usepackage{dsfont}
\usepackage{enumerate}
\usepackage{enumitem}

\usepackage{kotex}
\usepackage{hyperref}
\usepackage{amsmath}
\usepackage{amsthm}
\usepackage{amsfonts}
\usepackage{bbm}
\usepackage{dsfont}
\usepackage[Symbol]{upgreek}
\usepackage{lscape}
\usepackage{caption}
\usepackage{balance}

\usepackage{rotating}

\usepackage{multirow}

\usepackage{amssymb}
\usepackage{pifont}

\newcommand{\mc}[1]{\mathcal{#1}}
\newcommand{\bb}[1]{\mathbbm{#1}}

\theoremstyle{plain}
\newtheorem{thm}{Theorem}

\newtheorem{lem}{Lemma}

\theoremstyle{definition}
\newtheorem{defn}{Definition}

\newtheorem{assum}{Assumption}

\hypersetup{citecolor=blue}
\let\oldcite\citep
\renewcommand*\cite[1]{\oldcite{#1}}

\renewcommand*\eqref[1]{(\ref{#1})}

\newcommand{\note}[1]{\textcolor{black}{#1}}
\newcommand{\revision}[1]{\textcolor{black}{#1}}

\newcommand{\myparagraph}[1]{\vspace{0.0cm}\noindent\textbf{#1}~}

\def\code#1{\texttt{#1}}
\DeclarePairedDelimiter\norm{\lVert}{\rVert}

\theoremstyle{definition}

\usepackage{tabularx,booktabs,ragged2e}
\usepackage{diagbox}

\usepackage{array, boldline, rotating}
\newcommand{\thickhline}{\hlineB{4}}

%% file: intro_iclr.tex
\section{Introduction}\label{sec:intro}

\emph{Dataset
bias}~\citep{torralba2011unbiased,shrestha2021investigation} is a bad
training dataset problem that occurs when unintended easier-to-learn attributes~\emph{(i.e., bias attributes)}, having a high correlation with the target attribute, are present~\citep{shah2020pitfalls, ahmed2020systematic}. 
This is due to the fact that the model can infer outputs by focusing on the bias features, which could lead to testing failures.
For example, most ``camel'' images include a ``desert background,'' and this unintended
correlation can provide a false shortcut for answering ``camel'' on the basis of the ``desert.''  In~\citep{nam2020learning,lee2021learning}, samples of data that have a strong correlation (like \revision{the aforementioned desert/camel}) are called ``bias-aligned samples,'' while samples of data that have a weak correlation (like ``camel on the grass'' images) are termed ``bias-conflicting samples.''

To reduce the dataset bias, initial studies~\citep{kim2019learning, mcduff2019characterizing, singh2020don,
li2019repair} have frequently assumed a case where labels with bias attributes are provided, but these additional labels provided through human effort are expensive. Alternatively, the bias-type, such as ``background\revision{,}'' is assumed in~\citep{lee2019simple, geirhos2018imagenet, bahng2019learning,
cadene2019rubi, clark2019don}. However, assuming biased knowledge from humans is still unreasonable\revision{,} since even humans cannot predict the type of bias that may exist in a large dataset~\citep{schafer2016bias}. Data for deep learning is typically collected by web-crawling without thorough consideration of the dataset bias problem.

Recent studies~\citep{bras2020adversarial, nam2020learning, kim2021biaswap, lee2021learning, Seo_2022_CVPR, zhang2022correct} have replaced human intervention with DNN results. They have identified bias-conflicting samples by using empirical metrics for output space (\emph{e.g.,} training loss and accuracy). For example,~\citet{nam2020learning} suggested a ``relative difficulty'' based on per-sample training loss and thought that a sample with a high ``relative difficulty'' was \revision{a} bias-conflicting sample. Most of the previous research has focused on the output space, such as feature space (penultimate layer output)~\citep{lee2021learning, kim2021biaswap, bahng2019learning, Seo_2022_CVPR, zhang2022correct}, loss~\citep{nam2020learning}, and accuracy ~\citep{bras2020adversarial, liu2021just}. However, this limited output space can impose restrictions on describing the data in detail.


Recently, as an alternative, model parameter space (\emph{e.g.,} gradient~\citep{huang2021importance, killamsetty2021retrieve, mirzasoleiman2020coresets}) has been used to obtain high-performance gains compared to output space approaches for various target tasks. For example,~\citet{huang2021importance} used gradient-norm to detect out-of-distribution detection samples and showed that the gradient of FC layer $\in\bb{R}^{h\times c}$ could capture joint information between feature and softmax output, where $h$ and $c$ are the dimension\revision{s} of feature and output vector, respectively.  
Since the gradients of each data point $\in\bb{R}^{h\times c}$ constitute high\revision{-}dimensional information, it is much more informative than the output space, such as logit $\in\bb{R}^{c}$ and feature $\in\bb{R}^{h}$.
However, there is no approach to tackle the dataset bias problem using a gradient norm-based metric.


In this paper, we present a resampling method from the perspective of the per-sample gradient norm to mitigate dataset bias. Furthermore, we theoretically justify that the gradient-norm-based resampling method can be an excellent debiasing approach. Our key contributions can be summarized as follows:
\vspace{-15pt}
\begin{itemize}[leftmargin=*]\setlength\itemsep{0.1em}
    \item We propose \textbf{P}er-sample \textbf{G}radient-norm based \textbf{D}ebiasing \revision{(PGD)}, a simple and efficient gradient-norm-based debiasing method. PGD is motivated by prior research demonstrating~\cite{mirzasoleiman2020coresets, huang2021importance, killamsetty2021retrieve} that gradient is effective at finding rare samples, and it is also applicable to finding the bias-conflicting samples in the dataset bias problem (See Section~\ref{sec:method} and Appendix~\ref{app:add_exp}).
    \item PGD outperforms \revision{other dataset bias methods} on various benchmarks, such as colored MNIST (CMNIST), multi-bias MNIST (MBMNIST), corrupted CIFAR (CCIFAR), biased action recognition (BAR), biased FFHQ (BFFHQ), CelebA\revision{,} and CivilComments-WILD. In particular, for the colored MNIST case, the proposed method yielded higher unbiased test accuracies compared with the vanilla and the best method\revision{s} by $35.94\%$ and  $2.32\%$, respectively. (See Section~\ref{sec:exp})
    \item We provide theoretical evidence of the superiority of PGD. To this \revision{end}, we first explain that minimizing the trace of inverse Fisher information is a good objective to mitigate dataset bias. In particular, PGD, resampling based on the gradient norm computed by the biased model, is a possible optimizer for mitigating the dataset bias problem. (See Section~\ref{sec:analysis})
\end{itemize}

%% file: problem_iclr.tex
\vspace{-0.17in}
\section{Dataset Bias Problem}
\label{sec:prob}
\vspace{-0.1in}

\myparagraph{Classification model.}
We first describe the conventional supervised learning setting. Let us consider the classification problem when a training dataset $\mc{D}_{n} = \{(x_i,y_i) \}_{i=1}^n$\revision{,} with input image $x_i$ and corresponding label $y_i$\revision{,} is given. Assuming that there are $c \in \bb{N} \setminus\{1\}$ classes, $y_i$ is assigned to the one element in set $C=\{1,...,c \}$. 
Note that we focus on a situation where dataset $\mc{D}_{n}$ does not have noisy samples, for example, noisy labels or out-of-distribution samples (e.g., SVHN samples when the task is CIFAR-10).
When input $x_i$ is given, $f(y_i|x_i,\theta)$ represents the softmax output of the classifier for label $y_i$. \revision{This} is derived from the model parameter $\theta\in\bb{R}^d$.
The cross-entropy (CE) loss $\mc{L}_{\text{CE}}$ is frequently used to train the classifier, and it is defined as $\mc{L}_{\text{CE}} (x_i,y_i;\theta)= - \log f(y_i|x_i,\theta)$ \revision{when the label is one-hot encoded}.



\begin{wrapfigure}[9]{r}{0.33\textwidth}
    \vspace{-0.22in}
    \begin{minipage}{0.33\textwidth}
        \centering
        \includegraphics[width=\textwidth]{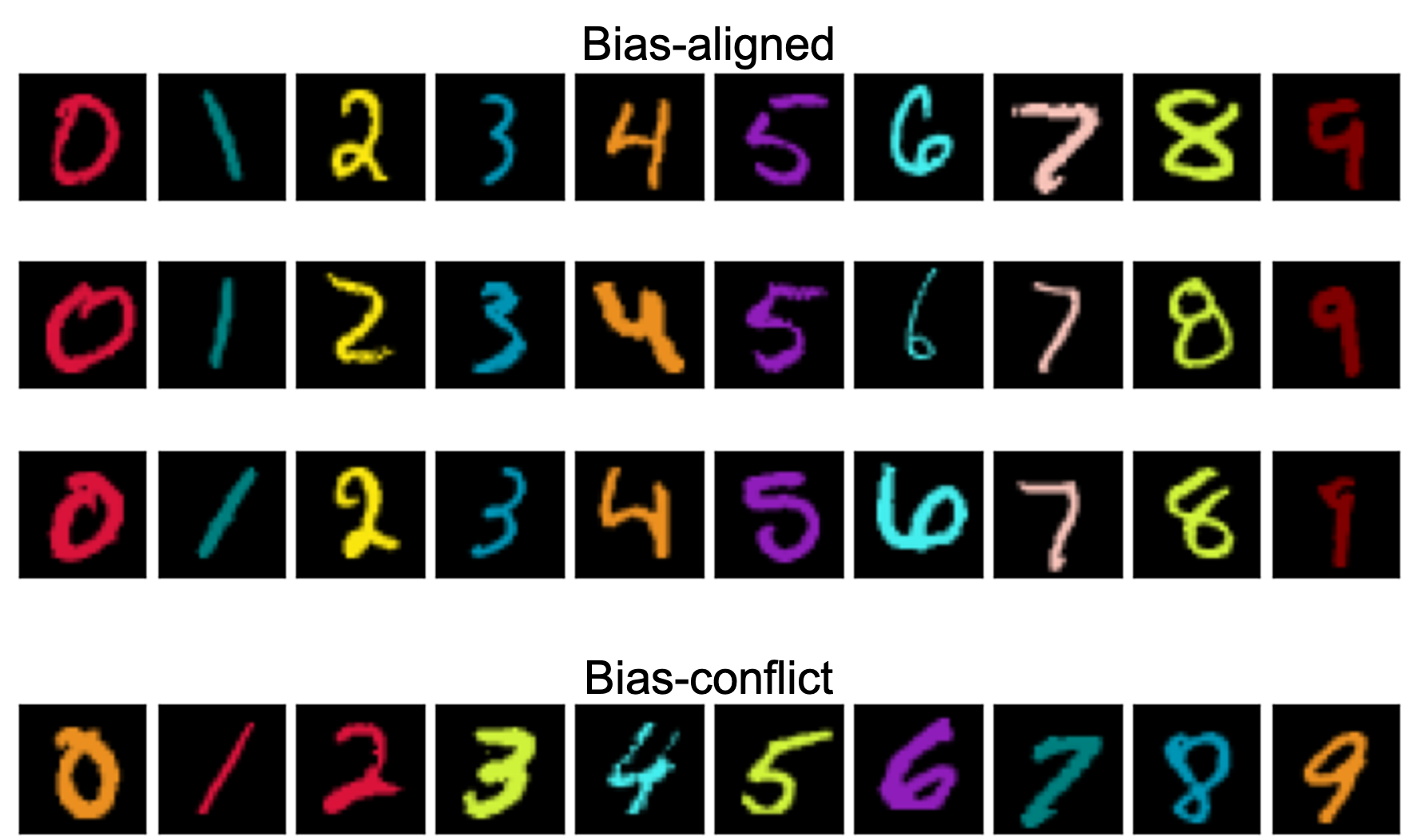}
        \vspace{-18pt}
        \caption{Target and bias attribute: \code{digit shape}, \code{color}.}
        \label{fig:bias_example}
    \end{minipage}
\end{wrapfigure}
\myparagraph{Dataset bias.}
Let us suppose that a training set\revision{,} $\mc{D}_{n}$\revision{,} is comprised of images, as shown in Figure~\ref{fig:bias_example}, and that the objective is to classify the \code{digits}. Each image can be described by a set of attributes, (\emph{e.g.,} for the first image in Figure~\ref{fig:bias_example}, it can be \{\code{digit 0}, \code{red}, \code{thin},...\}). The purpose of the training classifier is to find a model parameter $\theta$ that correctly predicts the target attributes, (\emph{e.g.,} \code{digit}). 
Notably, the target attributes are also interpreted as \emph{classes}. However, we focus on a case wherein another attribute that is strongly correlated to the target exists, and we call these attributes \emph{bias attributes}. For example, in Figure~\ref{fig:bias_example}, the bias attribute is \code{color}. 
Furthermore, samples whose bias attributes are highly correlated to the target attributes are called \emph{bias-aligned} (top three rows in Figure~\ref{fig:bias_example}).
Conversely, weakly correlated samples are called \emph{bias-conflicting} (see the bottom row of Figure~\ref{fig:bias_example}). 
Therefore, our main scope is that the training dataset which \revision{has} samples whose bias and target attributes are misaligned.
\footnote{Note that bias-alignment cannot always be strictly divisible in practice. For ease of explanation, we use the notations bias-conflicting/bias-aligned. }
According to~\citep{nam2020learning}, 
when the bias attributes are easier-to-learn than the target attributes,
dataset bias is problematic\revision{,} as the trained model may prioritize the bias attributes over the target attributes. For example, for a model trained on the images in Figure~\ref{fig:bias_example}, \revision{it} can output class $4$ when the (\code{Orange}, \code{0}) image (\emph{e.g.,} left bottom image) is given, due to the wrong priority, \code{color} which is an easier-to-learn attribute \citep{nam2020learning}.

%% file: method_iclr.tex
\vspace{-3pt}
\section{PGD: Per-sample Gradient-Norm-Based Debiasing}
\label{sec:method}
\vspace{-3pt}

\begin{wrapfigure}[14]{r}{0.43\textwidth}
\vspace{-0.32in}
\hspace{-0.1in}
    \scalebox{0.73}{
    \begin{minipage}[t]{0.62\textwidth}
    \begin{algorithm}[H]
    \caption{PGD: Per-sample Gradient-norm based Debiasing}
    \label{alg:pgd}
    \begin{algorithmic}[1]
    \STATE Input: dataset $\mc{D}_n$, learning rate $\eta$, iterations $T_b$, $T_d$, Batch size $B$, Data augmentation operation $\mc{A}(\cdot)$, Initial parameter $\theta_{0}$, GCE parameter $\alpha$\\
         \quad \textbf{\code{/** STEP 1: Train $f_b$ **/}} 
        \FOR{$t=1,2,\cdots,T_b$} 
            \STATE Construct a mini-batch $\mc{B}_{t}$ $=\{(x_{i},y_{i})\}_{i=1}^{B}\sim U$.
            \STATE Update $\theta_{t}$ as: \\
             \ $\theta_{t-1} - \frac{\eta}{B} \nabla_{\theta} \sum_{(x,y)\in B_{t}} \mc{L}_{\text{GCE}}(\mc{A}(x),y;\theta_{t-1},\alpha)$ 
        \ENDFOR\\
        \ \quad \textbf{\code{/** STEP 2: Calculate $h$ **/}}
            \STATE Calculate $h(x_i,y_i)$ for all $(x_i,y_i)\in\mc{D}_n$,~\eqref{eq:samprob}.\\ 
        \ \quad \textbf{\code{/** STEP 3: Train $f_d$ based on $h$ **/}}
        \FOR{$t=1,2,\cdots,T_d$}
            \STATE Construct a mini-batch 
            $\mc{B}'_{t}$ $=\{(x_{i},y_{i})\}_{i=1}^{B}\sim h$.            
            \STATE Update $\theta_{T_b+t}$ as:  \\
             \ $\theta_{T_b+t-1} -\frac{\eta}{B} \nabla_{\theta} \sum_{(x,y)\in \mc{B}'_{t}} \mc{L}_{\text{CE}}(\mc{A}(x),y;\theta_{T_b+t-1})$
        \ENDFOR
    \end{algorithmic}
    \end{algorithm}
    \end{minipage}}
\end{wrapfigure}

In this section, we propose a novel debiasing algorithm, coined as PGD. PGD consists of two models, biased $f_b$ and debiased $f_d$\revision{,} with parameters $\theta_b$ and $\theta_d$, respectively. Both models are trained sequentially. Obtaining the ultimately trained debiased model $f_d$ involves three steps: (1) train the biased model, (2) compute the sampling probability of each sample, and (3) train the debiased model. These steps are described in Algorithm~\ref{alg:pgd}.

\myparagraph{Step 1: Training the biased model.}
In the first step, the biased model is trained on the mini-batches sampled from a uniform distribution $U$, similar to conventional SGD-based training, with data augmentation $\mc{A}$. 
The role of the biased model is twofold: it detects which samples are bias-conflicting and calculates how much they should be highlighted. Therefore, we should make the biased model uncertain when it faces bias-conflicting samples. \revision{In doing so, the biased model, $f_b$, is trained on the generalized cross-entropy (GCE) loss $\mc{L}_{\text{GCE}}$ to amplify the bias of the biased model, motivated by~\citep{nam2020learning}.}
For an input image $x$\revision{,} the corresponding true class $y$, $\mc{L}_{\text{GCE}}$ is defined as 
$\mc{L}_{\text{GCE}} (x,y;\theta, \alpha) = \frac{1-f(y|x,\theta)^\alpha}{\alpha}$.
Note that $\alpha \in (0,1]$ is a hyperparameter that controls the degree of emphasizing the easy-to-learn samples, namely bias-aligned samples. Note that when $\alpha \to 0$, the GCE loss $\mc{L}_{\text{GCE}}$ is exactly the same as the conventional CE loss $\mc{L}_{\text{CE}}$. We set $\alpha=0.7$ as done by the authors of~\citet{zhang2018generalized}, \citet{nam2020learning} and \citet{lee2021learning}.

\myparagraph{Step 2: Compute the gradient-based sampling probability.}
In the second step, the sampling probability of each sample is computed from the trained biased model. Since rare samples have large gradient norm\revision{s} compared to the usual samples at the biased model~\citep{hsu2020generalized}, the sampling probability of each sample is computed to be proportional to its gradient norm so that bias-conflicting samples are over-sampled. Before computing the sampling probability, the per-sample gradient with respect to $\mc{L}_{\text{CE}}$ for all $(x_i,y_i) \in \mc{D}_n$ is obtained from the biased model.
We propose the following sampling probability of each sample $h(x_i,y_i)$ which is proportional to their gradient norm\revision{s,} as follows:
\vspace{-5pt}
\begin{equation}
    \label{eq:samprob}
    h(x_i,y_i) = \frac{\norm{\nabla_\theta \mc{L}_{\text{CE}}(x_i,y_i;\theta_b)}_s^r}{\sum_{(x_i,y_i) \in \mc{D}_{n}} \norm{\nabla_\theta \mc{L}_{\text{CE}}(x_i,y_i;\theta_{b})}_s^r},
\vspace{-5pt}
\end{equation}

where $\norm{\cdot}_s^r$ denotes $r$ square of the $L_s$ norm, and $\theta_b$ is the result of Step 1. Since, $h(x_i,y_i)$ is the sampling probability on $\mc{D}_{n}$, $h(x_i,y_i)$ is the normalized gradient-norm. 
Note that computing the gradient for all samples requires huge computing resources and memory. Therefore, we only extract the gradient of the final FC layer parameters. This is a frequently used technique for reducing the computational complexity~\citep{ash2019deep, mirzasoleiman2020coresets, killamsetty2021retrieve, killamsetty2021grad, killamsetty2020glister}.  In other words, instead of $h(x_i,y_i)$, we empirically utilize $\hat{h}(x_i,y_i) = \frac{\norm{\nabla_{\theta_{\text{fc}}} \mc{L}_{\text{CE}}(x_i,y_i;\theta_b)}_s^r}{\sum_{(x_i,y_i) \in \mc{D}_{n}} \norm{\nabla_{\theta_{\text{fc}}} \mc{L}_{\text{CE}}(x_i,y_i;\theta_{b})}_s^r}$, where $\theta_{\text{fc}}$ is the parameters of the final FC layer. We consider $r=1$ and $s=2$  (\emph{i.e.,} $L_2$), and deliver ablation studies on various $r$ and $s$ in Section~\ref{sec:exp}. 

\myparagraph{Step 3: Ultimate debiased model training.}
Finally, the debiased model\revision{,} $f_d$\revision{,} is trained using mini-batches sampled with the probability $h(x_i,y_i)$ obtained in Step 2. Note that, as described in Algorithm~\ref{alg:pgd}, our debiased model inherits the model parameters of the biased model $\theta_{T_b}$.
However, \citet{lee2021learning} argued that just oversampling bias-conflicting samples does not successfully debias, and this unsatisfactory result stems from the data diversity (\emph{i.e.,} data augmentation techniques are required). Hence, we used simple randomized augmentation operations $\mc{A}$ such as random rotation and random color jitter to oversample the bias-conflicting samples.

%% file: exp_iclr.tex
\vspace{-10pt}
\section{Experiments}
\label{sec:exp}
\vspace{-10pt}

In this section, we demonstrate the effectiveness of PGD for multiple benchmarks compared with previous proposed baselines. 
Detail analysis not in this section (\emph{e.g.,} training time, unbiased case study, easier to learn target attribute, sampling probability analysis, reweighting with PGD) are described in the Appendix~\ref{app:add_exp}.

\vspace{-5pt}
\subsection{Benchmarks}
\vspace{-5pt}

To precisely examine the debiasing performance of PGD, 
we used the Colored MNIST, Multi-bias MNIST, and Corrupted CIFAR datasets as synthetic datasets, which assume situations in which the model learns bias attributes first. BFFHQ, BAR, CelebA, and CivilComments-WILDS datasets obtained from the real-world \revision{are} used to observe the situations in which general algorithms have poor performance due to bias attributes. Note that BFFHQ and BAR are biased by using human prior knowledge, while CelebA and CivilComments-WILDS are \revision{naturally} biased datasets.
A detailed explanation of each benchmark are presented in Appendix~\ref{app:benchmark}.

\myparagraph{Colored MNIST (CMNIST).} CMNIST is a modified version of MNIST dataset~\citep{lecun2010mnist}, where \code{color} is the biased attribute and \code{digit} serv\revision{es} as the target. 
We randomly selected ten colors that will be injected into the digit. Evaluation was conducted for various ratios $\rho$ $\in$ $\{ 0.5\%,  1\%, 5\%\}$, where $\rho$ denotes the portion of bias-conflicting samples. Note that CMNIST has only one bias attribute: \code{color}.

\myparagraph{Multi-bias MNIST (MB-MNIST).} The authors of~\citep{shrestha2021investigation} stated that CMNIST is too simple to examine the applicability of debiaising algorithms for complex bias cases. 
However, the dataset that \citet{shrestha2021investigation} generated is also not complex\revision{,} since they did not use an \revision{real-world pattern} dataset (\emph{e.g.,} MNIST) and used simple \revision{artificial} patterns (\emph{e.g.,} straight line and triangle). Therefore, we generated a MB-MNIST; we used benchmark \revision{to reflect the real-worled} better than~\citep{shrestha2021investigation}. MB-MNIST consists of eight attributes: digit~\citep{lecun2010mnist}, alphabet~\citep{cohen2017emnist}, fashion object~\citep{xiao2017/online}, Japanese character~\citep{clanuwat2018deep}, digit color, alphabet color, fashion object color, Japanese character color. Among the eight attributes, the target attribute is digit shape and the others are the bias attributes. To construct MB-MNIST, we follow the CMNIST protocol, which generates bias by aligning two different attributes (\emph{i.e.,} digit and color) with probability $(1-\rho)$. MB-MNIST dataset is made by independently aligning the digit and seven other attributes with probabity $(1-\rho)$. Note that \revision{rarest} sample is generated with probability $\rho^7$. When $\rho$ is set as the CMNIST case, it is too low to generate sufficient misaligned samples. Therefore, we use $\rho \in \{10\%, 20\%, 30\% \}$ to ensure the trainability.

\myparagraph{Corrupted CIFAR (CCIFAR).} CIFAR10~\citep{krizhevsky2009learning} \revision{is comprised of} ten different objects, such as an \code{airplane} and a \code{car}. Corrupted CIFAR are biased with ten different types of texture bias (\emph{e.g.,} frost and brightness). The dataset was constructed by following the design protocol of~\citep{hendrycks2019benchmarking}, and the ratios $\rho$ $\in$ $\{ 0.5\%, 1\%, 5\%\}$ are used.


\myparagraph{Biased action recognition (BAR).}
Biased action recognition dataset was derived from~\citep{nam2020learning}. It comprised six classes for action, (\emph{e.g.,} \code{climbing} and \code{diving}), and each class is biased with place. For example, \code{diving} class pictures are usually taken \code{underwater}, while a few images are taken from the \code{diving pool}. 

\myparagraph{Biased FFHQ (BFFHQ).} BFFHQ dataset was constructed from the facial dataset, FFHQ~\citep{karras2019style}. It was first proposed in~\citep{kim2021biaswap} and was used in~\citep{lee2021learning}. It \revision{is comprised of} two gender classes, and each class is biased with age. For example, most female pictures are \emph{young} while male pictures are \emph{old}. This benchmark follows $\rho = 0.5\%$.


\myparagraph{CelebA.}
CelebA~\citep{DBLP:conf/iccv/LiuLWT15} is a common real-world face classification dataset. The goal is classifying the hair color (``blond" and ``not blond") of celebrities which has a spurious correlation with the gender (``male" or ``female") attribute. \revision{Hair color of almost all female images is blond}. We report the average accuracy and the worst-group accuracy on the test dataset.

\myparagraph{CivilComments-WILDS.} 
CivilComments-WILDS~\cite{DBLP:journals/corr/abs-1903-04561} is a dataset to classify whether an online comment is toxic or non-toxic. The mentions of certain demographic identities (male, female, White, Black, LGBTQ, Muslim, Christian, and other religion) cause the spurious correlation with the label.



\vspace{-5pt}
\subsection{Implementation.}
\vspace{-0.1in}

\myparagraph{Baselines.}
We select baselines available for the official code from the respective authors among debiasing methods without prior knowledge on the bias. Our baselines comprise six methods on the various tasks: vanilla network,
LfF~\citep{nam2020learning}, JTT~\citep{liu2021just}\footnote{In the case of JTT~\citep{liu2021just}, although the authors used bias label for validation dataset (especially, bias-conflicting samples), we tune the hyperparameters using a part of the biased training dataset for fair comparison. Considering that JTT does not show significant performance gain in the results, it is consistent with the existing results that the validation dataset is important in JTT, as described in ~\citep{idrissi2022simple}.}, Disen~\citep{lee2021learning}, 
EIIL~\citep{creager2021environment} and
CNC~\citep{zhang2022correct}.

\myparagraph{Implementation details.}
We use three types of networks: two types of simple convolutional networks (SimConv-1 and SimConv-2) and ResNet18~\citep{he2016deep}. Network imeplementation is described in Appendix~\ref{app:exp}. Colored MNIST is trained on SGD optimizer, batch size $128$, learning rate $0.02$, weight decay $0.001$, momentum $0.9$, learning rate decay $0.1$ every $40$ epochs, $100$ epochs training, and GCE parameter $\alpha$ $0.7$. For Multi-bias MNIST, it also utilizes SGD optimizer, and $32$ batch size, learning rate $0.01$, weight decay $0.0001$, momentum $0.9$, learning rate decay $0.1$ with decay step $40$. It runs $100$ epochs with GCE parameter $0.7$. For corrupted CIFAR and BFFHQ, it uses ResNet18 as a backbone network, and exactly the same setting presented by Disen~\citep{lee2021learning}. \footnote{\citet{lee2021learning} only reported bias-conflicting case for BFFHQ, but we report the unbiased test result.} \revision{For CelebA, we follows experimental setting of~\citep{zhang2022correct} which uses ResNet50 as a backbone network. For CivilComments-WILDS, we utilize exactly the same hyperparameters of~\cite{liu2021just} and utilize pretrained BERT.}   To reduce the computational cost in extracting the per-sample gradients, we  use only a fully connected layer, similar to~\citep{ash2019deep, mirzasoleiman2020coresets, killamsetty2021retrieve, killamsetty2021grad, killamsetty2020glister}. \revision{Except for CivilComments-WILDS and CelebA, we utilize data augmentation, such as color jitter, random resize crop and random rotation.} See Appendix~\ref{app:exp} for more details.

\begin{table}[t!]
\centering
    \caption{Average test accuracy and standard deviation (three runs) for
    experiments with the MNIST variants under various bias conflict ratios. The best
    accuracy is indicated in \textbf{bold} for each case. }
    \vspace{-5pt}
    \label{tab:control}
    \centering
    \begin{minipage}[b]{1.0\textwidth}
        \centering
        \resizebox{0.75\textwidth}{!}{
        \begin{tabular}{cccccc|c}
        \thickhline
Dataset                        & $\rho$ & Vanilla & LfF & JTT & Disen & PGD (Ours) \\ \hline
& 0.5\%             &  60.94{\tiny $\pm$ 0.97} & 91.35{\tiny $\pm$ 1.83} & 85.84{\tiny $\pm$ 1.32} & 94.56{\tiny $\pm$ 0.57} & \textbf{96.88{\tiny $\pm$ 0.28}}     \\ 
CMNIST & 1\%        &  79.13{\tiny $\pm$ 0.73} & 96.88{\tiny $\pm$ 0.20} & 95.07{\tiny $\pm$ 3.42} & 96.87{\tiny $\pm$ 0.64} & \textbf{98.35{\tiny $\pm$ 0.12}}    \\ 
& 5\%               &  95.12{\tiny $\pm$ 0.24} & 98.18{\tiny $\pm$ 0.05} & 96.56{\tiny $\pm$ 1.23} & 98.35{\tiny $\pm$ 0.20} & \textbf{98.62{\tiny $\pm$ 0.14}}   \\ \hline
& 10\%              &  25.23{\tiny $\pm$ 1.16} & 19.18{\tiny $\pm$ 4.45} & 25.34{\tiny $\pm$ 1.45} & 25.75{\tiny $\pm$ 5.38} & \textbf{61.38{\tiny $\pm$ 4.41}}      \\ 
MBMNIST & 20\%      &  62.06{\tiny $\pm$ 2.45} & 65.72{\tiny $\pm$ 6.23} & 68.02{\tiny $\pm$ 3.23} & 61.62{\tiny $\pm$ 2.60} &  \textbf{89.09{\tiny $\pm$ 0.97}}     \\ 
& 30\%              &  87.61{\tiny $\pm$ 1.60} & 89.89{\tiny $\pm$ 1.76} & 85.44{\tiny $\pm$ 3.44} & 88.36{\tiny $\pm$ 2.06} &  \textbf{90.76{\tiny $\pm$ 1.84}}     \\ \hline
& 0.5\%             &  23.06{\tiny $\pm$ 1.25} & 28.83{\tiny $\pm$ 1.30} & 25.34{\tiny $\pm$ 1.00} & 29.96{\tiny $\pm$ 0.71} &    \textbf{30.15{\tiny $\pm$ 1.22}}   \\ 
CCIFAR& 1\%         &  25.94{\tiny $\pm$ 0.54} & 33.33{\tiny $\pm$ 0.15} & 33.62{\tiny $\pm$ 1.05} & 36.35{\tiny $\pm$ 1.69} &   \textbf{42.02{\tiny $\pm$ 0.73}}    \\ 
& 5\%               &  39.31{\tiny $\pm$ 0.66} & 50.24{\tiny $\pm$ 1.41} & 45.13{\tiny $\pm$ 3.11} & 51.19{\tiny $\pm$ 1.38} &    \textbf{52.43{\tiny $\pm$ 0.14}}   \\ \thickhline

\end{tabular}}
\end{minipage}
\vspace{-5pt}
\end{table}

\begin{table}[]
        \centering
        \captionof{table}{Average test accuracy and standard deviation (three runs) for experiments with the raw image benchmarks: BAR and BFFHQ. The best accuracy is indicated in \textbf{bold}\revision{,} and for the overlapped best performance case is indicated in \underline{Underline}.}
        \vspace{-5pt}
        \label{tab:real_1}
        \resizebox{0.62\textwidth}{!}{
        \begin{tabular}{ccccccc}
            \thickhline
            Dataset                        &  Vanilla & LfF & JTT & Disen & PGD (Ours) \\ \hline
            BAR         &  63.15{\tiny $\pm$ 1.06}& 64.41{\tiny $\pm$ 1.30}& 63.62{\tiny $\pm$1.33} & \underline{64.70{\tiny $\pm$ 2.06}}& \textbf{65.39{\tiny $\pm$ 0.47}} \\ 
            BFFHQ       &  77.77{\tiny $\pm$ 0.45}& 82.13{\tiny $\pm$ 0.38}& 77.93{\tiny $\pm$ 2.16}& 82.77{\tiny $\pm$ 1.40}& \textbf{84.20{\tiny $\pm$ 1.15}} \\
            \thickhline
        \end{tabular}
        }\vspace{-10pt}
\end{table}

\vspace{-0.05in}
\subsection{Results and empirical analysis}

\myparagraph{Accuracy results.}
In Table~\ref{tab:control}, we present the comparisons of the image classification accuracy for the unbiased test sets. The proposed method outperforms the baseline methods for all benchmarks and for all different ratios.
For example, our model performance is $35.94\%$ better than that of the vanilla model for the colored MNIST benchmark with a ratio $\rho=0.5\%$. For the same settings, PGD performs better than Disen by $2.32\%$.

As pointed out in~\citep{shrestha2021investigation}, colored MNIST is too simple to evaluate debiasing performance on the basis of the performance of baselines. In Multi-bias MNIST case, other models 
fail to obtain
higher unbiased test results, even though the ratio is high, \emph{e.g.,} $10\%$. In this complex setting, PGD shows superior performance over other methods. For example, its performance is higher by $36.15\%$ and $35.63\%$ compared with the performance of vanilla model and Disen for the ratio of $10\%$.

\begin{table}[t!]
\centering
\caption{\revision{Average and worst test accuracy with the raw image benchmark: \textbf{CelebA} and raw NLP task: \textbf{CivilComments-WILDS}. The results of comparison algorithms  are the results reported in ~\cite{zhang2022correct}. The best worst accuracy is indicated in \textbf{bold}.}}
\vspace{-5pt}
\label{tab:real_2}
\revision{
\resizebox{0.62\textwidth}{!}{
\begin{tabular}{cc|cccccc}
\thickhline
& & Vanilla & LfF & EIIL & JTT & CNC & Ours \\ \hline
 \multicolumn{1}{c}{\multirow{2}{*}{{CelebA}}}             & \multicolumn{1}{|c|}{Avg.}  &         94.9 & 85.1 & 85.7 & 88.1 & 88.9 & 88.6 \\ 
                                     & \multicolumn{1}{|c|}{Worst} & 47.7 & 77.2 & 81.7 & 81.5 & \textbf{88.8} & \textbf{88.8} \\ \hline
\multicolumn{1}{c}{\multirow{2}{*}{CivilComments}} & \multicolumn{1}{|c|}{Avg.}  & 92.1 & 92.5 & 90.5 & 91.1 & 81.7 & 92.1 \\  
                                     & \multicolumn{1}{|c|}{Worst} & 58.6 & 58.8 & 67.0 & 69.3 & 68.9 & \textbf{70.6} \\ \thickhline
\end{tabular}}
}
\end{table}

Similar to the results for the bias-feature-injected benchmarks, as shown in Table~\ref{tab:real_1} and Table~\ref{tab:real_2}, PGD shows competitive performance among all the debiasing algorithms on the raw image benchmark (BAR, BFFHQ, and CelebA). For example, for the BFFHQ benchmark, the accuracy of PGD is $1.43\%$ better than that of Disen. As in Table~\ref{tab:real_2}, PGD outperforms the other baselines on CivilComments-WILDs, much more realistic NLP task. Therefore, we believe PGD also works well with transformer, and it is applicable to the real-world.

\begin{figure*}[t!]
    \vspace{-0.2in}
    \centering
    \hspace{1.0in}
    \includegraphics[width=0.3\textwidth]{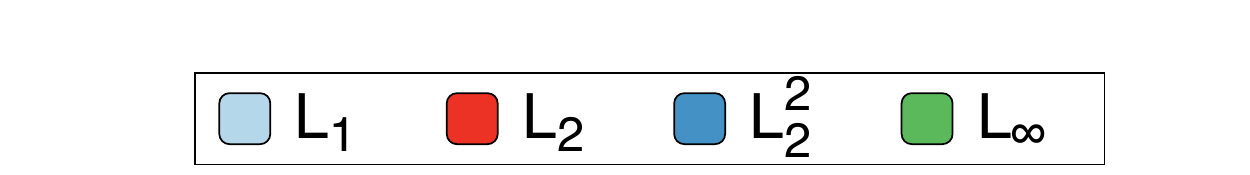}
    \vspace{-0.15in}
    \newline
    \centering
    \subfloat[Colored MNIST]{\includegraphics[width=0.3\textwidth]{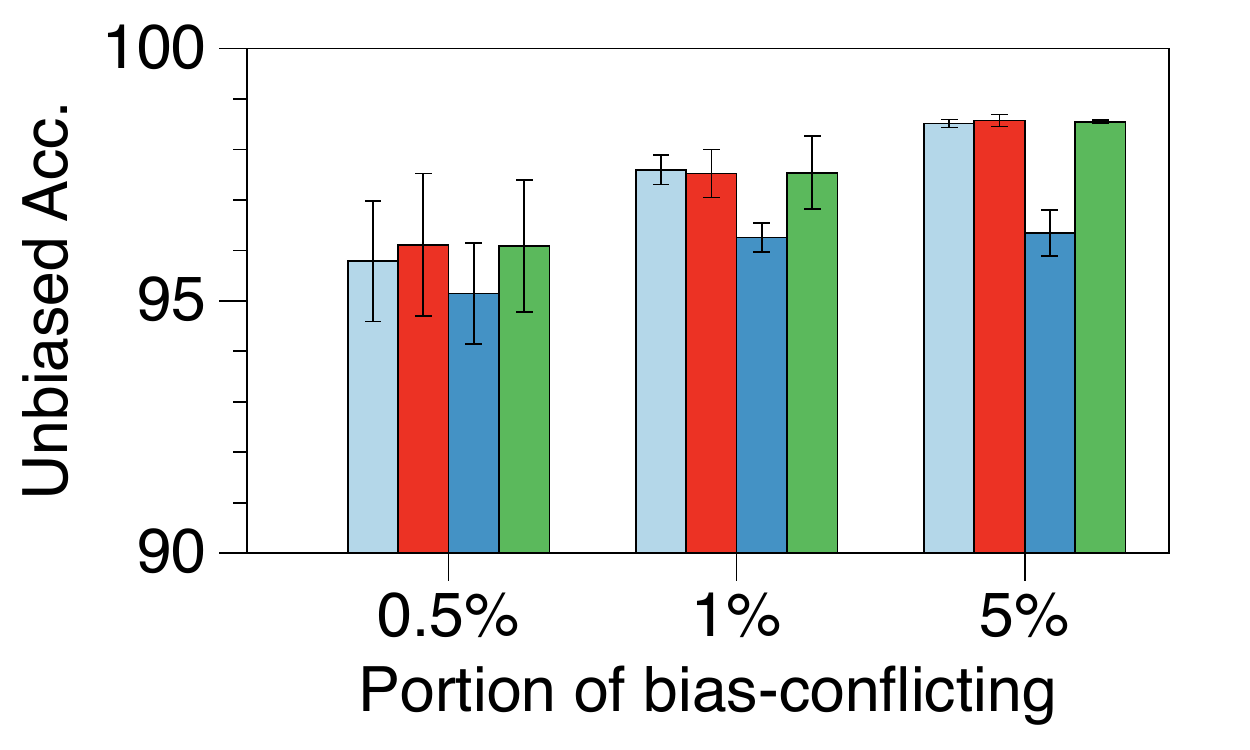} \label{fig:cm_norm}}
    \subfloat[Multi-bias MNIST]{\includegraphics[width=0.3\textwidth]{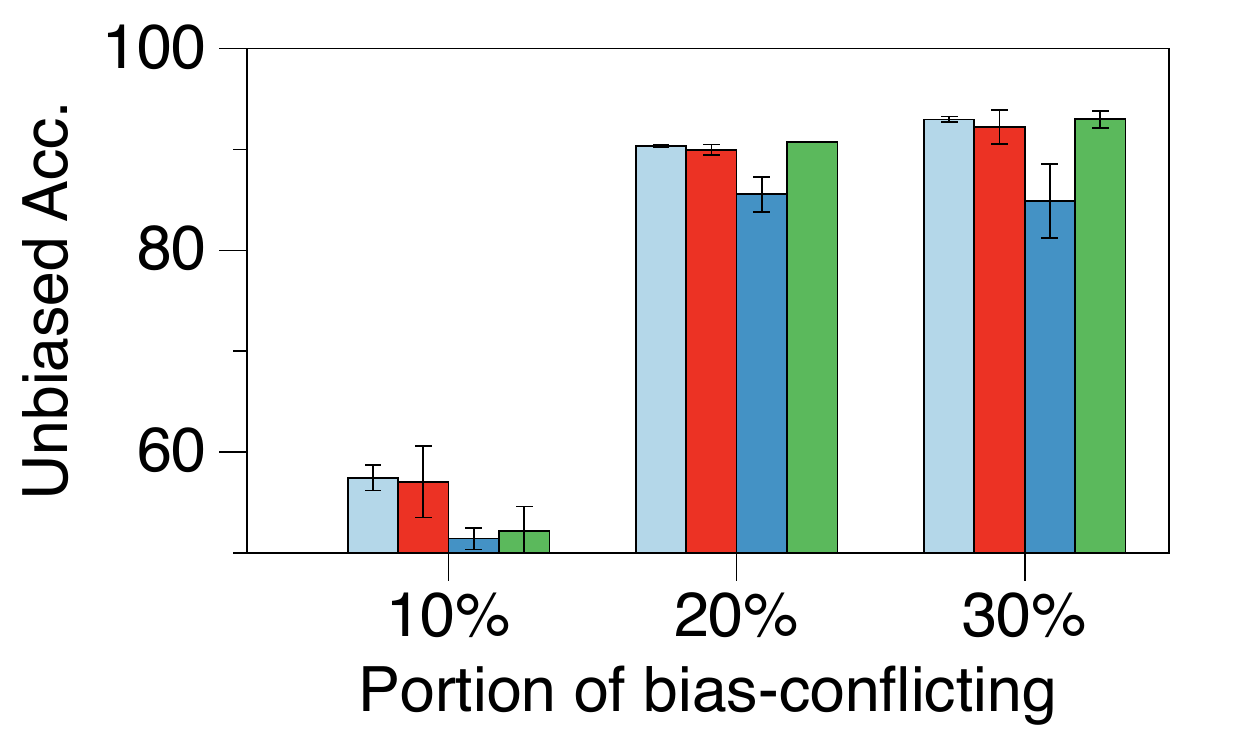} \label{fig:mm_norm}}
    \subfloat[Corrupted CIFAR]{\includegraphics[width=0.3\textwidth]{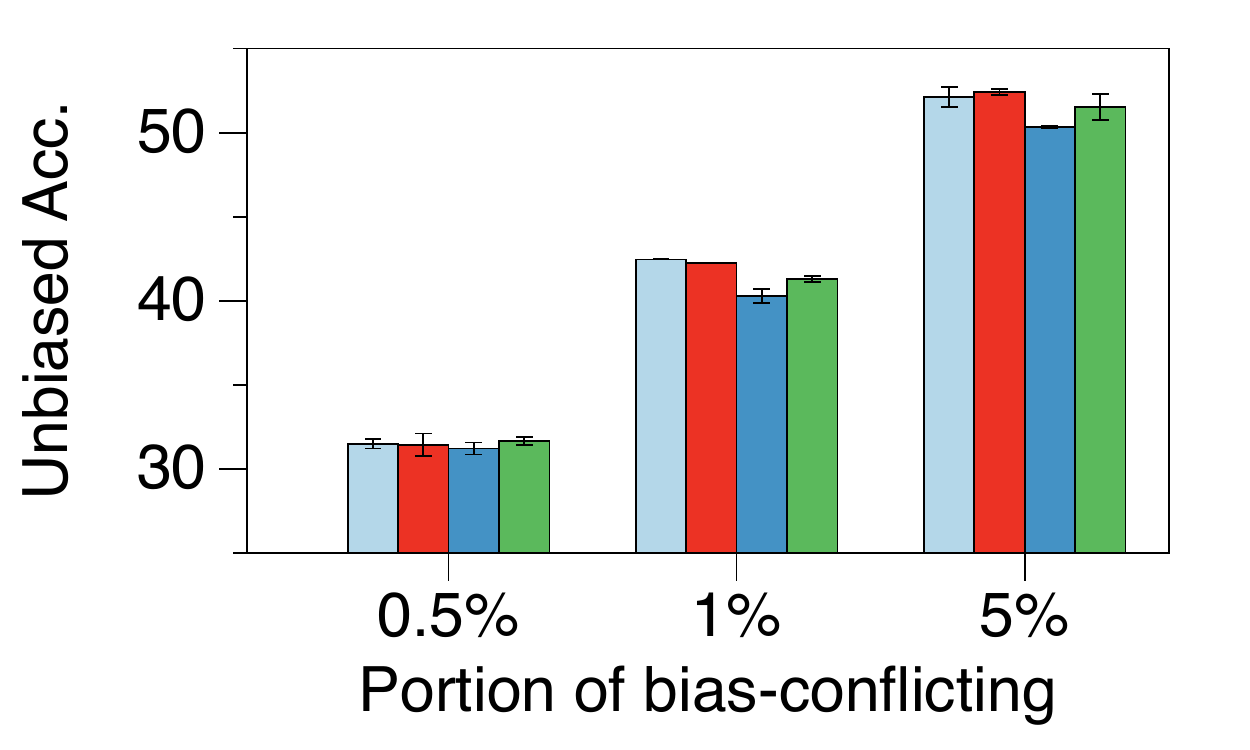} \label{fig:cc_norm}}
    \vspace{-0.13in}
    \caption{Average PGD results for various of norms, $\{L_1, L_2, L_2^2, L_\infty \}$, for the 
    feature-injected benchmarks. The error bars represent the standard deviation of three independent trials.}
    \vspace{-0.25in}
    \label{fig:norm}
\end{figure*}

\myparagraph{Unbiased test accuracy on various norms.} Since, gradient norm can have various candidates, such as order of the norm, we report four configurations of gradient norms. As shown in Figure~\ref{fig:norm}, all norms have significant unbiased test performance. Amongst them, the $L_2$-norm square case shows lower unbiased performance than the other cases. Therefore, it is recommended that any first power of $L_{\{1,2,\infty\}}$-norms be used in PGD for overcoming the dataset bias problem. This is quite different from the results \revision{in}~\citep{huang2021importance}, which suggested that $L_1$-norm is the best choice in the research field of out-of-distribution detection.

\begin{wraptable}[6]{r}{0.4\textwidth}
    \centering
        \vspace{-0.25in}
            \caption{Ablation studies on GCE and data augmentation (\checkmark for applied case).}
            \vspace{-0.07in}
            \label{tab:ablation}
            \resizebox{0.4\textwidth}{!}{
            \begin{tabular}{cc|cc}
                    \thickhline
                    GCE         & Aug.          & CMNIST (0.5\%)      & MB-MNIST (10\%)   \\ \hline
                                &               & 84.93 {\tiny $\pm$ 0.79}     & 40.58 {\tiny $\pm$ 3.39} \\ \hline
                    \checkmark  &               & 93.18 {\tiny $\pm$ 1.07}     & 45.70 {\tiny $\pm$ 2.91}    \\ \hline
                                &  \checkmark   & 91.19 {\tiny $\pm$ 0.97}     & 46.70 {\tiny $\pm$ 1.10}    \\ \hline
                    \checkmark  &  \checkmark   & \textbf{96.88 {\tiny $\pm$ 0.28}}     & \textbf{61.38 {\tiny $\pm$ 4.41}} \\ \thickhline
            \end{tabular}}
\end{wraptable}
\myparagraph{Ablation study.}
Table~\ref{tab:ablation} shows the importance of each module in our method: generalized cross entropy, and data augmentation modules. We set the ratio $\rho$ as $0.5\%$ and $10\%$ for CMNIST and MB-MNIST, respectively. 
We observe that GCE is more important that data augmentation for CMNIST. However, data augmentation shows better performance than GCE for MB-MNIST. 
\revision{In} all cases, the case where both are utilized outperforms the other cases.

%% file: analysis_iclr.tex
\vspace{-8pt}
\section{Mathematical Understanding of PGD}
\label{sec:analysis}
\vspace{-8pt}

This section provides a theoretical analysis of per-sample gradient-norm based debiasing. We first briefly summarize the maximum likelihood estimator (MLE) and Fisher information (FI),  which are ingredients of this section. We then interpret the debiasing problem as a min-max problem and deduce that solving \revision{it} the min-max problem can be phrased as minimizing the trace of the inverse FI. Since handling the trace of the inverse FI is very difficult owing to its inverse computation, we look at a glance by relaxing it into a one-dimensional toy example. In the end, we conclude that the gradient-norm based re-sampling method is an attempt to solve the dataset bias problem.


\vspace{-3pt}
\subsection{Preliminary}
\vspace{-5pt}


\myparagraph{Training and test joint distributions.}
The general joint distribution $\mc{P}(x,y|\theta)$ is assumed to be factored into the parameterized conditional distribution $f(y|x,\theta)$ and the marginal distribution $\mc{P}(x)$, which is independent of the model parameter $\theta$, (\emph{i.e.,} $\mc{P}(x,y|\theta)=\mc{P}(x)f(y|x,\theta)$).
We refer to the model  $f(y|x,\theta^\star)$ that produces the exact correct answer, as an oracle model, and to its parameter $\theta^\star$ as the oracle parameter. The training dataset $\mc{D}_{n}$ is sampled from $\{(x_i,y_i)\}_{i=1}^n \sim p(x)f(y|x,\theta^\star)$, where the training and test marginal distributions are denoted by $p(x)$ and $q(x)$, respectively. Here, we assume that both marginal distributions are defined on the marginal distribution space $\mc{M}= \{\mc{P}(x) | \int_{x \in \mc{X}} \mc{P}(x) \, dx = 1\}$, where $\mc{X}$ means the input data space, \emph{i.e.,} $p(x),\,q(x) \in\mc{M}$.



\myparagraph{The space $\mc{H}$ of sampling probability $h$.}
When the training dataset $\mc{D}_{n}$ is given, we denote the sampling probability as $h(x)$ which is defined on the probability space $\mc{H}$\footnote{Note that for simplicity, we abuse the notation $h(x,y)$  used in Section~\ref{sec:method} as $h(x)$. This is exactly the same for a given dataset $\mc{D}_{n}$ situation.}:
\vspace{-0.01in}
\begin{equation}
\label{eq:h_space}
\mc{H} = \{h(x) \,| \textstyle\sum_{(x_i,y_i) \in \mc{D}_{n}} h(x_i) = 1 \,,\, h(x_i) \ge 0 \quad \forall (x_i,y_i) \in \mc{D}_{n}\}. 
\vspace{-5pt}
\end{equation}

\myparagraph{Maximum likelihood estimator (MLE).}
When $h(x)$ is the sampling probability, we define MLE $\hat{\theta}_{h(x),\mc{D}_{n}}$ as follows:
\vspace{-10pt}
\begin{equation*}
{\hat{\theta}}_{h(x),\mc{D}_{n}}  
                                =\argmin{}_{\theta}   \,\, - \textstyle\sum_{(x_i,y_i) \in \mc{D}_{n}}\,h(x_i)\log f(y_i|x_i, \theta). 
\vspace{-7pt}
\end{equation*}
Note that MLE $\hat{\theta}_{h(x),\mc{D}_{n}}$ is a variable controlled by two factors: (1) a change in the training dataset $\mc{D}_{n}$ and (2) the adjustment of the sampling probability $h(x)$. If $h(x)$ is a uniform distribution $U(x)$, then $\hat{\theta}_{U(x),\mc{D}_{n}}$ is the outcome of empirical risk minimization (ERM).




\myparagraph{Fisher information (FI).}
FI, denoted by $I_{\mc{P}(x)}(\theta)$, is an information measure of samples from a given distribution $\mc{P}(x,y|\theta)$. It is defined as \revision{follows}:
\begin{equation}
\label{eq:decom_fi}
    I_{\mc{P}(x)}(\theta) = \bb{E}_{(x,y)\sim\mc{P}(x)f(y|x,\theta)}[\nabla_{\theta}\,\log\, f(y|x,\theta){{\nabla^\top_{\theta}}}\,\log\, f(y|x,\theta)].
\end{equation}
FI provides a guideline for understanding the test cross-entropy loss of MLE $\hat{\theta}_{U(x),\mc{D}_{n}}$. When the training set is sampled from $p(x)f(y|x,\theta^\star)$ and the test samples are generated from $q(x)f(y|x,\theta^\star)$, we can understand the test loss of MLE $\hat{\theta}_{U(x),\mc{D}_{n}}$ by using FI as follows.
\begin{thm}
\label{thm:thm1}
Suppose Assumption~\ref{app:main_assumption} 
 in Appendix~\ref{appsec:backgrounds}
and Assumption \ref{app:additional_assumption} in Appendix \ref{appsec:theorem_1} hold, then for sufficiently large $n=|\mc{D}_{n}|$, the following holds with high probability:

\begin{equation}
\resizebox{0.945\linewidth}{!}{
$\bb{E}_{(x,y)\sim q(x)f(y|x,\theta^\star)}\left[\bb{E}_{\mc{D}_{n}\sim p(x)f(y|x,\theta^\star)}\left[-\log{f(y|x,\hat{\theta}_{U(x),\mc{D}_n})}\right]\right] \leq\frac{1}{2n}\trace\left[I_{p(x)}(\hat{\theta}_{U(x),\mc{D}_n})^{-1}\right]\trace\left[I_{q(x)}(\theta^{\star})\right]$. \label{eq:fir}
}
\end{equation}
\end{thm}

Here is the proof sketch. The left-hand side of~\eqref{eq:fir} converges to the Fisher information ratio (FIR) $\trace \left[I_{p(x)}(\theta^\star)^{-1} I_{q(x)}(\theta^\star)\right]$-related term. Then, FIR can be decomposed into two trace terms with respect to the training and test marginal distributions $p(x)$ and $q(x)$. Finally, we show that the term $\trace[I_{p(x)}(\theta^\star)^{-1}]$ which is defined in the oracle model parameter can be replaced with $\trace [I_{p(x)}(\hat{\theta}_{U(x),\mc{D}_{n}})^{-1}]$. The proof of Theorem~\ref{thm:thm1} is in Appendix~D. 
Note that Theorem~\ref{thm:thm1} means that the upper bound of the test loss of MLE $\hat{\theta}_{U(x),\mc{D}_{n}}$ can be minimized by reducing $\trace[I_{p(x)}(\hat{\theta}_{U(x),\mc{D}_n})^{-1}]$.

\myparagraph{Empirical Fisher information (EFI).}
In practice, the exact FI~\eqref{eq:decom_fi} cannot be computed since we do not know the exact data generation distribution $\mc{P}(x)f(y|x,\theta)$. For practical reasons, the empirical Fisher information (EFI) is commonly used~\citep{jastrzkebski2017three, chaudhari2019entropy} to reduce the computational cost of gathering gradients for all possible classes when $x$ is given. In \revision{the present} study, we used a slightly more generalized EFI that involved the sampling probability $h(x)\in \mc{H}$ as follows: 
\begin{equation}
\label{eq:efi}
    \hat{I}_{h(x)}(\theta) = \textstyle\sum_{(x_i,y_i)\in\mc{D}_{n}} h(x_i)  \nabla_{\theta}\log\,f(y_i|x_i,\theta)\nabla^\top_{\theta}\log\,f(y_i|x_i,\theta).
\end{equation}
Note that the conventional EFI is the case when $h(x)$ is uniform. EFI provides a guideline for understanding the test cross-entropy loss of MLE $\hat{\theta}_{h(x),\mc{D}_{n}}$.


\subsection{Understanding dataset bias problem via min-max problem}

\myparagraph{Debiasing formulation from the perspective of min-max problem.} 
\revision{We formulate the dataset bias problem as described in Definition~\ref{dfn:debiasing_obj}. \eqref{eq:deb_obj} is a min-max problem formula, a type of robust optimization. Similar problem formulations for solving the dataset bias problem can be found in~\cite{arjovsky2019invariant, bao2021predict, zhang2022rich}. However, they assume that the training data is divided into several groups, and the model minimizes the worst inference error of the reweighted group dataset. In contrast, the objective of~\eqref{eq:deb_obj} minimizes the worst-case test loss without explicit data groups where the test distribution can be arbitrary.}




\begin{defn} When the training dataset $\mc{D}_n \sim p(x)f(y|x,\theta^\star)$ is given, the debiasing objective is 
    \label{dfn:debiasing_obj}
    \begin{equation}
        \label{eq:deb_obj}
        \min_{h(x) \in \mc{H}} \max_{q(x)\in\mc{M}} \bb{E}_{(x,y) \sim q(x)f(y|x,\theta^\star)}\left[-\log{f(y|x,{\hat{\theta}}_{h(x),\mc{D}_{n}})}\right].
    \end{equation}
\end{defn}
\vspace{-0.1in}
The meaning of Definition~\ref{dfn:debiasing_obj} is that we have to train the model $\hat{\theta}_{h(x),\mc{D}_{n}}$ so that the loss of the worst case test samples (\emph{$\max_{q(x)}$}) is minimized by controlling the sampling probability $h(x)$ (\emph{$\min_{h(x)}$}). Note that since we cannot control the given training dataset $\mc{D}_{n}$ and test marginal distribution $q(x)$, the only controllable term is the sampling probability $h(x)$.
Therefore, from Theorem~\ref{thm:thm1} and EFI, we design a practical objective function for the dataset bias problem as follows: 
\begin{equation}
    \label{eq:ultimate_obj}
    \min_{h(x) \in \mc{H}} \trace \left[ \hat{I}_{h(x)}(\hat{\theta}_{h(x),\mc{D}_n})^{-1}\right].
\end{equation}
\vspace{-0.3in}

\subsection{Meaning of PGD in terms of ~\eqref{eq:ultimate_obj}.}

In this section, we present an analysis of PGD with respect to~\eqref{eq:ultimate_obj}. To do so, we try to understand~\eqref{eq:ultimate_obj}, which is difficult to directly solve. It is because computing the trace of the inverse matrix is computationally expensive. Therefore, we intuitively understand~\eqref{eq:ultimate_obj} in the one-dimensional toy scenario.





\myparagraph{One-dimensional example.} We assume that $\mc{D}_{n}$ comprises sets $M$ and $m$ such that elements in each set share the same loss function.
\revision{For example, the loss function\revision{s} of the elements in set $M$ and $m$ are $\frac{1}{2} (\theta+a)^{2}$ and $\frac{1}{2}(\theta-a)^{2}$ with a given constant $a$, respectively.} 
We also assume that each sample of $M$ and $m$ has the set dependent probability mass $h_M(x)$ and $h_m(x)$, respectively. With these settings, our objective is to determine $h^\star (x)= \argmin_{h(x)\in\mc{H}} \trace [ \hat{I}_{h(x)} (\hat{\theta}_{h(x),\mc{D}_{n}})^{-1}]$. 
Thanks to the model's simplicity, we can easily find $h^\star(x)$ in a closed form with respect to the gradient at $\hat{\theta}_{U(x),\mc{D}_n}$ for each set, \emph{i.e.,} $g_M(\hat{\theta}_{U(x),\mc{D}_n})$ and $g_m(\hat{\theta}_{U(x),\mc{D}_n})$.


\begin{thm}\label{thm:norm}
    Under the above setting, the solution of $(h^\star_M(x),h^\star_m(x)) = \argmin_{h(x)\in\mc{H}} \trace [ \hat{I}_{h(x)}(\hat{\theta}_{h(x),\mc{D}_n})^{-1} ]$ is:
    \vspace{-0.07in}
    \begin{equation*}
    \label{eq:grad_prop}
        h^{\star}_{M}(x) ={|g_M({\hat{\theta}}_{U(x),\mc{D}_n})|}/ {Z}, \quad h^{\star}_{m}(x) ={|g_m({\hat{\theta}}_{U(x),\mc{D}_n})|}/ {Z},
    \end{equation*}
    where $Z = |M| |g_M(\hat{\theta}_{U(x),\mc{D}_n})| +  |m| |g_m(\hat{\theta}_{U(x),\mc{D}_n})|$, and $|M|$ and $|m|$ denote the cardinality of $M$ and $m$, respectively.
\end{thm}

\vspace{-0.05in}

The proof of Theorem~\ref{thm:norm} is provided in Appendix~E. 
Note that $h_M^\star(x)$ and $h_m^\star(x)$ are computed using the trained biased model with batches sampled from the uniform distribution $U(x)$. It is the same with the second step of PGD.

\begin{wrapfigure}[9]{r}{0.45\textwidth}
\vspace{-0.37in}
\begin{minipage}{0.45\textwidth}
    \centering
    \hspace{0.4in}
    \includegraphics[width=0.5\textwidth]{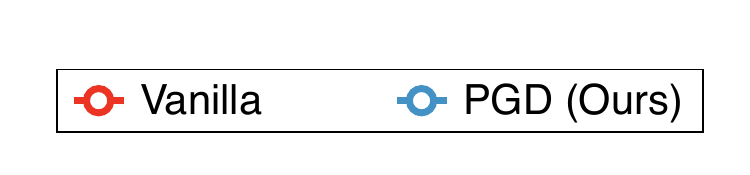}
    \vspace{-0.2in}
    \newline
    \subfloat[Colored MNIST]{\includegraphics[width=0.5\textwidth]{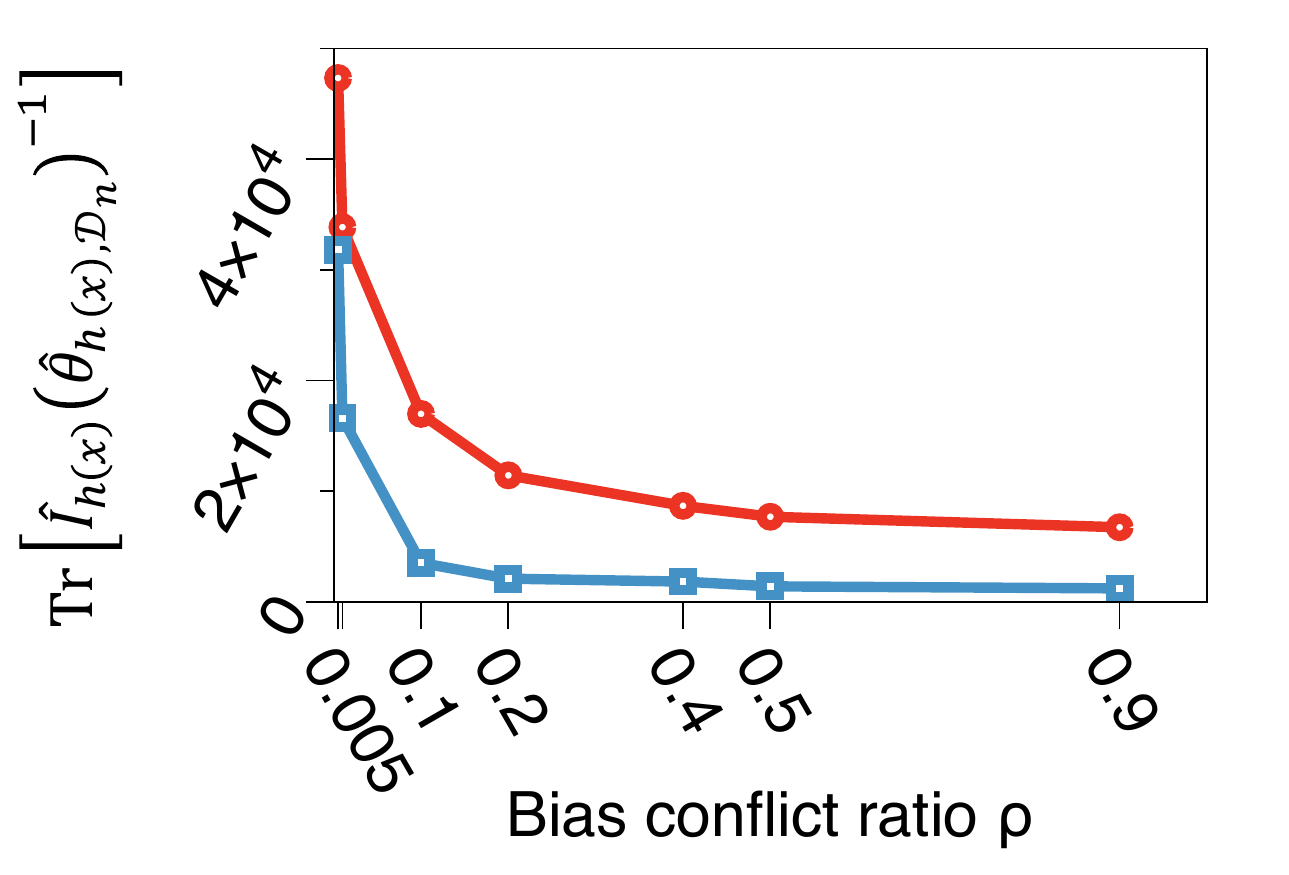} \label{fig:cm_invq_compare}}
    \subfloat[Multi-bias MNIST]{\includegraphics[width=0.5\textwidth]{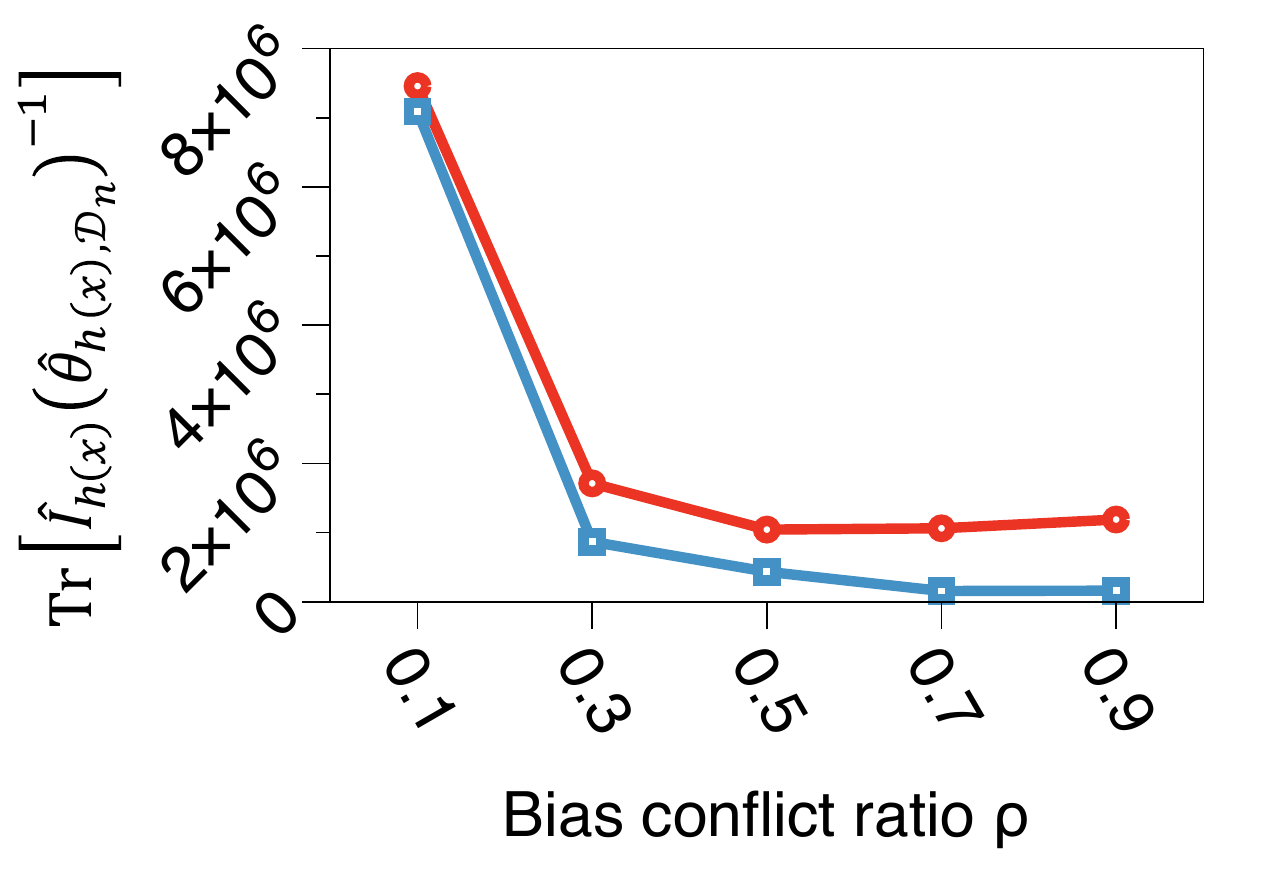} \label{fig:mm_innvq_compare}}
    \vspace{-0.1in}
    \caption{Target objective $\trace [ \hat{I}_{h} (\hat{\theta}_{h(x),\mc{D}_{n}})^{-1} ]$. PGD : $h(x)=\hat{h}(x)$, and vanilla: $h(x)=U(x)$.}
        \label{fig:invq_compare}
    \end{minipage}
\end{wrapfigure}

\myparagraph{PGD tries to minimize~\eqref{eq:ultimate_obj}.}
Theorem~\ref{thm:norm} implies that~\eqref{eq:ultimate_obj} can be minimized by sampling in proportion to their gradient norm. Because the \revision{basis} of PGD is oversampling based on the gradient norm from the biased model, we can deduce that PGD strives to satisfy~\eqref{eq:ultimate_obj}.
Furthermore, we empirically show that PGD reduces the trace of the inverse of EFI in the high-dimensional case, as evident in Figure~\ref{fig:invq_compare}.

%% file: related_iclr.tex
\section{Related Work}
\label{sec:related}

\myparagraph{Debiasing with bias label.} In~\cite{goyal2017making,
goyal2020rel3d}, a debiased dataset was generated using human labor. 
Various studies~\cite{alvi2018turning, kim2019learning, mcduff2019characterizing,
singh2020don, teney2021evading} have attempted to reduce dataset bias using \emph{explicit bias labels}. Some of these studies~\cite{alvi2018turning, kim2019learning,
mcduff2019characterizing, singh2020don, li2018resound, li2019repair}, used
bias labels for each sample to reduce the influence of the bias labels when
classifying target labels. Furthermore, \citet{tartaglione2021end} proposed the EnD
regularizer, which entangles target correlated features and disentangles biased
attributes. Several studies~\citep{alvi2018turning, kim2019learning,
teney2021evading} have designed DNNs as a shared feature extractors and multiple
classifiers.  In contrast to the shared feature extractor
methods,~\citet{mcduff2019characterizing} and \citet{ramaswamy2021fair} fabricated a
classifier and conditional generative adversarial networks, yielding test
samples to determine whether the classifier was biased. Furthermore, \citet{singh2020don} proposed
a new overlap loss defined by a class activation map (CAM). The overlap loss
reduces the overlapping parts of the CAM outputs of the two bias labels and target
labels. The authors of~\citep{li2019repair, li2018resound} employed bias labels to detect bias-conflicting samples and to oversample them to debias. In~\citep{liu2021just}, a reconstructing method based on the sample accuracy was proposed. \citet{liu2021just} used bias labels in the validation dataset to tune the hyper-parameters. On the other hand, there has been a focus on fairness within each attribute~\citep{hardt2016equality, Woodworth2017LearningNP, pleiss2017fairness, agarwal2018reductions}. Their goal is to prevent bias attributes from affecting the final decision of the trained model.  

\myparagraph{Debiasing with bias context.}
In contrast to studies assuming the explicit bias labels, a few studies~\citep{geirhos2018imagenet,
wang2019learning, lee2019simple, bahng2019learning, cadene2019rubi,
clark2019don} assumed that the bias context is known.
In~\citep{geirhos2018imagenet, wang2019learning, lee2019simple}, debiasing was
performed by directly modifying known context bias. In particular, the authors
of~\citep{geirhos2018imagenet} empirically showed that CNNs trained on
ImageNet~\citep{imagenet_cvpr09} were biased towards the image texture, and
they generated stylized ImageNet to mitigate the texture bias, while
\citet{lee2019simple} and \citet{wang2019learning} inserted a filter in front of the models so
that the influence of the backgrounds and colors of the images could be removed.
Meanwhile, some studies~\citep{bahng2019learning,clark2019don,cadene2019rubi}, mitigated bias
by reweighting bias-conflicting samples:~\citet{bahng2019learning}
used specific types of CNNs, such as BagNet~\citep{brendel2018approximating}, to
capture the texture bias, and the bias was reduced using the Hilbert-Schmidt
independence criterion (HSIC). In the visual question answering (VQA)
task,~\citet{clark2019don} and \citet{cadene2019rubi} conducted debiasing using the entropy regularizer or
sigmoid output of the biased model trained on the fact that the biased
model was biased toward the question.

\myparagraph{Debiasing without human supervision.} Owing to the impractical
assumption that bias information is given, recent studies have aimed to mitigate bias
without human supervision~\citep{bras2020adversarial, nam2020learning,
darlow2020latent, kim2021biaswap, lee2021learning}.
\citet{bras2020adversarial} identified bias-conflicting samples by sorting the average accuracy of multiple train-test iterations and performed debiasing by training on the samples with low average accuracy. In~\citep{ahmed2020systematic}, each class is divided into two clusters based on IRMv1 penalty~\citep{arjovsky2019invariant} using the trained biased model, and the deibased model is trained so that the output of two clusters become similar. Furthermore, ~\citet{kim2021biaswap} used Swap Auto-Encoder~\citep{park2020swapping} to generate bias-conflicting samples, and~\citet{darlow2020latent} proposed the modification of the latent representation to generate bias-conflicting samples by using an auto-encoder. \citet{lee2021learning} and \citet{nam2020learning} proposed a debiasing algorithm weighted training by using a relative difficulty score, which is measured by the per-sample training loss. Specifically,~\citet{lee2021learning} used feature\revision{-}mixing techniques to enrich the dataset feature information. \citet{Seo_2022_CVPR} and \citet{sohoni2020no} proposed unsupervised clustering based debiasing method. Recently, contrastive learning based method~\citep{zhang2022correct} and self-supervised learning method~\citep{kim2022learning} are proposed. On the other hand, there have been studies~\citep{li2021discover, lang2021explaining, krishnakumar2021udis} that identify the bias attribute of the training dataset without human supervision.

%% file: conclusion_iclr.tex
\section{Conclusion}
\label{sec:conclusion}
We propose a gradient-norm-based dataset oversampling method for mitigating the dataset bias problem. The main intuition of this work is that gradients contain abundant information about each sample. Since the bias-conflicting samples are relatively more difficult-to-learn than bias-aligned samples, the bias-conflicting samples have a higher gradient norm compared with the others. Through various experiments and ablation studies, we demonstrate the effectiveness of our gradient-norm-based oversampling method, called PGD. Furthermore, we formulate the dataset bias problem as a min-max problem, and show theoretically that it can be relaxed \revision{by} minimizing the trace of the inverse Fisher information. We provide empirical and theoretical evidence that PGD tries to solve the \revision{problem of} minimizing the trace of the inverse Fisher information problem. Despite this successful outcome and analysis, we are still working on two future projects: release approximations\revision{,} such as a toy example\revision{,} for understanding PGD and cases where the given training dataset is corrupted, such as with noisy labels. We hope that this study will help improve understanding of researchers about the dataset bias  problem.

%% file: appendix.tex
\appendix
\pagebreak

\begin{center}
    \noindent\\
    \vspace{0.2in}
    \textbf{\LARGE -- Appendix --}\\
    \vspace{0.1in}
    \textbf{\LARGE Mitigating Dataset Bias by Using Per-sample Gradient}\\
    \vspace{0.1in}
    \noindent
\end{center}

\setcounter{assum}{0}
\setcounter{thm}{0}
\setcounter{lem}{0}
\setcounter{equation}{0}

Due to the page constraint, this extra material includes additional results and theoretical proofs that are not in the original manuscript. Section~\ref{app:benchmark} demonstrates how to create datasets. Section~\ref{appsec:exp_setting} contains implementation details such as hyperparameters, computer resources, and a brief explanation of the baseline. Section \ref{app:comp_anal} and Section \ref{app:second_eas_exp} include case studies and empirical evidence of PGD. Section~\ref{app:add_exp} demonstrates additional experiment results. In Section~\ref{appsec:backgrounds}, we first provide a notation summary and some assumptions for theoretical analysis. Section~\ref{appsec:theorem_1} and Section~\ref{appsec:theorem_2} include proofs of Theorem~\ref{thm:thm1} and Theorem~\ref{thm:norm} with lemmas, respectively.

\input{./appendix/data}

\input{./appendix/exp}

\input{./appendix/theory}

\input{./appendix/thm1}
\input{./appendix/thm2}

%% file: appendix/data.tex
\section{Benchmarks and baselines}
\label{app:benchmark}

We explain the datasets utilized in Section~\ref{sec:exp}. In short, we build MNIST variants from scratch, while others get them directly from online repositories BAR\footnote{https://github.com/alinlab/BAR}, CCIFAR and BFFHQ\footnote{https://github.com/kakaoenterprise/Learning-Debiased-Disentangled}.

\subsection{Controlled Biased Benchmarks}
\label{app:data_con}

\begin{figure}[h!]
    \centering
    \begin{minipage}[c]{0.32\textwidth}
    \includegraphics[width=1\textwidth]{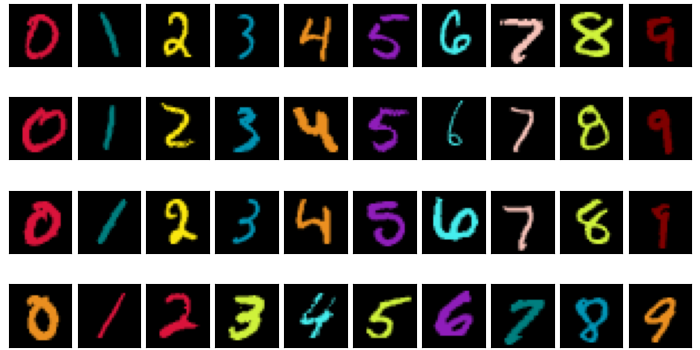} \label{fig_app:cm}
    \caption{Colored MNIST: The single bias attribute is color, and the target attribute is shape. The top 3 rows represent bias-aligned samples, and the bottom row samples are bias-conflicting examples.}
    \end{minipage}
    \begin{minipage}[c]{0.32\textwidth}
    \includegraphics[width=1\textwidth]{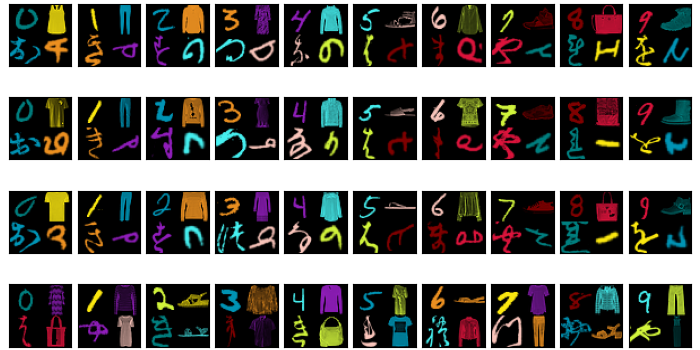} \label{fig_app:bm}
    \caption{Multi-bias MNIST: Multiple colors and objects bias, with digit shape as the target. The top 3 rows represent bias-aligned samples, and the bottom row samples are bias-conflicting examples.}
    \end{minipage}
    \begin{minipage}[c]{0.32\textwidth}
    \includegraphics[width=1\textwidth]{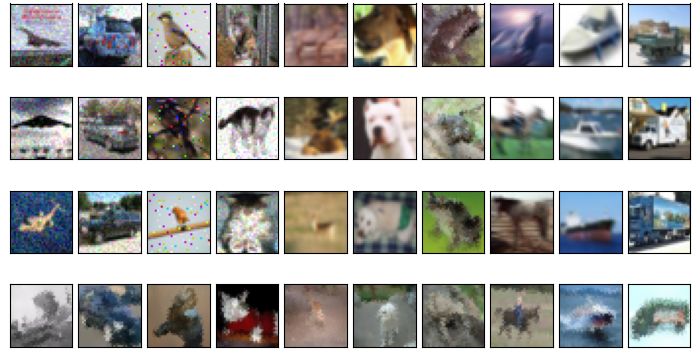} \label{fig_app:cc}
    \caption{Corrupted CIFAR: corruption is the bias attribute, while target attribute is object. The top three rows are bias-aligned samples, while the bottom row are bias-conflicting examples.}
    \end{minipage}
\end{figure}

\myparagraph{Colored MNIST (CMNIST)}
The MNIST dataset~\cite{lecun2010mnist} is composed of 1-dimensional grayscale handwritten images. The size of the image is $28 \times 28$. We inject color into these gray images to give them two main attributes: color and digit shape. This benchmark comes from related works~\cite{nam2020learning, kim2021biaswap, lee2021learning, bahng2019learning}. At the beginning of the generation, ten uniformly sampled RGB colors are chosen, $\{C_i\}_{i\in [10]} \in \bb{R}^{3 \times 10}$. When the constant $\rho$, a ratio of bias-conflicting samples, is given, each sample $(x,y)$ is colored by the following steps: (1) Choose bias-conflicting or bias-aligned samples: take a random sample and set it to bias-conflicting set when $u < \rho$ where $u \sim \mc{U}(0,1)$, otherwise bias-aligned. In experiments, we use $\rho \in \{0.5\%, 1\%, 5\% \}$. (2) Coloring: Note that each $C_i \in \bb{R}^{3}$ ($i\in [10]$) is a bias-aligned three-dimensional color vector for each digit $i\in\{10\}$. 
Then for bias-aligned images with the arbitrary digit $y$, color the digit with $c \sim \mc{N}(C_y, \sigma \mathit{I}_{3\times 3})$. In the case of bias conflicting images with the arbitrary digit $y$, first uniformly sample $C_{U_y}\in\{C_i\}_{i\in [10]\setminus y}$, and color the digit with $c \sim \mc{N}(C_{U_y}, \sigma \mathit{I}_{3\times 3})$. In the experiments, we set $\sigma$ as 0.0001. 
We use $55,000$ samples for training $5,000$ samples for validation (\emph{i.e.,} $10\%$), and $10,000$ samples for testing. Take note that test samples are unbiased, which means $\rho = 90\%$.

\myparagraph{Multi-Bias MNIST}
Multi-bias MNIST has images with size $56\times 56$. This dataset aims to test the case where there are multiple bias attributes. To accomplish this, we inject a total of seven bias attributes: digit color, fashion object, fashion color, Japanese character, Japanese character color, English character, and English character color, with digit shape serving as the target attribute. We inject each bias independently into each sample, as with the CMNIST case (\emph{i.e.,} sampling and injecting bias). We also set $\rho = 90\%$ for all bias attributes to generate an unbiased test set. As with CMNIST, we use $55,000$ samples for training and $5,000$ samples for validation, and $10,000$ samples for testing.

\myparagraph{Corrupted CIFAR}
This dataset was generated by injecting filters into the CIFAR10 dataset~\cite{krizhevsky2009learning}. The work~\cite{nam2020learning, lee2021learning} inspired this benchmark. In this benchmark, the target attribute and the bias attribute are object and corruption, respectively. \{Snow, Frost, Fog, Brightness, Contrast, Spatter, Elastic, JPEG, Pixelate, Saturate\} are examples of corruption. We downloaded this benchmark from the repository of the official code of Disen~\cite{lee2021learning}. This dataset contains $45,000$ in training samples, $5,000$ in validation samples, and $10,000$ in testing images. As with prior datasets, the test dataset is composed of unbiased samples (\emph{i.e.,} $\rho = 90\%$).

\subsection{Real-world benchmarks}
\label{app:data_real}

\begin{figure}[h!]
    \centering
    \begin{minipage}[c]{0.49\textwidth}
    \includegraphics[width=1\textwidth]{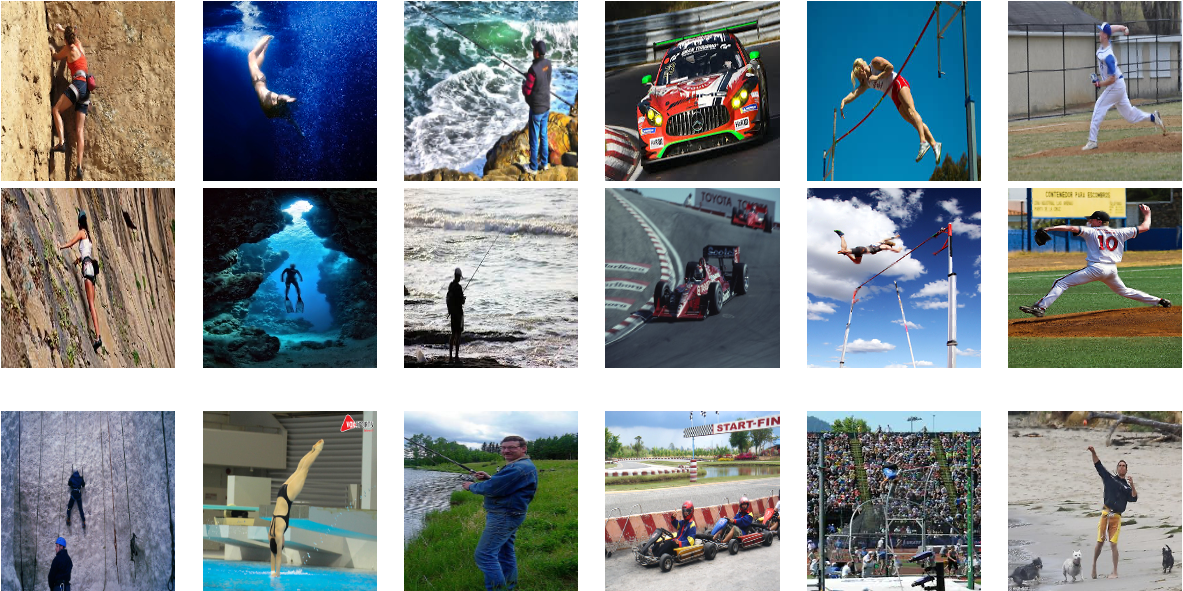} \label{fig_app:bar}
    \caption{Biased Action Recognition: The biased attribute is background, while the target attribute is action. The top 2 rows are bias-aligned samples, and the bottom row is bias-conflict samples.}
    \end{minipage}
    \hfill
    \begin{minipage}[c]{0.49\textwidth}
    \includegraphics[width=1\textwidth]{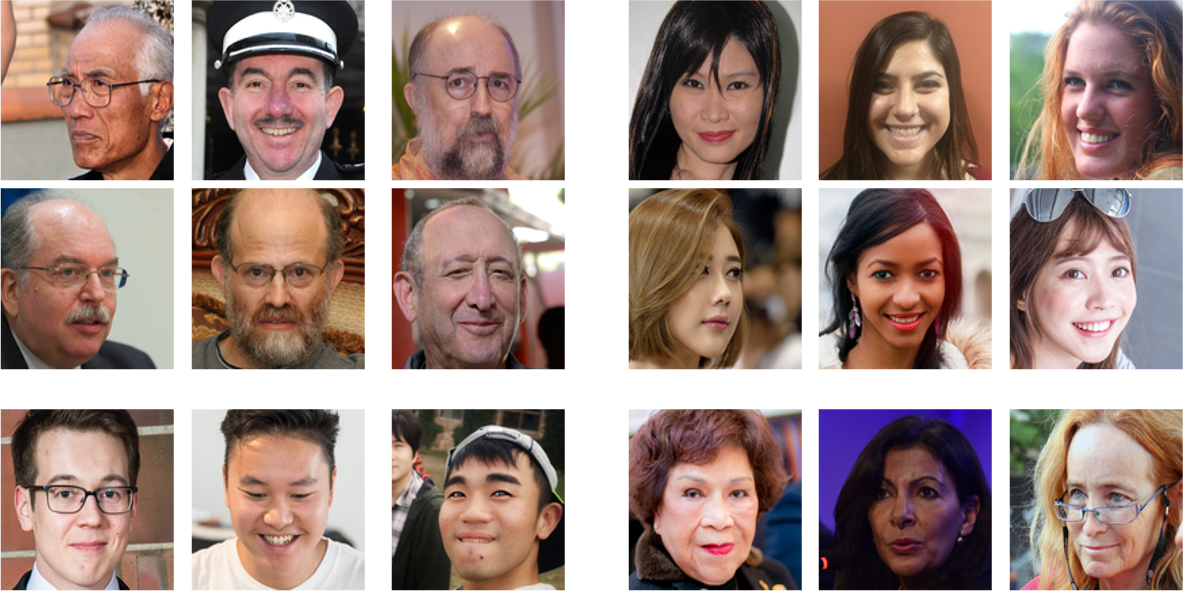} \label{fig_app:bffhq}
    \caption{BFFHQ: stands for biased FFHQ. Target attribute: gender, biased attribute: age. The samples in the top two rows are bias-aligned, while the samples in the bottom row are bias-conflict.}
    \end{minipage}
\end{figure}

\myparagraph{Biased Action Recognition (BAR)}
This dataset comes from the paper~\cite{nam2020learning} for real-world image testing. TThis benchmark aims to classify six actions  \{Climbing, Diving, Fishing, Racing, Throwing, Vaulting\} even though the places are biased. Target and bias attribute pairs are (Climbing,RockWall), (Diving,Underwater), (Fishing, WaterSurface), (Racing, APavedTrack), (Throwing, PlayingField), and (Vaulting, Sky). Bias-conflicting samples, for example, are  (Climbing, IceCliff), (Diving, Indoor), (Fishing, Forest), (Racing, OnIce), (Throwing, Cave), (Vaulting, Beach). There are $1,941$ samples for training and $6,54$ samples for testing. To split the training and validation samples, we used $10\%$ validation samples, i.e., $1,746$ images for training and $195$ validation. We download training datasets from the online repository of BAR.

\myparagraph{Biased FFHQ}
This BFFHQ benchmark was conducted in~\cite{lee2021learning, kim2021biaswap}. Target and bias attributes for bias-aligned samples are (Female, Young), and (Male, Old). Here, "Young" refers to people aged $10$ to $29$, while "old" refers to people aged $40$ to $59$. The bias-conflicting samples are (Female, Old) and (Male, Young). The number of training, validation, and test samples are $19,200$, $1,000$, and $1,000$, respectively.

\myparagraph{CelebA}
CelebA~\cite{DBLP:conf/iccv/LiuLWT15} is a common real-world face classification dataset, and each image has $40$ attributes. The goal is to classify the hair color ("blond" and "not blond") of celebrities, which has a spurious correlation with the gender ("male" or "female") attribute. In fact, only $6\%$ of blond hair color images are male. Therefore, ERM shows poor performance on the bias-conflicting samples. We report the average accuracy and the worst-group accuracy on the test dataset.

\myparagraph{CivilComments-WILDS} 
CivilComments-WILDS~\cite{DBLP:journals/corr/abs-1903-04561} is a dataset to classify whether an online comment is toxic or non-toxic. Each sentence is a real online comment, curated on the Civil Comments platform, a comment plug-in for independent news sites. The mentions of certain demographic identities (male, female, White, Black, LGBTQ, Muslim, Christian, and other religion) cause the spurious correlation with the label. Table~\ref{tab:portion of toxic} indicates the portion of toxic comments for each demographic identity.

\begin{table}[h!]
    \centering
    \resizebox{\textwidth}{!}{    
    \begin{tabular}{c|cccccccc}
        \thickhline
        Identity & Male & Female & White & Black & LGBTQ & Muslim & Christian & Other religions\\ \hline
        Portion(\%) of toxic & 14.9 & 13.7 & 28.0 & 31.4 & 26.9 & 22.4 & 9.1 & 15.3 \\
        \thickhline
    \end{tabular}
    }
    \caption{For each demographic identity, the portion of toxic comments in the CivilComments-WILDS.}
    \label{tab:portion of toxic}
\end{table}

%% file: appendix/exp.tex
\subsection{Baselines}
\label{app:exp_baseline}
In this section, we briefly describe how the baselines, such as LfF, JTT, Disen, GEORGE, BPA, CNC, and EIIL. Please refer to each paper for a detailed explanation because we briefly explain the algorithms.


(1) LfF~\cite{nam2020learning} trains the debiased model by weighting the bias-conflicting samples based on the "relative difficulty", computed by the two loss values from the biased model and the debiased model. To amplify the bias-conflicting samples, the authors employ generalized cross-entropy loss with parameter $\alpha=0.7$. We implement the LfF algorithm following the code officially offered by the authors. The loss functions  that this work proposes are as follows:
\begin{align*}
    \mc{L}_\text{LfF} &= W(z) \mc{L}_{\text{CE}} (C_d(z, y) + \lambda \mc{L}_{\text{GCE}} (C_b(z,y)),\\
    W(z)  &= \frac{\mc{L}_{\text{CE}}(C_b(z) , y)}{\mc{L}_{\text{CE}}(C_b(z),y) + \mc{L}_{\text{CE}}(C_d(z), y)}.
\end{align*}
Note that $W(z)$ is a relative difficulty and that GCE is a generalized cross-entropy. $z$ denotes feature, which is the output of the penultimate layer, and $C_{\cdot}$ is a fully connected layer.

(2) JTT~\cite{liu2021just} aims to debias by splitting the dataset into correctly and incorrectly learned samples. To do so, JTT trains the biased model first and splits the given training dataset as follows:
    \begin{equation}
        \mc{D}_\text{error-set} = \{(x, y) \text{ s.t. } y_\text{given} \neq  \argmax_c f_b(x)[c] \},
    \end{equation}
The ultimate debiased model is then trained by oversampling $\mc{D}_\text{error-set}$ with $\lambda_{up}$ times. We set $\lambda_{up}$ for all experiments as $1/\rho$. We reproduce the results by utilizing the official code offered by the authors. The main strength of PGD compared to JTT is that PGD does not need to set a hyperparameter $\lambda_{up}$.

(3) Disen~\citep{lee2021learning} aims to debias by generating abundant features from mixing features between samples. To do so, the author trains the biased and debiased model by aggregating features from both networks. This work also utilize the "relative difficulty" that is proposed in~\cite{lee2021learning}. We reproduced the results utilizing the official code offered by the authors. The loss function proposed in this work is as follows:
\begin{equation*}
    \mc{L}_\text{total} = \mc{L}_\text{dis} + \lambda_\text{swap}\mc{L}_\text{swap},
\end{equation*}
where 
\begin{align*}
    \mc{L}_\text{swap} &= W(z) \mc{L}_{\text{CE}} (C_d(z_\text{swap}, y) + \lambda_{\text{swap}_b} \mc{L}_{\text{GCE}} (C_b(z_\text{swap},\tilde{y})) \\
    \mc{L}_\text{dis} &= W(z) \mc{L}_{\text{CE}} (C_d(z, y) + \lambda_{\text{dis}} \mc{L}_{\text{GCE}} (C_b(z,y)),\\
    W(z)  &= \frac{\mc{L}_{\text{CE}}(C_b(z) , y)}{\mc{L}_{\text{CE}}(C_b(z),y) + \mc{L}_{\text{CE}}(C_d(z), y)}.
\end{align*}
Except for the swapped feature, $z_\text{swap}$ all terms are identical to those in LfF explanation.

(4) GEORGE~\cite{sohoni2020no} aims to debias by measuring and mitigating hidden stratification without requiring access to subclass labels. Assume there are $n$ data points $x_1,...,x_n \in \chi$ and associated superclass (target) labels $y_1,...,y_n \in \{1,\cdots ,C\}$. Furthermore, each datapoint $x_i$ is associated with a latent (unobserved) subclass label $z_i$. George consists of three steps. The author trains the biased model using ERM. Next, to estimate an approximate subclass (latent) label, apply UMAP dimensionality reduction~\cite{mcinnes2018uniform} to the features of a given training dataset at the ERM model. Here, GEORGE cluster the output of the reduced dimension for the data of each superclass into $K$ clusters, where $K$ is chosen automatically. The original paper contains a detailed description of the clustering process. Lastly, to improve performance on these estimated subclasses, GEORGE minimizes the maximum per-cluster average loss (\emph{i.e.,} $(x,y) \sim  {\hat{P}}_{\tilde{z}}$), by using the cluster as groups in the G-DRO objective~\cite{sagawa2019distributionally}. The loss function proposed in this work is as follows:
\begin{equation*}
    \mathop{\text{minimize}}_{L, f_{\theta}}\mathop{\,\,{\text{max}}}_{1\leq\tilde{z}\leq K}\,\,{\bb{E}}_{(x,y)\sim {\hat{P}}_{\tilde{z}}}[l(L \circ f_{\theta}(x), y)]
\end{equation*}
where $f_\theta$ and $L$ are parameterized feature extractor and classifier, respectively.

(5) BPA~\cite{Seo_2022_CVPR} aims to debias by using the technique of feature clustering and cluster reweighting. It consists of three steps. First, the author trains the biased model based on ERM. Next, at the biased model, cluster all training samples into $K$ clusters based on the feature, where $K$ is the hyperparameter.  Here, $h(x,y;\tilde{\theta})\in \mc{K}=\{1,\cdots, K\}$ denote the cluster mapping function of data $(x,y)$ derived by the biased model with parameter $\tilde{\theta}$. At the last step, BPA computes the proper importance weight, $w_k$ for the $k$-th cluster, where $k\in \mc{K}$ and the final objective of debiasing the framework is given by minimizing the weighted empirical risk as follows:
\begin{equation*}
    \mathop{\text{minimize}}_{\theta}\left\{{\bb{E}}_{(x,y)\sim P}\left[w_{h(x,y;\tilde{\theta})}(\theta)l(x,y;\theta)\right]\right\},
\end{equation*}
Concretely, for any iteration number $T$, the momentum method based on the history set $\mc{H}_{T}$, which is defined as:
\begin{equation*}
\mc{H}_T=\left\{1\leq t\leq T\,|\,\frac{\bb{E}_{(x,y)\sim P_k}\left[l((x,y);\theta_{t})\right]}{N_k}\right\},    
\end{equation*}
where $N_k$ is the number of the data belonging to $k$-th cluster.

(6) CNC~\cite{zhang2022correct} aims to debias by learning representation such that samples in the same class are close but different groups are far. CNC is composed of two steps: (1) inferring pseudo group label, (2) supervised contrastive learning. Get the ERM-based model $f$ and the pseudo prediction $\hat{y}$ first, then standard argmax over the final layer outputs of the model $f$. Next, CNC trains the debiased model based on supervised contrastive learning using pseudo prediction$\hat{y}$. The detailed process of contrastive learning for each iteration is as follows:

\begin{itemize}
    \item From the selected batch, sample the one anchor data (x,y).
    \item Construct the set of positives samples  $\{(x_m^{+}, y_m^{+} )\}$ which is belong to the batch, satisfying $y_m^{+}=y$ and $\hat{y}_{m}^{+}\neq \hat{y}$.
    \item Similarly, construct the set of negative samples  $\{(x_n^{-}, y_n^{-} )\}$ which is belong to the batch, satisfying $y_n^{-}\neq y$ and $\hat{y}_{n}^{-}=\hat{y}$.
    \item With the loss of generality, assume the cardinality of the positive and negative sets are $M$ and $N$, respectively.
    \item Weight update based on the gradient of the loss function $\hat{L}(f_{\theta};x,y)$, the detail is like below:
    \begin{equation*}
       \hat{L}(f_{\theta};x,y)=\lambda \hat{L}_{\text{con}}^{\text{sup}}(x,\{x_m^{+}\}_{m=1}^{M}, \{x_n^{-}\}_{n=1}^{N};f_{\text{enc}})+(1-\lambda)\hat{L}_{\text{cross}}(f_{\theta};x,y). 
    \end{equation*}
    Here, $\lambda \in [0,1]$ is a hyperparameter and $\hat{L}_{\text{cross}}(f_{\theta};x,y)$ is an average cross-entropy loss over $x$, the $M$ positives, and $N$ negatives. Moreover, $f_{\text{enc}}$ is the feature extractor part of $f_{\theta}$ and the detail formulation of $\hat{L}_{\text{con}}^{\text{sup}}(x,\{x_m^{+}\}_{m=1}^{M}, \{x_n^{-}\}_{n=1}^{N};f_{\text{enc}})$ is like below:
    \begin{equation*}
       -\frac{1}{M}\sum_{r=1}^{M}\log\frac{\exp({f_{\text{enc}}(x)}^{T}f_{\text{enc}}(x_r^{+})/\tau)}{\sum_{m=1}^{M}\exp({f_{\text{enc}}(x)}^{T}f_{\text{enc}}(x_m^{+})/\tau)+\sum_{n=1}^{N}\exp({f_{\text{enc}}(x)}^{T}f_{\text{enc}}(x_n^{-})/\tau)}.
    \end{equation*}
\end{itemize}

(7) EIIL~\cite{creager2021environment} proposes a novel invariant learning framework that does not require prior knowledge of the environment. EIIL is composed of three steps: (i) training based on ERM, (ii) environment inference (EI), and (iii) invariant learning (IL). In the first step, EIIL gets a biased model by minimizing ERM. Next, based on the feature extractor trained in (i), optimize the EI objective to infer environments from each training dataset. The object of EI is to sort training examples that maximally separate the spurious features so that they facilitate effective invariant learning. Lastly, to get a debiased model, optimize the classifier and feature extractor by minimizing the invariant learning objective.

\newpage
\section{Experiment details and additional analysis} 
\label{app:exp}

\subsection{Settings}
\label{appsec:exp_setting}
This section discusses how our experiment was set up, including architecture, image processing, and implementation details.

\myparagraph{Architecture.}
For the colored MNIST, we use simple convolutional networks consisting of three CNN layers with kernel size $4$ and channel sizes \{8, 32, 64\} for each layer. Also, we utilize average pooling at the end of each layer. Batch normalization and dropout techniques are used for regularization. Detailed network configurations are below. Similarly, for the multi-bias MNIST, we use four CNN layers with kernel size \{7, 7, 5, 3\} and channel size \{8, 32, 64, 128\}, respectively. For corrupted CIFAR, BAR, and BFFHQ, we utilize ResNet-18, which is provided by the open-source library torchvision. For CelebA, we follow the experimental setting of CNC~\cite{zhang2022correct}, which uses ResNet-50 as a backbone network. For CivilCOmments-WILDS, we use pretrained-BRET for the backbone network and exactly the same hyperparameters as in~\cite{zhang2022correct}.

\myparagraph{SimConv-1.}

\code{(conv1): Conv2d(3, 8, kernel\_size=(4, 4), stride=(1, 1))\\
(bn1): BatchNorm2d(8, eps=1e-05, momentum=0.1, affine=True, track\_running\_stats=True)\\
(relu1): ReLU()\\
(dropout1): Dropout(p=0.5, inplace=False)\\
(avgpool1): AvgPool2d(kernel\_size=2, stride=2, padding=0)\\
(conv2): Conv2d(8, 32, kernel\_size=(4, 4), stride=(1, 1))\\
(bn2): BatchNorm2d(32, eps=1e-05, momentum=0.1, affine=True, track\_running\_stats=True)\\
(relu2): ReLU()\\
(dropout2): Dropout(p=0.5, inplace=False)\\
(avgpool2): AvgPool2d(kernel\_size=2, stride=2, padding=0)\\
(conv3): Conv2d(32, 64, kernel\_size=(4, 4), stride=(1, 1))\\
(relu3): ReLU()\\
(bn3): BatchNorm2d(64, eps=1e-05, momentum=0.1, affine=True, track\_running\_stats=True)\\
(dropout3): Dropout(p=0.5, inplace=False)\\
(avgpool3): AdaptiveAvgPool2d(output\_size=(1, 1))\\
(fc): Linear(in\_features=64, out\_features=\textbf{\$num\_class}, bias=True)
}

\myparagraph{SimConv-2.}

\code{(conv1): Conv2d(3, 8, kernel\_size=(7, 7), stride=(1, 1))\\
(bn1): BatchNorm2d(8, eps=1e-05, momentum=0.1, affine=True, track\_running\_stats=True)\\
(relu1): ReLU()\\
(dropout1): Dropout(p=0.5, inplace=False)\\
(avgpool1): AvgPool2d(kernel\_size=3, stride=3, padding=0)\\
(conv2): Conv2d(8, 32, kernel\_size=(7, 7), stride=(1, 1))\\
(bn2): BatchNorm2d(32, eps=1e-05, momentum=0.1, affine=True, track\_running\_stats=True)\\
(relu2): ReLU()\\
(dropout2): Dropout(p=0.5, inplace=False)\\
(avgpool2): AvgPool2d(kernel\_size=3, stride=3, padding=0)\\
(conv3): Conv2d(32, 64, kernel\_size=(5, 5), stride=(1, 1))\\
(relu3): ReLU()\\
(bn3): BatchNorm2d(64, eps=1e-05, momentum=0.1, affine=True, track\_running\_stats=True)\\
(dropout3): Dropout(p=0.5, inplace=False)\\
(conv4): Conv2d(64, 128, kernel\_size=(3, 3), stride=(1, 1))\\
(relu4): ReLU()\\
(bn4): BatchNorm2d(128, eps=1e-05, momentum=0.1, affine=True, track\_running\_stats=True)\\
(dropout4): Dropout(p=0.5, inplace=False)\\
(avgpool): AdaptiveAvgPool2d(output\_size=(1, 1))\\
(fc): Linear(in\_features=128, out\_features=\textbf{\$num\_class}, bias=True)}

\subsection{PGD Implementation Details}

\myparagraph{Image processing}
We train and evaluate with a fixed image size. For colored MNIST case  ($28 \times 28$), multi-bias MNIST ($56 \times 56$), corrupted CIFAR ($32 \times 32$), and the remains ($224 \times 224$). For the CMNIST, MBMNIST, CCIFAR, BAR, and BFFHQ, we utilize random resize crop, random rotation, and color jitter to avoid overfitting. We use normalizing with a mean of $(0.4914, 0.4822, 0.4465)$, and standard deviation of $(0.2023, 0.1994, 0.2010)$ for CCIFAR, BAR, and BFFHQ cases.

\myparagraph{Implementation} For table~\ref{tab:control} and Table~\ref{tab:real_1} reported in Setion~\ref{sec:exp}, we reproduce all experimental results referring to other official repositories:
\footnote{https://github.com/alinlab/LfF}
\footnote{https://github.com/clovaai/rebias}
\footnote{https://github.com/kakaoenterprise/Learning-Debiased-Disentangled}
\footnote{https://github.com/anniesch/jtt}.The differences compared to the baseline codes are network architecture for CMNIST and usage of data augmentation. Here, we use the same architecture for CMNIST and data augmentation for all algorithms for a fair comparison. Except for JTT, all hyperparameters for CCIFAR and BFFHQ follow previously reported parameters in repositories. We grid-search for other cases, MNIST variants, and BAR. We set the only hyperparameter of PGD, $\alpha = 0.7$, as proposed by the original paper~\cite{zhang2018generalized}. A summary of the hyperparameters that we used is reported in Table~\ref{tab:hyperparam}.

\begin{table}[h]
    \centering
    \resizebox{\textwidth}{!}{
        \begin{tabular}{c|c|c|c|c|c|c|c}
            \thickhline
              & Colored MNIST & Multi-bias MNIST & Corrupted CIFAR & BAR & Biased FFHQ & CelebA & CivilComments-WILDS \\
              \hline
Optimizer     & SGD           & SGD              & Adam            & SGD                       & Adam & Adam & SGD \\
Batch size    & 128           & 32               & 256             & 16                        & 64 & 256 & 16 \\
Learning rate & 0.02          & 0.01             & 0.001           & 0.0005                    & 0.0001 & 0.0001 & 0.00001 \\
Weight decay  & 0.001         & 0.0001           & 0.001           & 1e-5                      & 0.0 & 0.01 & 0.01 \\
Momentum      & 0.9           & 0.9              & -               & 0.9                       & - & - & 0.9 \\
Lr decay      & 0.1           & 0.1              & 0.5             & 0.1                       & 0.1 & Cosine annealing & 0.1 \\
Decay step    & 40            & -                & 40              & 20                        & 32 & - & - \\
Epoch         & 100           & 100              & 200             & 100                       & 160 & 100 & 5  \\
GCE $\alpha$  & 0.7           & 0.7              & 0.7             & 0.7                       & 0.7 & 0.7 & 0.7 \\
        \thickhline
\end{tabular}}
\caption{Hyperparameter details}
\label{tab:hyperparam}
\end{table}

For Table \ref{tab:real_2} reported in Section~\ref{sec:exp}, we follow the implementation settings of CelebA and CivilComments-WILDS, suggested by \citet{Seo_2022_CVPR} and \citet{liu2021just}, respectively. A summary of the hyperparameters that we used is reported in Table \ref{tab:hyperparam}. We conduct our experiments mainly using a single Titan XP GPU for all cases.


\newpage

\section{Case studies on PGD}
\label{app:comp_anal}
In this section, we analyze PGD in many ways. Most analyses are based on the CMNIST dataset, and the experimental setting is the same as the existing setting unless otherwise noted. \revision{For example, all experiments used the same data augmentation, color jitter, resize crop, and random rotation.}

\subsection{Study 1: ablation study on GCE parameter $\alpha$}
\label{app:gce}

\begin{wraptable}[9]{r}{0.5\textwidth}
    \vspace{-0.25in}
    \centering
    \resizebox{0.5\textwidth}{!}{
    \begin{tabular}{ccccc}
            \thickhline
            Colored MNIST    & $\alpha=0.3$    & $\alpha=0.5$ & $\alpha=0.7$ & $\alpha=0.9$\\ \hline
            \multicolumn{5}{c}{Debiased model}                \\ \hline
            $\rho=0.5\%$        & 89.93\revision{\tiny $\pm$ 0.19}  & 94.70\revision{\tiny $\pm$ 0.23} & 96.88\revision{\tiny $\pm$ 0.28} & 96.79\revision{\tiny $\pm$ 0.04} \\ 
            $\rho=1\%$          & 96.32\revision{\tiny $\pm$ 0.12} & 97.27\revision{\tiny $\pm$ 0.16} & 97.35\revision{\tiny $\pm$ 0.12} & 97.59\revision{\tiny $\pm$ 0.04} \\ 
            $\rho=5\%$          & 98.80\revision{\tiny $\pm$ 0.05} & 98.82\revision{\tiny $\pm$ 0.02} & 98.62\revision{\tiny $\pm$ 0.14} & 98.78\revision{\tiny $\pm$ 0.02} \\ \hline
            \multicolumn{5}{c}{Biased model}                \\ \hline
            $\rho=0.5\%$        & 19.86\revision{\tiny $\pm$ 0.93}  & 18.70\revision{\tiny $\pm$ 0.89} & 18.12\revision{\tiny $\pm$ 0.09} & 17.39\revision{\tiny $\pm$ 0.86} \\ 
            $\rho=1\%$          & 22.40\revision{\tiny $\pm$ 2.12} & 21.04\revision{\tiny $\pm$ 1.54} & 19.71\revision{\tiny $\pm$ 0.24} & 19.12\revision{\tiny $\pm$ 1.12} \\ 
            $\rho=5\%$          & 53.24\revision{\tiny $\pm$ 1.18} & 49.46\revision{\tiny $\pm$ 1.13} & 43.97\revision{\tiny $\pm$ 1.75} & 39.40\revision{\tiny $\pm$ 2.65} \\ \hline
            \thickhline
    \end{tabular}
    }
    \caption{Ablation study on GCE parameter $\alpha$.}
    \label{tab:gce_ablation}
\end{wraptable}

The only hyper-parameter used in PGD is the GCE parameter $\alpha$. We experimented with this value at $0.7$ according to the protocol of LfF~\cite{nam2020learning}. However, we need to compare the various cases of $\alpha$. To analyze this, we run PGD with various $\alpha$ and report the performance in Table~\ref{tab:gce_ablation}. As in Table~\ref{tab:gce_ablation}, the debiased model performs best when the GCE parameter is $0.9$. This is because the biased model is fully focused on the bias feature, rather than the target feature, which can be seen from the unbiased test accuracy of the biased model, as in the bottom of Table~\ref{tab:gce_ablation}.

\subsection{Study 2: multi-stage vs single-stage} 
\label{app:stage}

\begin{wraptable}[6]{r}{0.5\textwidth}
    \vspace{-0.15in}
    \centering
    \resizebox{0.5\textwidth}{!}{
    \begin{tabular}{c|cc|cc}
            \thickhline
                            & \multicolumn{2}{c}{Training time}        & \multicolumn{2}{c}{Test Acc.}        \\\hline
            Colored MNIST    & Single-stage    & Multi-stage        & Single-stage & Multi-stage\\ \hline
            $\rho=0.5\%$        & 2h 53m 40s  & 33m 39s   & 92.19\revision{\tiny $\pm$ 0.12}  & 96.88\revision{\tiny $\pm$ 0.28}    \\ 
            $\rho=1\%$        & 2h 54m 45s  & 32m 28s   & 97.23\revision{\tiny $\pm$ 0.34}  & 97.35\revision{\tiny $\pm$ 0.12}    \\ 
            $\rho=5\%$        & 2h 49m 31s  & 34m 13s   & 98.44\revision{\tiny $\pm$ 0.17}  & 98.62\revision{\tiny $\pm$ 0.14}    \\ 
            \thickhline
    \end{tabular}
    }
    \caption{Multi-stage vs Single-stage}
    \label{tab:single_stage}
\end{wraptable}
PGD computes the per-sample gradient norm only once, between training the biased model and the debiased model. However, an update of the per-sample gradient can be performed repeatedly at each epoch (i.e., single-stage). In other words, PGD can be modeled to run the following loop: updating the biased model, updating the sampling probability, and updating the debiased model. In this section, we justify why we use the multi-stage approach. We report the performance of multi-stage and single-stage PGD on the colored MNIST dataset. As in Table~\ref{tab:single_stage}, the single-stage method has two characteristics: (1) it requires more training time than the multi-stage method. (2) It has lower unbiased accuracy compared to the multi-stage method. The longer training time is due to the high computational resources required to compute the per-sample gradient norm. Moreover, because the single-stage method's sampling probability changes the training distribution over epochs, the debiased model suffers from unstable training and loses debiasing performance.

\subsection{Study 3: Resampling vs Reweighting}
\label{app:reweight}

\begin{wraptable}[6]{r}{0.4\textwidth}
    \vspace{-0.15in}
    \centering
    \resizebox{0.4\textwidth}{!}{
    \begin{tabular}{ccc}
            \thickhline
            Colored MNIST    & Resampling    & Reweighting\\ \hline
            $\rho=0.5\%$        & 96.88\revision{\tiny $\pm$ 0.28}  & 94.70\revision{\tiny $\pm$ 0.4} \\ 
            $\rho=1\%$        & 97.35\revision{\tiny $\pm$ 0.12}  & 97.20\revision{\tiny $\pm$ 0.1} \\ 
            $\rho=5\%$        & 98.62\revision{\tiny $\pm$ 0.14}  & 98.51\revision{\tiny $\pm$ 0.03} \\ 
            \thickhline
    \end{tabular}
    }
    \caption{Reweighting vs resampling}
    \label{tab:reweighting}
\end{wraptable}

To support our algorithm design, we provide further experimental analysis, i.e., resampling versus reweighting. Reweighting~\cite{nam2020learning, lee2021learning} and resampling~\cite{liu2021just} are the two main techniques to debias by up-weighting bias-conflicting samples. PGD is an algorithm that modifies the sampling probability by using the per-sample gradient norm. To check whether PGD works with reweighting, we examine the results of PGD with reweighting on colored MNIST dataset and report them in Table~\ref{tab:reweighting}. We compute the weight for each sample as follows: $w(x_i,y_i) = |\mc{D}_n| \times \frac{\norm{\nabla_\theta \mc{L}_{\text{CE}}(x_i,y_i;\theta_b)}_2}{\sum_{(x_i,y_i) \in \mc{D}_n} \norm{\nabla_\theta \mc{L}_{\text{CE}}(x_i,y_i;\theta_{b})}_2}$. As in Table~\ref{tab:reweighting}, PGD with resampling slightly outperforms PGD with reweighting. As argued in~\cite{an2020resampling}, this gain from resampling can be explained by the argument that resampling is more stable and better than reweighting.

\newpage
\subsection{Study 4: Analysis of the pure effect of gradient-norm-based score}
\label{app:effect_grad_score}

\begin{wraptable}[8]{r}{0.5\textwidth}
    \vspace{-0.1in}
    \centering
    \resizebox{0.5\textwidth}{!}{
    \begin{tabular}{cccc}
            \thickhline
            Method    & $\rho=0.5\%$    &$\rho=1\%$        & $\rho=5\%$ \\ \hline
            LfF        & 91.35\revision{\tiny $\pm$ 1.35}   & 96.88\revision{\tiny $\pm$ 0.2}   & 98.18\revision{\tiny $\pm$ 0.05}  \\ 
            PGD + Loss        & 30.63\revision{\tiny $\pm$ 2.23}  & 34.04\revision{\tiny $\pm$ 3.00}   & 78.48\revision{\tiny $\pm$ 1.41}  \\ 
            LfF + Gradient     & 93.29\revision{\tiny $\pm$ 0.39}  & 97.55\revision{\tiny $\pm$ 0.24}   & 98.37\revision{\tiny $\pm$ 0.20}  \\ 
            PGD        & 96.88\revision{\tiny $\pm$ 0.28}  & 98.35\revision{\tiny $\pm$ 0.12}   & 98.62\revision{\tiny $\pm$ 0.14}  \\ 
            \thickhline
    \end{tabular}
    }
    \caption{loss score vs gradient norm score}
    \label{tab:effect_grad_score}
\end{wraptable}

\revision{PGD performance improvement comes not only from the two-stage and resampling modules that we wrote about in Appendix~\ref{app:stage} and Appendix~\ref{app:reweight}, but also from the gradient score. To verify this, we report the following two results: (1) resample based on per-sample loss rather than per-sample gradient norm in PGD, and (2) change the relative difficulty score of LfF to gradient norm. As shown in Table~\ref{tab:effect_grad_score}, we can conclude the following two results: (1) loss of the last epoch of the first stage in a two-stage approach is not suitable for resampling, and (2) the results of replacing the relative difficulty metric of LfF with a gradient norm show that the gradient norm has better discriminative properties for bias-conflicting samples than loss.}

\begin{wraptable}[6]{r}{0.5\textwidth}
    \vspace{-0.1in}
    \centering
    \resizebox{0.5\textwidth}{!}{
    \begin{tabular}{ccccc}
            \thickhline
                                    & Vanilla & LfF & Disen & ours \\ \hline
            Computation time        & 14m 59s  & 21m 35s   & 23m 18s & 33m 31s \\ \thickhline
                                    & Step 1 & Step 2 & Step 3  &\\ \hline
            Computation time        & 15m 19s  & 1m 26s   & 16m 46s &   \\ \thickhline
            
    \end{tabular}
    }
    \vspace{-5pt}
    \caption{Computation cost}
    \label{tab:compute_time}
\end{wraptable}

\subsection{Study 5: Computation cost}
\label{app:compute_cost}
Debiasing algorithms require an additional computational cost. To evaluate the computational cost of PGD, we report the training time in Table~\ref{tab:compute_time}. We conduct this experiment by using the colored MNIST with $\rho=0.5\%$. As in the top of Table~\ref{tab:compute_time}, we report the training time of four methods: vanilla, LfF, Disen, and PGD. Here, PGD spends a longer amount of training time. This is because there is no module for computing per-sample gradient in a batch manner. At the bottom of Table~\ref{tab:compute_time}, we report part-by-part costs to see which parts consume the most time. Note that Steps 1, 2, and 3 represent training the biased model, computing the per-sample gradient-norm, and training the debiased model, respectively. We can conclude with the following two facts. (1) Step 2 (computing the per-sample gradient norm and sampling probability) spends $4.3\%$ of training time. (2) Resampling based on the modified sampling probability $h(x)$ requires an additional cost of 1m 27s by seeing the difference between the computing times of Step 3 and Step 1.

\subsection{Study 6: What if PGD runs on unbiased datasets} 
\label{app:unbiased}

\vspace{0.2in}
\begin{wraptable}[9]{r}{0.5\textwidth}
    \vspace{-0.15in}
    \centering
    \resizebox{0.5\textwidth}{!}{
    \begin{tabular}{cccc}
            \thickhline
            Vanilla    & LfF & Disen & Ours \\ \hline
            \multicolumn{4}{c}{$\rho=90\%$ Colored MNIST}                \\ \hline
            99.04\revision{\tiny $\pm$ 0.05} & 98.75\revision{\tiny $\pm$ 0.07} & 99.31\revision{\tiny $\pm$ 0.1} & 98.43\revision{\tiny $\pm$ 0.11} \\  \hline
            \multicolumn{4}{c}{Natrual CIFAR10}                \\ \hline
            94.24\revision{\tiny $\pm$ 0.01}  & - & - & 94.79\revision{\tiny $\pm$ 0.02} \\ 
            \thickhline
    \end{tabular}
    }
    \caption{Results on unbiased CMNIST and natural CIFAR10 cases.}
    \label{tab:natural}
\end{wraptable}

We examine whether PGD fails when an unbiased dataset is given. To verify this, we report two types of additional results: (1) unbiased CMNIST (\emph{i.e.,} $\rho = 90\%$) and (2) conventional public dataset (\emph{i.e.,} CIFAR10). We follow the experimental setting of CMNIST for the unbiased CMNIST case. On the other hand, we train ResNet18~\cite{he2016deep} for CIFAR10 with the SGD optimizer, $0.9$ momentum, $5e-4$ weight decay, $0.1$ learning rate, and Cosine Annealing LR decay scheduler. As shown in Table~\ref{tab:natural}, PGD does not suffer significant performance degradation in unbiased CMNIST. Furthermore, it performs better than the vanilla model on the CIFAR10 dataset. This means that the training distribution that PGD changes do not cause significant performance degradation. In other words, PGD works well, regardless of whether the training dataset is balanced or unbiased.

\newpage
\section{Empirical evidence of PGD}
\label{app:second_eas_exp}

\revision{As same with the setting of Appendix~\ref{app:comp_anal}, in this section, we also use the existing CMNIST setting, such as data augmentation, hyperparameters.}

\subsection{Correlation between gradient norm and bias-alignment of the CMNIST} 
\label{app:gradient_norm}
\begin{figure*}[h]
    \centering
    \subfloat[Colored MNIST $\rho=0.5\%$]{\includegraphics[width=0.32\textwidth]{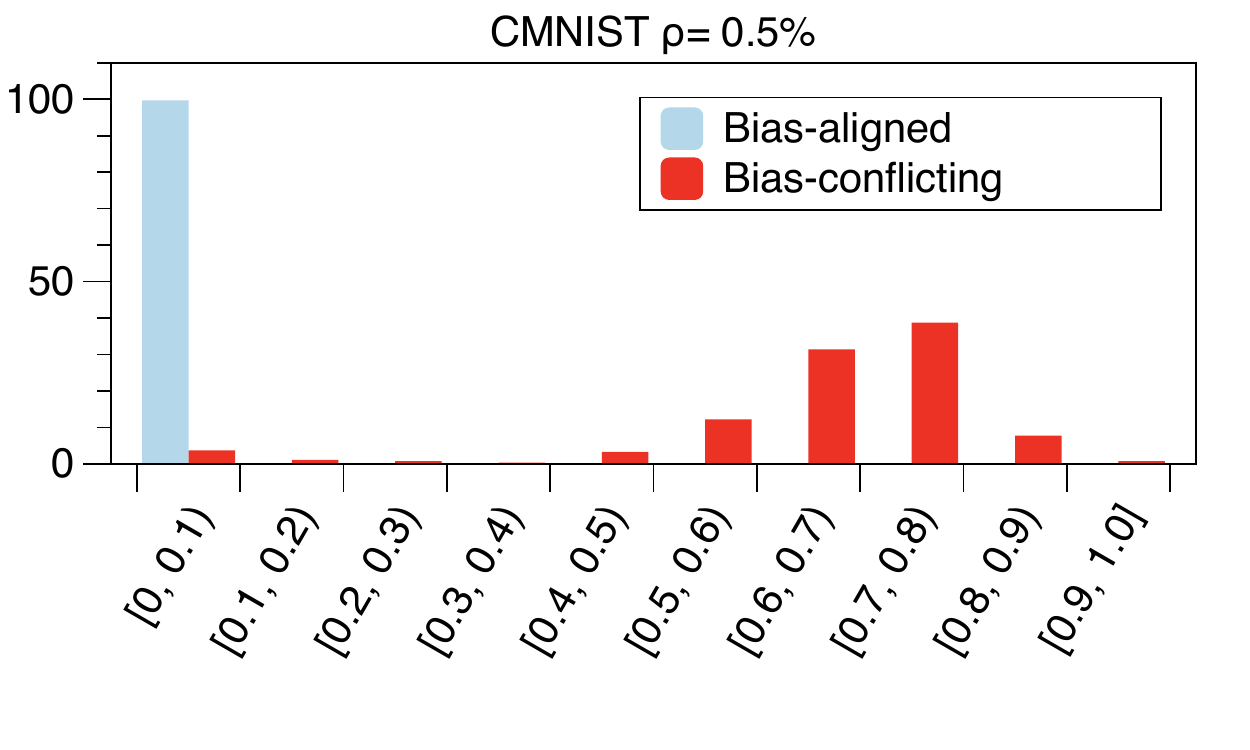} \label{fig:hist_0.5}}
    \subfloat[Colored MNIST $\rho=1\%$]{\includegraphics[width=0.32\textwidth]{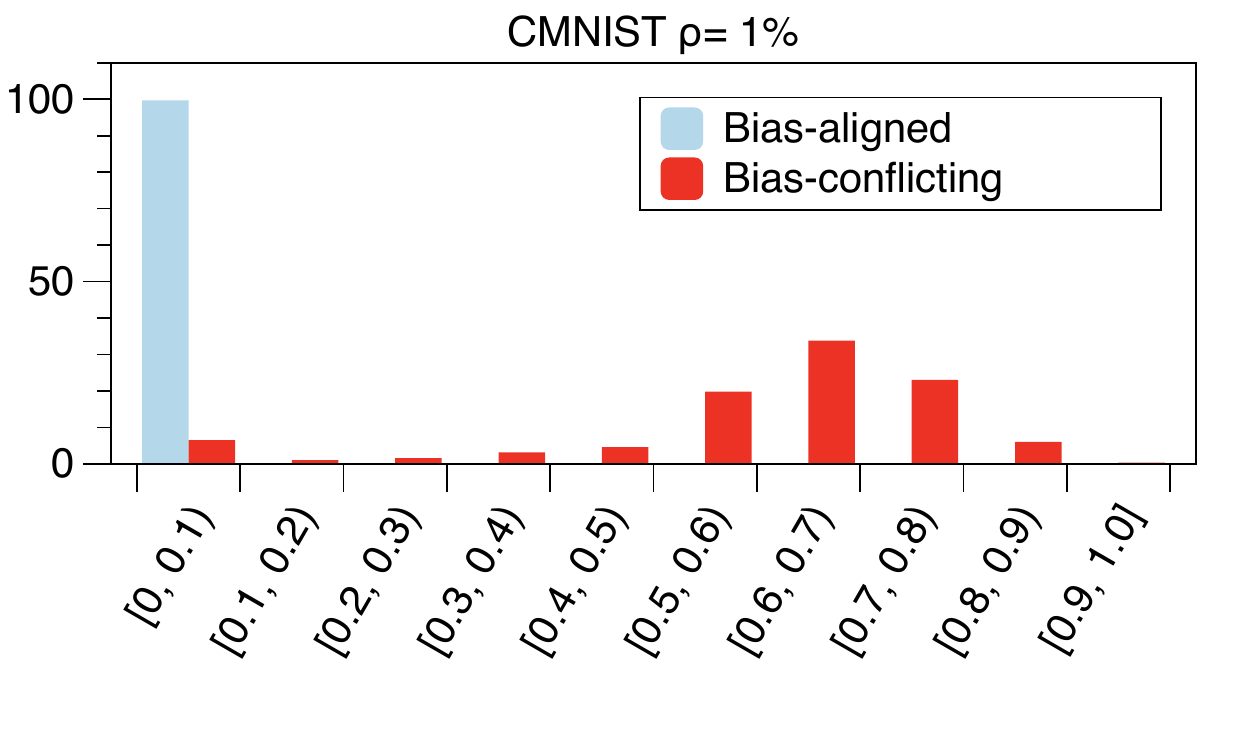} \label{fig:hist_1}}
    \subfloat[Colored MNIST $\rho=5\%$]{\includegraphics[width=0.32\textwidth]{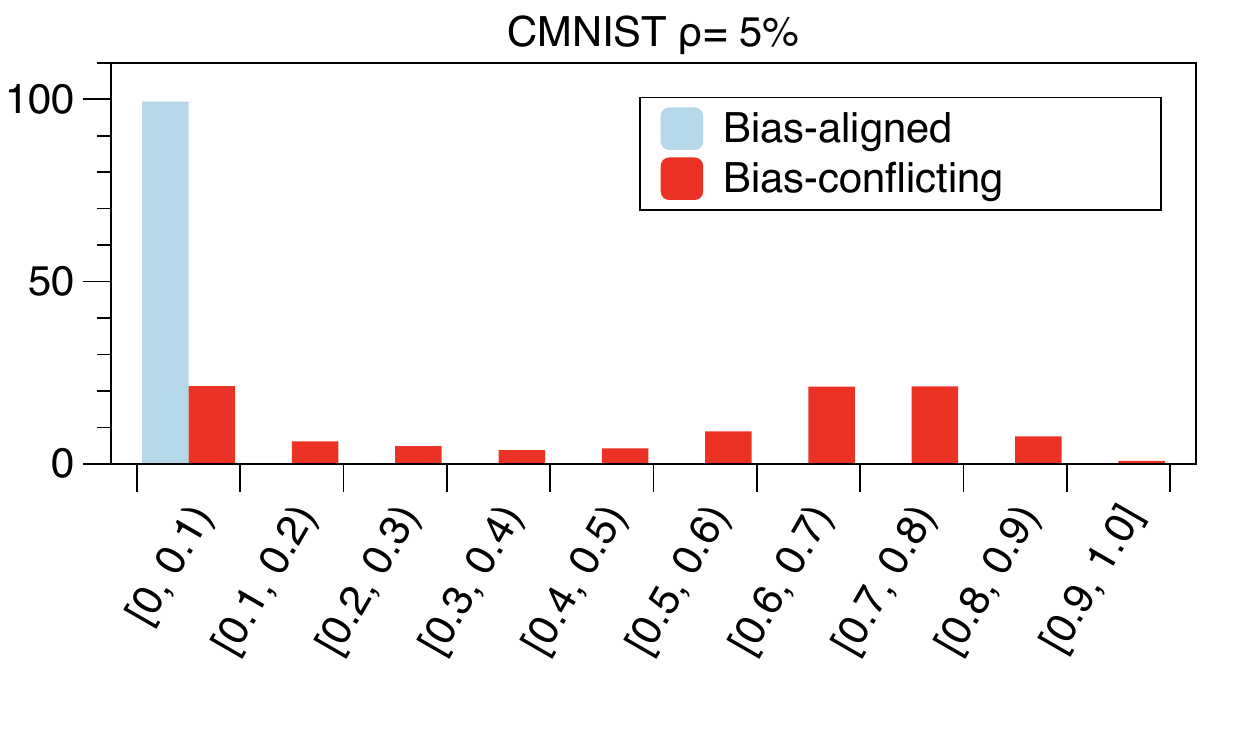} \label{fig:hist_5}}
    \caption{Histogram of per-sample gradient norm.}
    \label{fig:ab_hist}
\end{figure*}

To check if the per-sample gradient norm efficiently separates the bias-conflicting samples from the bias-aligned samples, we plot the gradient norm distributions of the colored MNIST (CMNIST). For comparison, we normalized the per-sample gradient norm as follows: $\frac{\norm{\nabla_\theta \mc{L}_{\text{CE}}(x_i,y_i;\theta_b)}}{\max_{(x_i,y_i) \in \mc{D}_n} \norm{\nabla_\theta \mc{L}_{\text{CE}}(x_i,y_i;\theta_{b})}}$. As in Figure~\ref{fig:ab_hist}, the bias-aligned sample has a lower gradient norm (blue bars) than the bias-conflicting samples (red bars).

\subsection{PGD does not learn only the second-easiest feature}
\label{app:easier_to_learn}
We provide the results of the following experimental settings: The target feature is color, and the bias feature is digit shape, i.e., the task is to classify the color, not the digit shape. Let us give an example of this task. When one of the target classes is red, this class is aligned with one of the digits (e.g., "0"). In other words, the bias-aligned samples in this class are (Red, "0"), and the bias-conflicting samples are (e.g., {(Red, "1"), (Red, "2"),.., (Red, "9")}). Note that, as shown in LfF~\citep{nam2020learning}, color is empirically known to be easier to learn than digit shape; we think that the above scenario reflects the concern: whether PGD is only targeting the second-easiest feature (digit shape). Therefore, if the concern is correct, PGD may fail in this color target MNIST scenario since the model will learn digit shape. However, as shown in the table below, vanilla, PGD, and LfF perform well in that case.

\begin{table}[h]
    \centering
    \resizebox{\textwidth}{!}{
        \begin{tabular}{c|cccccc}
            \thickhline
               & Vanilla (Digit) & \textbf{Vanilla (Color)} & LfF (Digit) & \textbf{LfF (Color)} & PGD (Digit) & \textbf{PGD (Color)} \\
              \hline
$\rho=0.5\%$     & 60.94 & \textbf{90.33} & 91.35  & \textbf{91.16} & 96.88 & \textbf{98.92}  \\
$\rho=1\%$    & 79.13 & \textbf{92.53} & 96.88 & \textbf{96.12} & 98.35 & \textbf{99.58}  \\
$\rho=5\%$ & 95.12 & \textbf{96.96} & 98.18 & \textbf{99.11} & 98.62 & \textbf{99.7}  \\ \thickhline
\end{tabular}}
\caption{Digit target MNIST vs Color target MNIST}
\label{tab:second_easiest}
\end{table}
We can also support this result by seeing the distribution of the normalized gradient norms, $\norm{\nabla_\theta \mc{L}_{\text{CE}}(x_i,y_i;\theta_b)}_{2}/\max_{(x_i,y_i) \in \mc{D}_n} \norm{\nabla_\theta \mc{L}_{\text{CE}}(x_i,y_i;\theta_{b})}_{2}\in\,[0,1]$, extracted from the biased model $\theta_b$ (trained in Step 1 of Algorithm \ref{alg:pgd} of Section \ref{sec:method}). 


\begin{table}[h]
    \centering
    \resizebox{\textwidth}{!}{
        \begin{tabular}{c|cccccccccc}
            \thickhline
               & [0.0,0.1)& [0.1,0.2) & [0.2,0.3) & [0.3,0.4) & [0.4,0.5) & [0.5,0.6) & [0.6,0.7)&[0.7,0.8)&[0.8,0.9)&[0.9,1.0]\\
              \hline
Bias-aligned & 53504 & 88 & 40 & 21 & 13 & \textbf{18} & \textbf{16} & \textbf{17} & \textbf{8} & \textbf{4}\\
Bias-conflicting & 212 & 18 & 7 & 10 & 8 & \textbf{6} & \textbf{5} & \textbf{4} & \textbf{0} & \textbf{1}\\ \thickhline
\end{tabular}}
\caption{Number of samples at each bin: Color target MNIST ($\rho=0.5\%$)}
\label{tab:stat_reverse}
\end{table}

The numbers filled in Table~\ref{tab:stat_reverse} are the number of data items belonging to each bin category. We can check that \emph{there are no bias-conflicting samples whose gradient norm is significantly larger than the bias-aligned samples.} In other words, PGD \emph{does not force the debiased model to learn the digit shape (i.e., the second-easiest feature) in this scenario.} This scenario brings similar performance to Vanilla.

\begin{figure*}[h]
    \centering
    \subfloat[Colored MNIST $\rho=0.5\%$]{\includegraphics[width=0.32\textwidth]{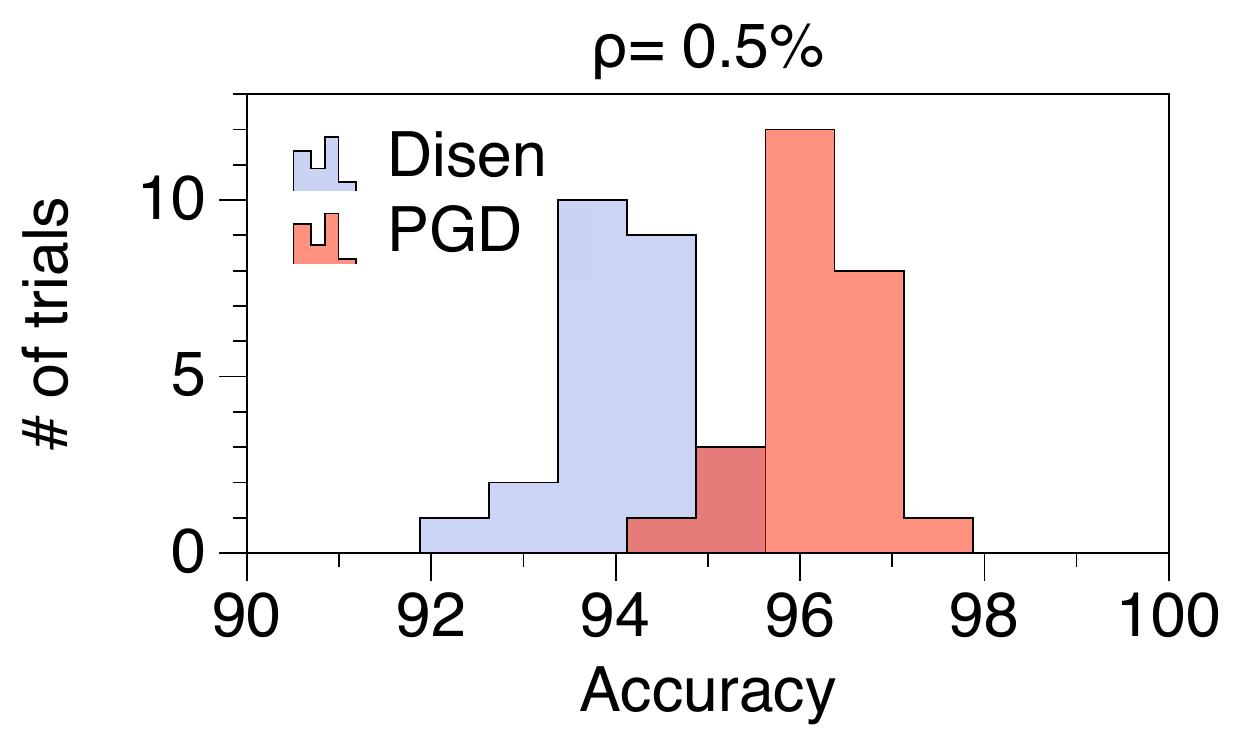} \label{fig:t_hist_0.5}}
    \subfloat[Colored MNIST $\rho=1\%$]{\includegraphics[width=0.32\textwidth]{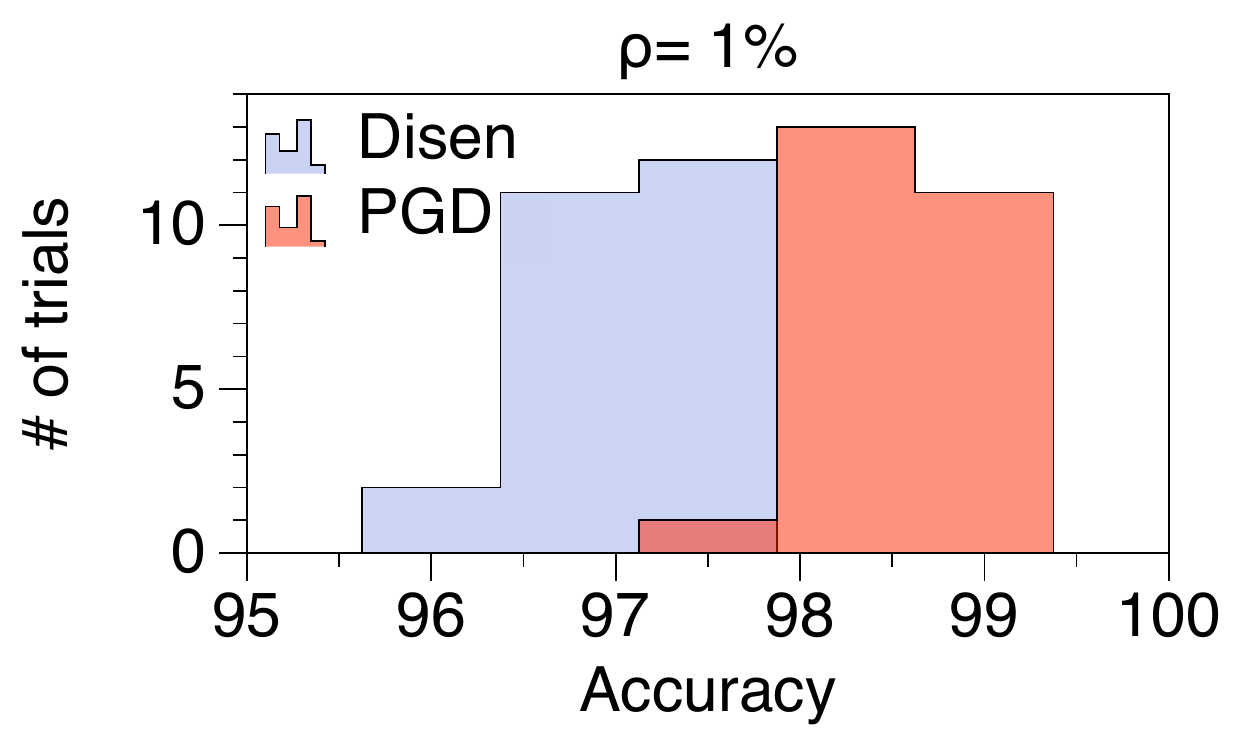} \label{fig:t_hist_1}}
    \subfloat[Colored MNIST $\rho=5\%$]{\includegraphics[width=0.32\textwidth]{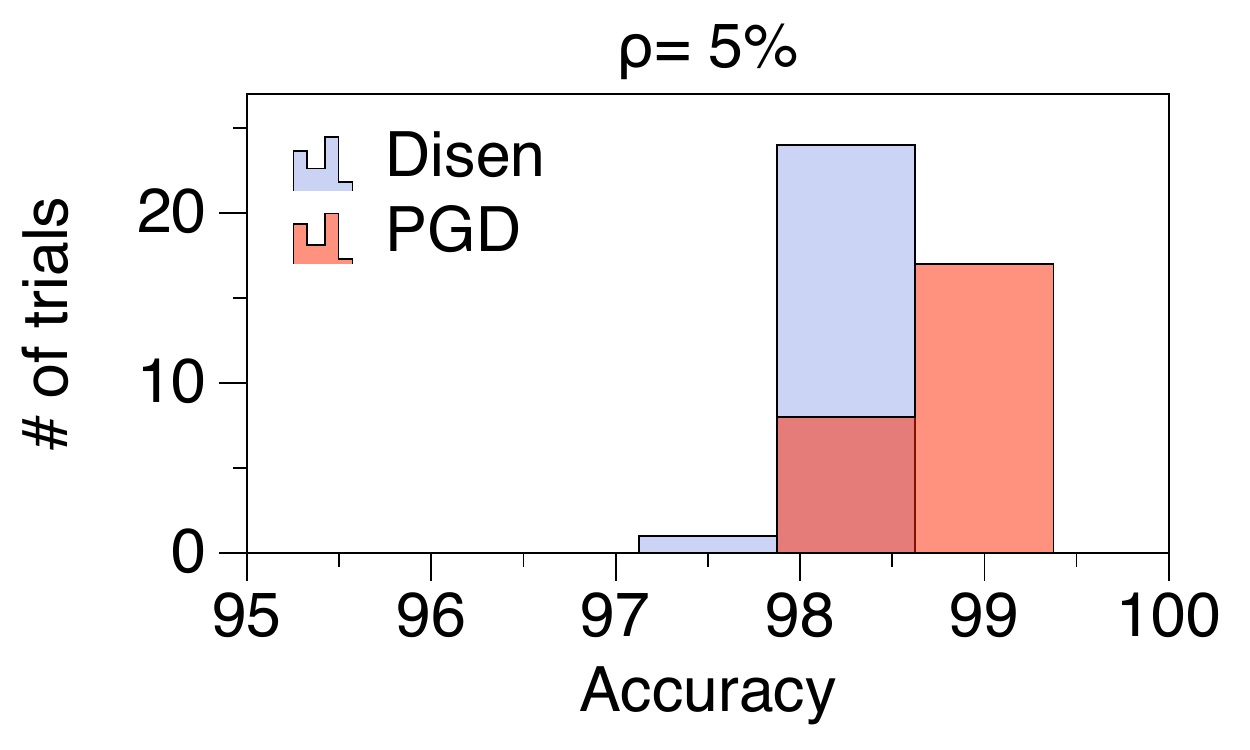} \label{fig:t_hist_5}}
    \caption{Histogram of unbiased test accuracy among 25 trials for each.}
    \label{fig:hist_trial}
\end{figure*}

\revision{
\subsection{Histogram of the results of 25 trials in CMNIST}
\label{app:histogram_trial}
For in-depth analysis, we provide the results of 25 tests on CMNIST in Figure~\ref{fig:hist_trial}. We compare with Disen, which shows the best performance except for PGD. However, very few cases overlap, as shown in Figure~\ref{fig:hist_trial}. We conduct a t-test for a more in-depth analysis of this. When the alternative hypothesis is established that PGD is superior to Disen, the p-value for it has values of 0.01572, 0.01239, and 0.29370, respectively. In other words, it can be said that as the bias becomes worse, the superiority of PGD stands out.
}

\section{Additional Results on other datasets}   
\label{app:add_exp}

\subsection{CelebA with ResNet18 Setting.}

\begin{table}[h!]
\centering
\vspace{-5pt}
\resizebox{0.75\textwidth}{!}{
\begin{tabular}{cc|ccccc}
\thickhline
& & Vanilla & LfF & GEORGE & BPA & Ours \\ \hline
 \multicolumn{1}{c}{\multirow{2}{*}{{CelebA}{$^\dagger$}}}             & \multicolumn{1}{|c|}{Avg.}  &         80.52 & 84.89 & 83.13  & 90.18  & 89.27 \\ 
                                     & \multicolumn{1}{|c|}{Worst} & 41.02 & 57.96 & 65.45 &    82.54 & \textbf{82.73} \\ \thickhline
\end{tabular}}
\caption{Average and worst test accuracy with CelebA setting of~\cite{Seo_2022_CVPR}. The results of comparison algorithms for CelebA $^\dagger$ are the results reported in~\cite{Seo_2022_CVPR}. The best worst accuracy is indicated in \textbf{bold}.}
\label{tab:real_3}
\end{table}

We reported the results of CelebA in Table \ref{tab:real_2} of section \ref{sec:exp}, following the settings of  \citep{zhang2022correct}. For comparison with more diverse algorithms, we further report the CelebA results according to the settings of \cite{Seo_2022_CVPR}. Note that the \cite{zhang2022correct} and \cite{Seo_2022_CVPR} used a different model, each using ResNet50 and ResNet18, respectively. As in Table~\ref{tab:real_3}, PGD shows competitive performance among the other baselines.

%% file: appendix/theory.tex
\newpage
\section{Backgrounds For Theoretical Analysis}
\label{appsec:backgrounds}
\subsection{Notations Summary.}
\label{appsubsec:notation}

For convenience, we describe notations used in Section~\ref{sec:analysis}, Appendix~\ref{appsec:backgrounds},~\ref{appsec:theorem_1}, and~\ref{appsec:theorem_2}.

\newcolumntype{Y}{>{\RaggedRight\arraybackslash}p{1.6cm}} 
\newcolumntype{L}{>{\RaggedRight\arraybackslash}X}

\begin{table}[h]
\caption{Notation Table}
\label{tab:not1}
\resizebox{1.0\textwidth}{!}{
\begin{tabular}{clc}
\thickhline
\hline
Notation & Description & Remark\\
\hline\hline
    
\multicolumn{3}{c}{Variables}\\\hline
  $(x,y)$ 
  & (image, label) 
  & $x \in \bb{R}^d, y \in C=\{1,...,c\}$\\

  $y_{true}(x)$ & the true label of image $x$
  & labeled by the oracle model $f(y|x,\theta^\star)$\\
  
  $\theta$ 
  & model parameter
  & -\\ 

  $\theta^\star$ 
  & oracle model parameter 
  & satisfying $f(y_{true}(x)|x,\theta^\star)=1$ for any $x$\\ 

  $\mc{D}_{n}$ & training dataset 
  & composed of $\{(x_i,y_i)\}_{i=1}^{n} \sim p(x,y|\theta^\star)$\\

  $h(x)$ & sampling probability of each sample in $\mc{D}_{n}$
  & satisfying $h(x) \in \mc{H}$ \\

\midrule

\multicolumn{3}{c}{Distributions}\\\hline
   $\mc{P}(x)$ 
  & general distribution of input image $x$ 
  & - \\ 

  $p(x)$  
  & distribution of training image
  & - \\

  $q(x)$  
  & distribution of test image 
  & - \\

  $f(y|x,\theta)$ 
  & conditional distribution with model parameter $\theta$ 
  & - \\
  
  $\mc{P}(x,y|\theta)$  
  & general joint distribution with model parameter $\theta$ 
  & $\mc{P}(x)f(y|x,\theta)$ \\

  $p(x,y|\theta^\star)$  
  & joint distribution of the training dataset 
  & $p(x)f(y|x,\theta^\star)$\\ 

  $q(x,y|\theta^\star)$  
  & joint distribution of the test dataset 
  & $q(x)f(y|x,\theta^\star)$ \\

\midrule
  
\multicolumn{3}{c}{Estimators}\\\hline
  ${\hat{\theta}}_{h(x),\mc{D}_{n}}$ 
  & MLE solution on the $\mc{D}_{n}$ with $h$
  & $\triangleq \argmax_{\theta} \quad \sum_{i=1}^{n}\,h(x_i)\log f(y_i|x_i, \theta)$ \\

  ${\hat{\theta}}_{U(x),\mc{D}_{n}}$ 
  & MLE solution on the $\mc{D}_{n}$ with uniform distribution $U$
  & solution of ERM  \\

\midrule
\multicolumn{3}{c}{Fisher Information}\\\hline
   $I_{\mc{P}(x)}(\theta)$
  & Fisher Information 
  & $ \bb{E}_{(x,y)\sim\mc{P}(x)f(y|x,\theta)}[\nabla_{\theta}\,\log\, f(y|x,\theta){{\nabla^\top_{\theta}}}\,\log\,f(y|x,\theta)]$ \\

  $\hat{I}_{h(x)}(\theta)$ 
  & Empirical Fisher Information
  & $ \sum_{i=1}^{n}h(x_i)\nabla_{\theta}\log\,f(y_i|x_i,\theta)\nabla^\top_{\theta}\log\,f(y_i|x_i,\theta)$ \\

  \midrule
\multicolumn{3}{c}{Set}\\\hline
  $\mc{H}$ 
  & set of all possible $h(x)$ on $\mc{D}_{n}$ 
  & $\{h(x)|\textstyle\sum_{(x_i,y_i) \in \mc{D}_{n}} h(x_i) = 1$ and $h(x_i) \ge 0 \quad \forall (x_i,y_i) \in \mc{D}_{n}\}$ \\

  $\mc{M}$ 
  & set of all possible marginal $\mc{P}(x)$ on input space $\mc{X}$ 
  & $\{\mc{P}(x)|\int_{x\in\mc{X}}\mc{P}(x)\,dx=1\}$ \\

  $\mc{W}$ 
  & set of all possible $(x,y_{true}(x))$
  & -\\

   $supp(\mc{P}(x,y|\theta))$
  & Support set of $\mc{P}(x,y|\theta)$
  & $\{(x,y)\in X \times \{1, \cdots,\,c\}\,\,|\,\,\mc{P}(x,y|\theta)\neq 0\},\,\,\forall\,\, \mc{P}(x,y|\theta)$ \\

\midrule
\multicolumn{3}{c}{Order notations in probability}\\\hline

 $O_{\text{p}}$
& Big $O$, stochastic boundedness
& - \\

 $o_{\text{p}}$
& Small $o$, convergence in probability
& - \\

\midrule
\multicolumn{3}{c}{Toy example (Appendix~\ref{appsec:theorem_2})}\\\hline

 $M$
& set of majority (\emph{i.e.,} bias-aligned) samples
& - \\

 $m$
& set of minority (\emph{i.e.,} bias-conflicting) samples
& - \\

 $g_M(\theta)$
& gradient of samples in $M$ at $\theta$
& - \\

 $g_m(\theta)$
& gradient of samples in $m$ at $\theta$
& - \\

 $h_M^\star(x)$
& Optimal sampling probability of samples in $M$
& - \\

 $h_m^\star(x)$
& Optimal sampling probability of samples in $m$
& - \\
  
\thickhline
\end{tabular}}
\end{table}

\subsection{Main Assumption}
\label{appsubsec:main_assumption}
Here, we organize the assumptions that are used in the proof of Theorems. These are basically used when analyzing models through Fisher information. The assumptions are motivated by~\cite{sourati2016asymptotic}.

\begin{assum}
\,
\label{app:main_assumption}

\begin{enumerate}[start=0,label={(A\arabic*).}, ref=\arabic*]
\item The general joint distribution $\mc{P}(x,y|\theta)$ is factorized into the conditional distribution $f(y|x,\theta)$ and the marginal distribution $\mc{P}(x)$, not depend on model parameter $\theta$, that is:
\begin{equation}
\mc{P}(x,y|\theta)=\mc{P}(x)f(y|x,\theta). 
\end{equation}
Thus, the joint distribution is derived from model parameter $\theta$ and the marginal distribution $\mc{P}(x)$, which is determined from the task that we want to solve. Without loss of generality, we match the joint distribution's name with the marginal distribution. 
\item \emph{(Identifiability):} The CDF $\bm{\mc{P}}_{\theta}$ (whose density is given by $\mc{P}(x,y|\theta)$) is identifiable for different parameters. Meaning that for every distinct parameter vectors $\theta_1$ and $\theta_2$ in $\Omega,$ $\bm{\mc{P}}_{\theta_1}$ and $\bm{\mc{P}}_{\theta_2}$ are also distinct. That is,
\begin{equation*}
\forall \theta_1 \neq \theta_2 \in \Omega, \quad \exists A \subseteq X \times \{1, \cdots,\,c\}\quad \text{s.t.} \quad \bm{\mc{P}}_{\theta_1}(A)\neq \bm{\mc{P}}_{\theta_2}(A),
\end{equation*}
where $X$, $\{1, \cdots, c\}$ and $\Omega$ are input, label, and model parameter space, respectively.
\item The joint distribution $\bm{\mc{P}}_{\theta}$ has common support for all $\theta\in\Omega$.

\item \emph{(Model Faithfulness)}:
For any $x\in X$, we assume an oracle model parameter $\theta^\star$ that generates $y_{\text{true}}(x)$, a true label of input $x$ with the conditional distribution $f(y_{\text{true}}(x)|x, \theta^{\star})=1$. 

\item \emph{(Training joint):} Let $p(x)$ denote the training marginal with no dependence on the parameter. Then, the set of observations in $\mc{D}_{n}\triangleq\{(x_1, y_1). \cdots \, (x_n, y_n)\}$ are drawn independently from the training/proposal joint distribution of the form $p(x,y|\theta^{\star}) \triangleq p(x)f(y|x, \theta^{\star})$, because we do not think the existence of mismatched label data situation in the training data.

\item \emph{(Test joint):} Let $q(x)$ denote the test marginal without dependence on the parameter. The unseen test pairs are distributed according to the test/true joint distribution of the form $q(x,y|\theta^{\star}) \triangleq q(x)f(y|x, \theta^{\star})$, because we do not think the existence of mismatched label data situation in the test task. 

\item \emph{(Differentiability):} The log-conditional  $\log\, f(y|x,\theta)$ is of class $\mc{C}^{3}(\Omega)$ for all $(x,y) \in X\times\{1,2, \cdots, \, c\}$, when being viewed as a function of the parameter.\footnote{We say that a function $f:X\xrightarrow[]{}Y$ is of $\mc{C}^{p}(X)$, for an integer $p>0$, if its derivatives up to $p$-th order exist and are continuous at all points of $X$.}
\item The parameter space $\Omega$ is compact, and there exists an open ball around the true parameter of the model $\theta^{\star}\in\Omega$.
\item \emph{(Invertibility):} The arbitrary Fisher information matrix $I_{\mc{P}(x)}(\theta)$ is positive definite and therefore invertible for all $\theta \in \Omega$.
\item $\{(x,y)\in supp(q(x,y|\theta^{\star}))\,\,|\,\,\nabla^{2}_{\theta}\log\,q(x,y|\theta^{\star})$ is singular$\}$ is a measure zero set. 
\end{enumerate}
\end{assum}

In contrast to~\cite{sourati2016asymptotic}, we modify (A3) so that the oracle model always outputs a hard label, \emph{i.e.,} $f(y_{true}(x)|x, \theta^\star) = 1$ and add (A9) which is not numbered but noted in the statement of Theorem 3 and Theorem 11 in~\cite{sourati2016asymptotic}.

\subsection{Preliminaries}\label{appsubsec:preliminaries}
We organize the two types of background knowledge, maximum likelihood estimator (MLE) and Fisher information (FI), needed for future analysis.


\subsubsection{Maximum Likelihood Estimator (MLE)}
In this section, we derive the maximum likelihood estimator in a classification problem with sampling probability $h(x)$. 
Unless otherwise specified, training set $\mc{D}_{n}=\{(x_i, y_i)\}_{i=1}^{n}$ is sampled from $p(x,y|\theta^\star)$. 
For given probability mass function (PMF) $h(x)$ on $\mc{D}_n$, we define MLE ${\hat{\theta}}_{h(x),\mc{D}_{n}}$ as follows:
\begin{align}
{\hat{\theta}}_{h(x),\mc{D}_{n}}   &\triangleq\argmax_{\theta} \quad \log \bb{P}(\mc{D}_{n}|\theta;h(x))\notag\\ 
                                &=\argmin_{\theta} \quad -\sum_{i=1}^{n}\,h(x_i)\log p(x_i, y_i|\theta)\label{appeq:mle_step1}\\
                                &=\argmin_{\theta} \quad -\sum_{i=1}^{n}\,h(x_i)\log f(y_i|x_i, \theta)\label{appeq:mle_step2}\\
                                &=\argmin_{\theta} \quad \sum_{i=1}^{n}\,h(x_i)\,\mc{L}_{\text{CE}}(x_i,y_i;\theta). \label{appeq:mle_step3}
\end{align}

In \eqref{appeq:mle_step1} and \eqref{appeq:mle_step2}, (A0) and (A4) of Assumption~\ref{app:main_assumption} in Appendix~\ref{appsec:backgrounds} were used, respectively. It is worth noting that MLE $\hat{\theta}_{h(x),\mc{D}_{n}}$ is a variable that is influenced by two factors: (1) a change in the training dataset $\mc{D}_{n}$ and (2) the adjustment of the sampling probability $h(x)$. If $h(x)$ is a uniform distribution $U(x)$, then the result of empirical risk minimization (ERM) is $\hat{\theta}_{U(x),\mc{D}_{n}}$.

\subsubsection{Fisher information (FI)}
\myparagraph{General definition of FI.} 
Fisher information (FI), denoted by $I_{\mc{P}(x)}(\theta)$, is a measure of sample information from a given distribution $\mc{P}(x,y|\theta)\triangleq\mc{P}(x)f(y|x,\theta)$. It is defined as the expected value of the outer product of the score function $\nabla_{\theta}\,\log\, \mc{P}(x,y|\theta)$ with itself, evaluated at some $\theta \in \Omega$. 

\begin{equation}\label{appeq:fisher_information}
I_{\mc{P}(x)}(\theta)\triangleq\bb{E}_{(x,y)\sim \mc{P}(x,y|\theta)}[\nabla_{\theta}\,\log\, \mc{P}(x,y|\theta){{\nabla^\mathsf{T}_{\theta}}}\,\log\, \mc{P}(x,y|\theta)].
\end{equation}

\myparagraph{Extended version of FI.}
Here, we summarize the extended version of FI, which can be derived by making some assumptions. These variants of FI are utilized in the proof of Theorems.

\begin{itemize}
\item \emph{(Hessian version)} Under the differentiability condition (A6) of Assumption \ref{app:main_assumption} in Appendix \ref{appsec:backgrounds}, 
FI can be written in terms of the Hessian matrix of the log-likelihood:
\begin{equation}
\label{appeq:fisher_hessian}
I_{\mc{P}(x)}(\theta) =- \bb{E}_{(x,y)\sim\mc{P}(x,y|\theta)}[{{\nabla^{2}_{\theta}}}\,\log\, \mc{P}(x,y|\theta)].
\end{equation}

\item \emph{(Model decomposition version)} 
Under the factorization condition (A0) of Assumption \ref{app:main_assumption} in Appendix \ref{appsec:backgrounds}, 
\eqref{appeq:fisher_information} and \eqref{appeq:fisher_hessian} can be transformed as follows:
\begin{align}
\label{appeq:varied_fisher_information_1}
I_{\mc{P}(x)}(\theta)&=\bb{E}_{(x,y)\sim \mc{P}(x)f(y|x,\theta)}[\nabla_{\theta}\,\log\, f(y|x,\theta){{\nabla^\mathsf{T}_{\theta}}}\,\log\, f(y|x,\theta)]\\
\label{appeq:varied_fisher_information_2}
&=- \bb{E}_{(x,y)\sim \mc{P}(x)f(y|x,\theta)}[{{\nabla^{2}_{\theta}}}\,\log\, f(y|x,\theta)].
\end{align}
Specifically, \eqref{appeq:varied_fisher_information_1} and \eqref{appeq:varied_fisher_information_2} can be unfolded as follows:
\begin{align}
\label{appeq:folded_fisher}
I_{\mc{P}(x)}(\theta) &= \int_{x\in X} \mc{P}(x)\sum_{y=1}^{c}\left[f(y|x, \theta)\cdot {{\nabla_{\theta}}}\,\log\,f(y|x,\theta){{\nabla^\mathsf{T}_{\theta}}}\,\log\,f(y|x,\theta)\right]dx \\
&= -\int_{x\in X} \mc{P}(x)\sum_{y=1}^{c}\left[f(y|x, \theta)\cdot {{\nabla^{2}_{\theta}}}\,\log\, f(y|x,\theta)\right]dx \nonumber
\end{align}
From now on, we define $I_{p(x)}(\theta)$ and $I_{q(x)}(\theta)$ as the FI derived from the training and test marginal, respectively.

\end{itemize}



\subsubsection{Empirical Fisher information (EFI)}\label{appsubsubsec:efi}

When the training dataset $\mc{D}_{n}$ is given, we denote the sampling probability as $h(x)$ which is defined on the probability space $\mc{H}$:
\begin{equation}
\label{appeq:h_space}
\mc{H} = \{h(x) \,| \textstyle\sum_{(x_i,y_i) \in \mc{D}_{n}} h(x_i) = 1 \,,\, h(x_i) \ge 0 \quad \forall (x_i,y_i) \in \mc{D}_{n}\}. \footnote{Note that for simplicity, we abuse the notation $h(x,y)$  used in Section~\ref{sec:method} as $h(x)$. This is exactly the same for a given dataset $\mc{D}_{n}$ situation.}
\end{equation}
Practically, the training dataset $\mc{D}_{n}$ is given as deterministic. Therefore,~\eqref{appeq:varied_fisher_information_1} can be refined as empirical Fisher information (EFI). This reformulation is frequently utilized, \emph{e.g.,} in ~\cite{jastrzkebski2017three, chaudhari2019entropy}, to reduce the computational complexity of gathering gradients for all possible classes (\emph{i.e.,} expectation with respect to $f(y|x,\theta)$ as in~\eqref{appeq:varied_fisher_information_1}). Refer the $\sum_{y=1}^{c}$ term of \eqref{appeq:folded_fisher}. Different from prior EFI, which is defined on the case when $h(x)$ is uniform, $U(x)$, we generalize the definition of EFI in terms of $h(x) \in \mc{H}$ as follows:
\begin{align}
    \hat{I}_{h(x)}(\theta) 
    &:= \bb{E}_{h(x)} \left[ \nabla_{\theta}\log\,f(y|x,\theta)\nabla^\top_{\theta}\log\,f(y|x,\theta) \right] \nonumber \\
    &\overset{(a)}{:=} 
    \textstyle\sum_{(x_i,y_i)\in\mc{D}_{n}} h(x_i)  \nabla_{\theta}\log\,f(y_i|x_i,\theta)\nabla^\top_{\theta}\log\,f(y_i|x_i,\theta).
\end{align}
Note that (a) holds owing to~\eqref{appeq:h_space}.

\subsubsection{Stochastic Order Notations $o_{\text{p}}$ and $O_{\text{p}}$}\label{appsubsubsec:stochastic_order_notations}
For a set of random variables $X_n$ and a corresponding set of constant $a_n$, the notation $X_n=o_{\text{p}}(a_n)$
means that the set of values ${X_n}/{a_n}$ converges to zero in probability as $n$ approaches an appropriate limit. It is equivalent with ${X_n}/{a_n}=o_{\text{p}}(1)$, where $X_n=o_{\text{p}}(1)$ is defined as:
\begin{equation*}
    \lim_{n\rightarrow\infty}\bb{P}(|X_n|\geq\epsilon)=0 \quad\forall\,\epsilon\geq 0.
\end{equation*}

The notation $X_n=O_{\text{p}}(a_n)$ means that the set of values ${X_n}/{a_n}$ is stochastically bounded. That is $\forall\,\epsilon>0, \exists\,\text{finite}\,M>0,\,N>0$ s.t. $\bb{P}(|{X_n}/{a_n}|>M)<\epsilon\,\,\text{for any}\,n>N$.

%% file: appendix/thm1.tex
\newpage
\section{Theorem 1}\label{appsec:theorem_1}
In this section, we deal with some required sub-lemmas that are used for the proof of Lemma~\ref{applem:lemma7}, which is the main ingredient of the proof of Theorem~\ref{appthm:theorem_1}.

\subsection{Sub-Lemmas}\label{appsubsec:sub_lemmas}

\begin{lem}[\cite{LehmCase98}, Theorem 5.1]\label{applem:lemma-1} When $\xrightarrow[]{P}$ denotes convergence in probability, and if (A0) to (A7) of the Assumption \ref{app:main_assumption} in Appendix \ref{appsec:backgrounds} hold, then there exists a sequence of MLE solutions $\{{\hat{\theta}}_{U(x),\mc{D}_{n}}\}_{n\in\bb{N}}$ that ${\hat{\theta}}_{U(x),\mc{D}_{n}}\xrightarrow[]{P}\theta^{\star}$ as ${n}\xrightarrow[]{}\infty$, where $\theta^{\star}$ is the `true' unknown parameter of the distribution of the sample.
\end{lem}

\begin{proof}
We refer to~\cite{LehmCase98} for detailed proof.
\end{proof}

\begin{lem}[\cite{LehmCase98}, Theorem 5.1]\label{applem:lemma-2} Let $\{{\hat{\theta}}_{U(x),\mc{D}_{n}}\}_{n\in\bb{N}}$ be the MLE based on the training data set $\mc{D}_{n}$. If (A0) to (A8) of the Assumption \ref{app:main_assumption} in Appendix \ref{appsec:backgrounds} hold, then the MLE ${\hat{\theta}}_{U(x),\mc{D}_{n}}$ has a zero-mean normal asymptotic distribution with a covariance equal to the inverse Fisher
information matrix, and with the convergence rate of 1/2:
\begin{equation*}
\sqrt{n}({\hat{\theta}}_{U(x),\mc{D}_{n}}-\theta^{\star})\xrightarrow[]{D}\mathcal{N}(\vec{0}, {I_{p(x)}(\theta^{\star})}^{-1}),
\end{equation*}
where $\xrightarrow[]{D}$ represents convergence in distribution.
\end{lem}

\begin{proof}
We refer to~\cite{LehmCase98} for detailed proof, based on Lemma~\ref{applem:lemma-1}.
\end{proof}

\begin{lem}[\cite{Wassermann04}, Theorem 9.18]
\label{applem:lemma1}
    Under the (A0) to (A8) of the Assumption \ref{app:main_assumption} in Appendix \ref{appsec:backgrounds} hold, we get
    \begin{equation*}
        \sqrt{n}{I_{p(x)}(\hat{\theta}_{U(x),\mc{D}_{n}})}^{1/2}(\hat{\theta}_{U(x),\mc{D}_{n}}-\theta^{\star})\xrightarrow[]{D}\mathcal{N}(\vec{0}, \bb{I}_{d}),
    \end{equation*}
    where $\xrightarrow[]{D}$ represents convergence in distribution.
\end{lem}

\begin{proof}
We refer to~\cite{Wassermann04} for detailed proof, based on Lemma~\ref{applem:lemma-2}.
\end{proof}



\begin{lem}(\cite{Serf:appr:1980}, Chapter 1)
\label{applem:lemma3}
Let $\{\theta_{n}\}$ be a sequence of random vectors. If there exists a random vector $\tilde{\theta}$ such that $\theta_{n}\xrightarrow[]{D}\tilde{\theta}$, then $\|\theta_{n}-\tilde{\theta}\|_{2}=O_{\text{p}}(1)$, where $\|\cdot\|_2$ denote the $L_2$ norm.
\end{lem}

\begin{proof}
We refer to~\cite{Serf:appr:1980} for detailed proof.
\end{proof}

\begin{lem}(\cite{sourati2016asymptotic}, Theorem 27)\label{applem:lemma4}
Let $\{\theta_{n}\}$ be a sequence of random vectors in a convex and compact set $\Omega\subseteq\bb{R}^{d}$ and $\theta^{\star}\in\Omega$ be a constant vector such that $\|\theta_{n}-\theta^{\star}\|_{2}=O_{\text{p}}(a_{n})$ where $a_n\xrightarrow[]{}0$ (as $n\xrightarrow[]{}\infty$). If $g:\Omega\xrightarrow[]{}\bb{R}$ is a $\mc{C}^{3}$ function, then
\begin{equation*}
    g(\theta_{n})=g(\theta^{\star})+\nabla^\mathsf{T}_{\theta}g(\theta^{\star})(\theta_{n}-\theta^{\star})+\frac{1}{2}(\theta_{n}-\theta^{\star})^\mathsf{T}\nabla^{2}_{\theta}g(\theta^{\star})(\theta_{n}-\theta^{\star})+o_{\text{p}}({a_{n}}^{2}).
\end{equation*}
\end{lem}

\begin{proof}
We refer to~\cite{Serf:appr:1980} for detailed proof.
\end{proof}

\begin{lem}\label{applem:lemma5}
If (A0) and (A3) of the Assumption \ref{app:main_assumption} in Appendix \ref{appsec:backgrounds} hold, then $\nabla_{\theta}\log\,\mc{P}(x,y_{true}(x)|\theta^{\star})=\vec{0}$ for any joint distribution $\mc{P}(x,y|\theta^\star)$. 
\end{lem}
\begin{proof}
\begin{align*}
\nabla_{\theta}\log\,\mc{P}(x,y_{true}(x)|\theta^{\star})&=\nabla_{\theta}\log\,f(y_{true}(x)|x,\theta^{\star})\\&=\nabla_{\theta}\log\,1\\&=\vec{0}.
\end{align*}

On the first equality, (A0) of Assumption \ref{app:main_assumption} in Appendix \ref{appsec:backgrounds} is used. At the second equality, (A3) of Assumption \ref{app:main_assumption} in Appendix \ref{appsec:backgrounds} is used.

\end{proof}

\begin{lem}\label{applem:lemma6}
If (A0) to (A8) of the Assumption \ref{app:main_assumption} in Appendix \ref{appsec:backgrounds} hold and the case $\nabla^{2}_{\theta}\log\,p(x,y_{true}(x)|\theta^{\star})$ is non-singular for given data $(x,y_{true}(x))$ satisfies, then the asymptotic distribution of the log-likelihood ratio is a mixture of first-order Chi-square distributions, and the convergence rate is one. More specifically:
\begin{equation}
    n\left(\log\,\frac{p(x,y_{true}(x)|\theta^{\star})}{p(x,y_{true}(x)|{\hat{\theta}}_{U(x),\mc{D}_{n}})}\right)\xrightarrow[]{D}\frac{1}{2}\sum_{i=1}^{d}\lambda_{i}(x,y_{true}(x))\cdot\chi^{2}_{1},
\end{equation}
where $\{\lambda_{i}(x,y_{true}(x))\}_{i=1}^{d}$ are eigenvalues of $I_{p(x)}(\theta^{\star})^{-\frac{1}{2}}\left\{-\nabla^{2}_{\theta}\log\,p(x,y_{true}(x)|\theta^{\star})\right\}I_{p(x)}(\theta^{\star})^{-\frac{1}{2}}$.
\end{lem}
\begin{proof}
The proof is based on the Taylor expansion theorem.  
Remind that we deal with the data $(x,y_{true}(x))$ satisfying  ${\nabla^{2}_{\theta}}\log\,p(x,y_{true}(x)|\theta^{\star})$ is non-singular.
From the property $\sqrt{n}({\hat{\theta}}_{U(x),\mc{D}_{n}}-\theta^{\star})\xrightarrow[]{D}\mathcal{N}(\vec{0}, {I_{p(x)}(\theta^{\star})}^{-1})$ derived from Lemma \ref{applem:lemma1}, one concludes that $\sqrt{n}\|{\hat{\theta}}_{U(x),\mc{D}_{n}}-\theta^{\star}\|_{2}=O_{\text{p}}(1)$ and therefore $\|{\hat{\theta}}_{U(x),\mc{D}_{n}}-\theta^{\star}\|_{2}=O_{\text{p}}(\frac{1}{\sqrt{n}})$ by the Lemma \ref{applem:lemma3}.

Thus, by the Lemma \ref{applem:lemma4},
\begin{align*}
    \log\,p(x,y_{true}(x)|&{\hat{\theta}}_{U(x),\mc{D}_{n}})=\log\,p(x,y_{true}(x)|\theta^{\star})+({\hat{\theta}}_{U(x),\mc{D}_{n}}-\theta^{\star})^{\mathsf{T}}\nabla_{\theta}\log\,p(x,y_{true}(x)|\theta^{\star})\\
    &+ \frac{1}{2}({\hat{\theta}}_{U(x),\mc{D}_{n}}-\theta^{\star})^{\mathsf{T}}{\nabla^{2}_{\theta}}\log\,p(x,y_{true}(x)|\theta^{\star})({\hat{\theta}}_{U(x),\mc{D}_{n}}-\theta^{\star})+o_{\text{p}}\left(\frac{1}{n}\right)
\end{align*}
holds. By the Lemma \ref{applem:lemma1} and the property $\nabla_{\theta}\log\,p(x,y_{true}(x)|\theta^{\star})=\vec{0}$ derived by Lemma \ref{applem:lemma5}, we can obtain
\begin{align*}
&n\left[\log\,p(x,y_{true}(x)|\theta^{\star})-\log\,p(x,y_{true}(x)|{\hat{\theta}}_{U(x),\mc{D}_{n}})\right]\\
&=-\frac{1}{2}\left[\sqrt{n}({\hat{\theta}}_{U(x),\mc{D}_{n}}-\theta^{\star})\right]^{\mathsf{T}}{\nabla^{2}_{\theta}}\log\,p(x,y_{true}(x)|\theta^{\star})\left[\sqrt{n}({\hat{\theta}}_{U(x),\mc{D}_{n}}-\theta^{\star})\right]+o_{\text{p}}(1)\\
&\xrightarrow[]{D}\frac{1}{2}{\mathcal{N}\left(\vec{0}, {I_{p(x)}(\theta^{\star})}^{-1}\right)}^{\mathsf{T}}\left[-\nabla^{2}_{\theta}\log\,p(x,y_{true}(x)|\theta^{\star})\right]\mathcal{N}\left(\vec{0}, {I_{p(x)}(\theta^{\star})}^{-1}\right)\\
&=\frac{1}{2}{\mathcal{N}\left(\vec{0}, \bb{I}_{d}\right)}^{\mathsf{T}}\left[-{I_{p(x)}(\theta^{\star})}^{-\frac{1}{2}}\nabla^{2}_{\theta}\log\,p(x,y_{true}(x)|\theta^{\star}){I_{p(x)}(\theta^{\star})}^{-\frac{1}{2}}\right]\mathcal{N}\left(\vec{0}, \bb{I}_{d}\right).
\end{align*}
    
Define $\Upgamma(x,y_{true}(x))$ as $-{I_{p(x)}(\theta^{\star})}^{-\frac{1}{2}}\nabla^{2}_{\theta}\log\,p(x,y_{true}(x)|\theta^{\star}){I_{p(x)}(\theta^{\star})}^{-\frac{1}{2}}$ and rewrite the right-hand-side element-wise\footnote{Suppose $\Upgamma=U\Sigma\,U^\mathsf{T}$ and $\Sigma=diag(\lambda_1,\cdots\lambda_{d})$. Then, $\mathcal{N}\left(\vec{0}, \bb{I}_{d}\right)^{\mathsf{T}}U\sim\mathcal{N}\left(\vec{0}, UU^{\mathsf{T}}\right)=\mathcal{N}\left(\vec{0}, \bb{I}_{d}\right)$.\\ Thus, $\mathcal{N}\left(\vec{0}, \bb{I}_{d}\right)^{\mathsf{T}}\Upgamma\mathcal{N}\left(\vec{0}, \bb{I}_{d}\right)=\mathcal{N}\left(\vec{0}, \bb{I}_{d}\right)^{\mathsf{T}}\Sigma\mathcal{N}\left(\vec{0}, \bb{I}_{d}\right)=\sum_{i=1}^{d}\lambda_{i}\mathcal{N}(0,1)^{2}$.} as
\begin{align*}
    \frac{1}{2}\mathcal{N}\left(\vec{0}, \bb{I}_{d}\right)^{\mathsf{T}}\Upgamma(x,y_{true}(x))&\mathcal{N}\left(\vec{0}, \bb{I}_{d}\right)\\
    &=\frac{1}{2}\sum_{i=1}^{d}\lambda_{i}(x,y_{true}(x))\cdot\mathcal{N}(0,1)^{2}=\frac{1}{2}\sum_{i=1}^{d}\lambda_{i}(x,y_{true}(x))\cdot{\chi_{1}}^{2},
\end{align*}
where $\{\lambda_{i}(x,y_{true}(x))\}_{i=1}^{d}$ are eigenvalues of $\Upgamma(x,y_{true}(x))$. Thus, the desired property is obtained.
\end{proof}


\subsection{Main Lemma}\label{appsubsec:main_lemma}
In this section, we derive the main Lemma, which represents the test cross-entropy loss and can be understood as Fisher information ratio (FIR)~\cite{sourati2016asymptotic}.

\subsubsection{Main Lemma statement and proof}\label{appsubsubsec:main_lemma_statement}
\begin{lem}[FIR in expected test cross entropy loss with MLE]\label{applem:lemma7}
If the Assumption \ref{app:main_assumption} in Appendix \ref{appsec:backgrounds} holds, then
    \begin{equation}\label{appeq:fir_in_debiasing}
        \begin{split}
            \lim_{n\to\infty} n\bb{E}_{(x,y)\sim q(x)f(y|x,\theta^\star)}
            &\left[\bb{E}_{\mc{D}_{n} \sim p(x)f(y|x,\theta^\star)}\left[-\log{f(y|x,{\hat{\theta}}_{U(x),\mc{D}_{n}})}\right]\right]\\ 
            &=\frac{1}{2}\trace\left[I_{p(x)}(\theta^{\star})^{-1}I_{q(x)}(\theta^{\star})\right].
        \end{split}
    \end{equation}
\end{lem}

\begin{proof}
We prove Lemma~\ref{applem:lemma7} via two steps. First we show that the expected cross-entropy loss term can be rewritten in terms of the log-likelihood ratio. Then, we prove that the expected log-likelihood ratio can be asymptotically understood as FIR.

\myparagraph{\underbar{Step 1: Log-likelihood ratio}} We show that the expected log-likelihood ratio can be formulated as the expected test cross-entropy loss as follows:


This property holds because,
\begin{align}
\bb{E}&_{(x,y)\sim q(x)f(y|x,\theta^\star)}\left[\bb{E}_{\mc{D}_{n}\sim p(x)f(y|x,\theta^\star)}\left[\log\frac{p(x,y|\theta^{\star})}{p(x,y|{\hat{\theta}}_{U(x),\mc{D}_{n}})}\right]\right] \nonumber \\
&=\bb{E}_{(x,y)\sim q(x)f(y|x,\theta^\star)}
\left[\bb{E}_{\mc{D}_{n}\sim p(x)f(y|x,\theta^\star)}\left[\log\frac{f(y|x,\theta^{\star})}{f(y|x,{\hat{\theta}}_{U(x),\mc{D}_{n}})}\right]\right]\label{appeq:ideal_debias_step1}\\
&=\bb{E}_{(x,y)\sim q(x)f(y|x,\theta^\star)}
\left[\bb{E}_{\mc{D}_{n}\sim p(x)f(y|x,\theta^\star)}\left[\log\frac{f(y|x,\theta^{\star})}{f(y|x,{\hat{\theta}}_{U(x),\mc{D}_{n}})}\right]\bb{1}_{Supp(q(x,y|\theta^\star))}\right]\label{appeq:ideal_debias_step2}\\
&=\bb{E}_{(x,y)\sim q(x)f(y|x,\theta^\star)}
\left[\bb{E}_{\mc{D}_{n}\sim p(x)f(y|x,\theta^\star)}\left[\log\frac{f(y|x,\theta^{\star})}{f(y|x,{\hat{\theta}}_{U(x),\mc{D}_{n}})}\,\,\bb{1}_{Supp(q(x,y|\theta^\star))}\right]\right]\notag\\ 
&=\bb{E}_{(x,y)\sim q(x)f(y|x,\theta^\star)}
\left[\bb{E}_{\mc{D}_{n}\sim p(x)f(y|x,\theta^\star)}\left[-\log{f(y|x,{\hat{\theta}}_{U(x),\mc{D}_{n}})}\,\,\bb{1}_{Supp(q(x,y|\theta^\star))}\right]\right] \label{appeq:ideal_debias_step3}\\
&=\bb{E}_{(x,y)\sim q(x)f(y|x,\theta^\star)}
\left[\bb{E}_{\mc{D}_{n}\sim p(x)f(y|x,\theta^\star)}\left[-\log{f(y|x,{\hat{\theta}}_{U(x),\mc{D}_{n}})}\right]\right]\notag
\end{align}

Since (A0) of Assumption \ref{app:main_assumption} in Appendix \ref{appsec:backgrounds}, \eqref{appeq:ideal_debias_step1} and \eqref{appeq:ideal_debias_step2} hold. At \eqref{appeq:ideal_debias_step3}, the properties, (i) $Supp(q(x,y|\theta^\star))\subseteq \mc{W}$ and (ii) $f(y|x,\theta^{\star})=1$ $\forall\,(x,y)\in\mc{W}$, was used which is derived by (A3) of the Assumption \ref{app:main_assumption} in Appendix \ref{appsec:backgrounds}.

\myparagraph{\underbar{Step 2: FIR}} Here, we show that the expected test loss of MLE can be understood as FIR.

By (A0) in Assumption~\ref{app:main_assumption} in Appendix~\ref{appsec:backgrounds}, trivially 
\begin{equation}\label{appeq:lemma7_used_prop1}
    \begin{split}
        \{(x,y)\in &Supp(q(x,y|\theta^\star))\,\,|
        \,\,\nabla^{2}_{\theta}\log\,q(x,y|\theta^{\star})\,\text{is singular}\}\\
        &=\{(x,y)\in Supp(q(x,y|\theta^\star))\,\,|\,\,\nabla^{2}_{\theta}\log\,p(x,y|\theta^{\star})\,\text{is singular}\}.
    \end{split}
\end{equation}
holds.
Since~\eqref{appeq:lemma7_used_prop1}, and (A9) in Assumption~\ref{app:main_assumption} in Appendix~\ref{appsec:backgrounds}, 
$Supp(q(x,y|\theta^\star))$ can be replaced by,
\begin{equation}\label{appeq:lemma7_used_prop2}
    S\triangleq Supp(q(x,y|\theta^\star))\setminus\{(x,y)\in \mc{W}\,\,|\,\, \nabla^{2}_{\theta}\,\log\,p(x,y|\theta^{\star})\,\,\text{is singular}\}     
\end{equation}
when calculate expectation.

We can get a result of Lemma~\ref{applem:lemma7} as follows:
\begin{align}
&\lim_{n\to\infty} n\bb{E}_{(x,y)\sim q(x)f(y|x,\theta^\star)}\left[\bb{E}_{\mc{D}_{n} \sim p(x)f(y|x,\theta^\star)}\left[-\log{f(y|x,{\hat{\theta}}_{U(x),\mc{D}_{n}})}\right]\right] \nonumber \\ 
&=\lim_{n\to\infty} n\bb{E}_{(x,y)\sim q(x)f(y|x,\theta^{\star})}
\left[\bb{E}_{\mc{D}_{n}\sim p(x)f(y|x,\theta^{\star})}\left[\log\frac{p(x,y|\theta^{\star})}{p(x,y|{\hat{\theta}}_{U(x),\mc{D}_{n}})}\right]\bm{1}_{Supp(q(x,y|\theta^\star))}\right]\label{appeq:main_lemma_step1}\\
&=\lim_{n\to\infty} n\bb{E}_{(x,y)\sim q(x)f(y|x,\theta^{\star})}
\left[\bb{E}_{\mc{D}_{n}\sim p(x)f(y|x,\theta^{\star})}\left[\log\frac{p(x,y|\theta^{\star})}{p(x,y|{\hat{\theta}}_{U(x),\mc{D}_{n}})}\right]\bm{1}_{S}\right]\label{appeq:main_lemma_step2}\\
&=\bb{E}_{(x,y)\sim q(x)f(y|x,\theta^{\star})}
\left[\lim_{n\to\infty}\bb{E}_{\mc{D}_{n}\sim p(x)f(y|x,\theta^{\star})}\left[n\log\frac{p(x,y|\theta^{\star})}{p(x,y|{\hat{\theta}}_{U(x),\mc{D}_{n}})}\right]\bm{1}_{S}\right]\notag
\end{align}
\begin{align}
&=\bb{E}_{(x,y_{true}(x))\sim q(x)f(y|x,\theta^{\star})}
\left[\lim_{n\to\infty}\bb{E}_{\mc{D}_{n}\sim p(x)f(y|x,\theta^{\star})}\left[n\log\frac{p(x,y_{true}(x)|\theta^{\star})}{p(x,y_{true}(x)|{\hat{\theta}}_{U(x),\mc{D}_{n}})}\right]\bm{1}_{S}\right]\label{appeq:main_lemma_step3}\\
&=
\bb{E}_{(x,y_{true}(x))\sim q(x)f(y|x,\theta^{\star})}
\left[\left(\frac{1}{2}\sum_{i=1}^{d}\lambda_{i}(x,y_{true}(x))\bb{E}\left[\chi^{2}_{1}\right]
\right)\bm{1}_{S}\right]
\label{appeq:main_lemma_step4}\\
&=\bb{E}_{(x,y_{true}(x))\sim q(x)f(y|x,\theta^{\star})}
\left[\left(\frac{1}{2}\sum_{i=1}^{d}\lambda_{i}(x,y_{true}(x))\right)\bm{1}_{S}\right]\label{appeq:main_lemma_step5}\\
&=\bb{E}_{(x,y_{true}(x))\sim q(x)f(y|x,\theta^{\star})}
\left[\frac{1}{2}\trace\left[I_{p(x)}(\theta^{\star})^{-\frac{1}{2}}\left\{-\nabla^{2}_{\theta}\log\,p(x,y_{true}(x)|\theta^{\star})\right\}I_{p(x)}(\theta^{\star})^{-\frac{1}{2}}\right]\bm{1}_{S}\right]\label{appeq:main_lemma_step6}\\
&=\bb{E}_{(x,y)\sim q(x)f(y|x,\theta^{\star})}
\left[\frac{1}{2}\trace\left[I_{p(x)}(\theta^{\star})^{-\frac{1}{2}}\left\{-\nabla^{2}_{\theta}\log\,p(x,y|\theta^{\star})\right\}I_{p(x)}(\theta^{\star})^{-\frac{1}{2}}\right]\bm{1}_{S}\right]\label{appeq:main_lemma_step7}\\
&=\frac{1}{2}\trace\left[I_{p(x)}(\theta^{\star})^{-\frac{1}{2}} 
\bb{E}_{(x,y)\sim q(x)f(y|x,\theta^{\star})}
\left[-\nabla^{2}_{\theta}\log\,p(x,y|\theta^{\star})\,\,\bm{1}_{S}\right]I_{p(x)}(\theta^{\star})^{-\frac{1}{2}}\right]\notag\\
&=\frac{1}{2}\trace\left[I_{p(x)}(\theta^{\star})^{-\frac{1}{2}} 
\bb{E}_{(x,y)\sim q(x)f(y|x,\theta^{\star})}
\left[-\nabla^{2}_{\theta}\log\,q(x,y|\theta^{\star})\,\,\bm{1}_{S}\right]I_{p(x)}(\theta^{\star})^{-\frac{1}{2}}\right]\label{appeq:main_lemma_step8}\\
&=\frac{1}{2}\trace\left[I_{p(x)}(\theta^{\star})^{-\frac{1}{2}} 
\bb{E}_{(x,y)\sim q(x)f(y|x,\theta^{\star})}
\left[-\nabla^{2}_{\theta}\log\,q(x,y|\theta^{\star})\right]I_{p(x)}(\theta^{\star})^{-\frac{1}{2}}\right]\notag\\
&=\frac{1}{2}\trace\left[I_{p(x)}(\theta^{\star})^{-\frac{1}{2}} 
I_{q(x)}(\theta^{\star})I_{p(x)}(\theta^{\star})^{-\frac{1}{2}}\right]\notag\\
&=\frac{1}{2}\trace\left[I_{p(x)}(\theta^{\star})^{-1}I_{q(x)}(\theta^{\star})\right].\label{appeq:main_lemma_step9}
\end{align}

\eqref{appeq:main_lemma_step1} holds from Step 1. From~\eqref{appeq:lemma7_used_prop2},~\eqref{appeq:main_lemma_step2} satisfied. 
\eqref{appeq:main_lemma_step3} and~\eqref{appeq:main_lemma_step7} holds because $(x,y)$ is sampled from $q(x)f(y|x.\theta^\star)$. From \eqref{appeq:main_lemma_step4} to \eqref{appeq:main_lemma_step6}, the result of Lemma \ref{applem:lemma6} is used. \eqref{appeq:main_lemma_step8} can be obtained thanks to (A0).
Lastly, \eqref{appeq:main_lemma_step9} holds because trace satisfies the commutative law about matrix multiplication. 

The final term, $\trace\left[I_{p(x)}(\theta^{\star})^{-1}I_{q(x)}(\theta^{\star})\right]$, is known as the Fisher information ratio (FIR) because it can be expressed as a ratio in the scalar case.

\end{proof}

\subsection{Theorem 1}
\label{appsubsec:theorem_1}
In this section, we finally prove Theorem~\ref{appthm:theorem_1}. To do so, we additionally follow assumptions in~\cite{sourati2016asymptotic}.
\subsubsection{Additional Assumption}\label{appsubsubsec:additional_assumption}

\begin{assum}\label{app:additional_assumption}
We assume to exist four positive constants $L_1, L_2, L_3, L_4\geq0$ such that following properties hold  $\forall x \in X, y \in\{1,\cdots,c\}$ and $\theta\in\Omega:$
\begin{itemize}
\item $ I(\theta,x)=-\nabla^{2}_{\theta}\log\,f(y|x,\theta)\,\,\text{is independent of the class labels}\,\,y.$
\item ${\nabla_{\theta}\log\,f(y|x,\theta^{\star})}^{\mathsf{T}}{I_{q(x)}(\theta^{\star})}^{-1}{\nabla_{\theta}\log\,f(y|x,\theta^{\star})}\leq L_{1}$
\item $\norm{I_{q(x)}(\theta^{\star})^{-1/2}I(\theta^{\star},x)I_{q(x)}(\theta^{\star})^{-1/2}}\leq L_{2} $
\item $\norm{I_{q(x)}(\theta^{\star})^{-1/2}(I(\theta^{'},x)-I(\theta^{''},x))I_{q(x)}(\theta^{\star})^{-1/2}}\leq L_{3}(\theta^{'}-\theta^{''})^{\mathsf{T}}I_{q(x)}(\theta^{\star})(\theta^{'}-\theta^{''})$
\item $-L_{4}\|\theta-\theta^{\star}\|I(\theta^{\star},x)\preceq I(\theta,x)-I(\theta^{\star},x)\preceq L_{4}\|\theta-\theta^{\star}\|I(\theta^{\star},x)$
\end{itemize}
\end{assum}

\subsubsection{Replacing $\theta
^{\star}$ by ${\hat{\theta}}_{U(x),\mc{D}_{n}}$}\label{appsubsubsec:theorem_1}

\begin{lem}
\label{applem:lemma9}
Suppose Assumption \ref{app:main_assumption} in Appendix \ref{appsec:backgrounds} and Assumption \ref{app:additional_assumption} in Appendix \ref{appsec:theorem_1} hold, then with high probability:
\begin{equation}
\trace \left[I_{p(x)}(\theta^{\star})^{-1}\right] = \lim_{n\rightarrow\infty}\trace\left[I_{p(x)}(\hat{\theta}_{U(x),\mc{D}_n})^{-1}\right].    
\end{equation}
\end{lem}

\begin{proof}

It is shown in the proof of Lemma 2 in \cite{chaudhuri2015convergence} that under the assumptions mentioned in  Assumption \ref{app:additional_assumption}, the following inequalities hold with probability $1-\delta(n)$:
\begin{equation}\label{eq:approximation_reference}
    \frac{\beta(n)-1}{\beta(n)}I(\theta^{\star},x)\preceq I(\hat{\theta}_{U(x),\mc{D}_n},x)\preceq \frac{\beta(n)+1}{\beta(n)}I(\theta^{\star},x),
\end{equation}
where $\beta(n)$ and $1-\delta(n)$ are proportional to $n$, which is the size of the training set $\mc{D}_{n} $.

Because of the independence for the class labels $y$ of $I(\theta,x)$, $\mc{P}(x)$, $I_{\mc{P}(x)}(\theta)=\bb{E}_{x \sim  \mc{P}(x)}\left[I(\theta, x)\right]$ holds for any marginal distribution $\mc{P}(x)$.\footnote{$I_{\mc{P}(x)}(\theta)=\bb{E}_{(x,y)\sim\mc{P}(x,y|\theta)}\left[-\nabla^{2}_{\theta}\log\,f(y|x,\theta)\right]=\bb{E}_{x\sim\mc{P}(x)}\left[\bb{E}_{y\sim f(y|x, \theta)}\left[-\nabla^{2}_{\theta}\log\,f(y|x,\theta)\right]\right]=\bb{E}_{x\sim \mc{P}(x)}\left[\bb{E}_{y\sim f(y|x, \theta)}\left[I(\theta,x)\right]\right]=\bb{E}_{x\sim \mc{P}(x)}\left[I(\theta,x)\right]$.}
 
 Taking the expectation to the \eqref{eq:approximation_reference} with respect to the marginal $p(x)$ and $q(x)$, then:
 \begin{equation}\label{appeq:assumption_derived_result}
    \frac{\beta(n)-1}{\beta(n)}I_{p(x)}(\theta^{\star})\preceq I_{p(x)}(\hat{\theta}_{U(x),\mc{D}_n})\preceq \frac{\beta(n)+1}{\beta(n)}I_{p(x)}(\theta^{\star}).     
 \end{equation}

 \begin{equation*}
    \frac{\beta(n)-1}{\beta(n)}I_{q(x)}(\theta^{\star})\preceq I_{q(x)}(\hat{\theta}_{U(x),\mc{D}_n})\preceq \frac{\beta(n)+1}{\beta(n)}I_{q(x)}(\theta^{\star}).     
 \end{equation*}
 
 Since $I_{p(x)}(\theta^{\star})$ and $I_{p(x)}(\hat{\theta}_{U(x),\mc{D}_n})$ are assumed to be positive definite, we can write \eqref{appeq:assumption_derived_result} in terms of inverted matrices\footnote{For $\forall$ two positive definite matrices $A$ and $B$, we have that $A\succeq B\Rightarrow A^{-1}\preceq B^{-1}$}:
 \begin{equation}\label{appeq:fisher_asymptotic}
    \frac{\beta(n)}{\beta(n)+1}{{I_{p(x)}}(\theta^{\star})}^{-1}\preceq {{I_{p(x)}}(\hat{\theta}_{U(x),\mc{D}_n})}^{-1}\preceq \frac{\beta(n)}{\beta(n)-1}{I_{p(x)}(\theta^{\star})}^{-1}.     
 \end{equation}
 
 \eqref{appeq:fisher_asymptotic} is equivalent to
 \begin{equation}\label{appeq:fisher_asymptotic_final}
    \frac{\beta(n)-1}{\beta(n)}{{I_{p(x)}}(\hat{\theta}_{U(x),\mc{D}_n})}^{-1}\preceq {{I_{p(x)}}(\theta^\star)}^{-1}\preceq \frac{\beta(n)+1}{\beta(n)}{{I_{p(x)}}(\hat{\theta}_{U(x),\mc{D}_n})}^{-1}.    
 \end{equation}
 From \eqref{appeq:fisher_asymptotic_final},
 \begin{equation}\label{appeq:trace_fisher_asymptotic}
 \frac{\beta(n)-1}{\beta(n)}\trace\left[{{I_{p(x)}}(\hat{\theta}_{U(x),\mc{D}_n})}^{-1}\right]\leq\trace\left[{{I_{p(x)}}(\theta^\star)}^{-1}\right]\leq\frac{\beta(n)+1}{\beta(n)}\trace\left[{{I_{p(x)}}(\hat{\theta}_{U(x),\mc{D}_n})}^{-1}\right]    
 \end{equation}
 satisfies.\footnote{If $A \preceq B$, $\trace\left[A\right]\leq\trace\left[B\right]$ holds.\\
 ($\because$)$\,A \preceq B\Rightarrow B-A\succeq\mc{O}$ and $B-A:=U\Sigma U^{T}$, where $U=[u_1|\cdots,|u_d].$\\
 Then, $\trace(B-A)=\sum_{i=1}^{d}u_{i}(B-A)u_{i}^{T}\leq 0$ because of the positive definite property of $B-A$.\\
$\trace(B-A)\leq0\Rightarrow\trace(B)\leq\trace(A).$
 }
 
Thus,
\begin{equation}\label{appeq:asymptotic_trace}
\lim_{n\to\infty}\trace\left[{{I_{p(x)}}(\hat{\theta}_{U(x),\mc{D}_n})}^{-1}\right]=\trace\left[{{I_{p(x)}}(\theta^\star)}^{-1}\right]    
\end{equation}
holds when taking $n\rightarrow\infty$ to the \eqref{appeq:trace_fisher_asymptotic}. Note that $\beta(n)$ is proportional to $n$.

\end{proof}

\subsection{Statement and proof of Theorem~\ref{appthm:theorem_1}}



\begin{thm}
\label{appthm:theorem_1}
Suppose Assumption~\ref{app:main_assumption} 
 in Appendix~\ref{appsec:backgrounds}
and Assumption \ref{app:additional_assumption} in Appendix \ref{appsec:theorem_1} hold, then for sufficiently large $n=|\mc{D}_{n}|$, the following holds with high probability:
\begin{equation}
\begin{split}
\bb{E}_{(x,y)\sim q(x)f(y|x,\theta^\star)}&\left[\bb{E}_{\mc{D}_{n}\sim p(x)f(y|x,\theta^\star)}\left[-\log{f(y|x,\hat{\theta}_{U(x),\mc{D}_n})}\right]\right] \cr &\leq\frac{1}{2n}\trace\left[I_{p(x)}(\hat{\theta}_{U(x),\mc{D}_n})^{-1}\right]\trace\left[I_{q(x)}(\theta^{\star})\right]. 
\end{split}\label{appeq:theorem_1}
\end{equation}
\end{thm}

\begin{proof}
Because of the (A8) of Assumption \ref{app:main_assumption} in Appendix \ref{appsec:backgrounds}, $I_{p(x)}(\theta^{\star})^{-1}$ and $I_{q(x)}(\theta^{\star})$ are positive definite matrix. Thus, $\trace\left[I_{p(x)}(\theta^{\star})^{-1}I_{q(x)}(\theta^{\star})\right]\leq\trace\left[I_{p(x)}(\theta^{\star})^{-1}\right]\trace\left[I_{q(x)}(\theta^{\star})\right]$ holds.\footnote{For $\forall$ two positive definite matrices $A$ and $B$, $\trace\left[AB\right]\leq\trace\left[A\right]\trace\left[B\right]$ satisfies.}

From the result of Lemma~\ref{applem:lemma7} and~\ref{applem:lemma9} in Appendix \ref{appsec:theorem_1}, 
\begin{align*}
\lim_{n\to\infty} n\bb{E}_{(x,y)\sim q(x)f(y|x,\theta^\star)}&\left[\bb{E}_{\mc{D}_{n}\sim p(x)f(y|x,\theta^\star)}\left[-\log{f(y|x,\hat{\theta}_{U(x),\mc{D}_n})}\right]\right]\\&=\frac{1}{2}\trace\left[I_{p(x)}(\theta^\star)^{-1}I_{q(x)}(\theta^{\star})\right]\\
&\leq\frac{1}{2}\trace\left[I_{p(x)}(\theta^\star)^{-1}\right]\trace\left[I_{q(x)}(\theta^{\star})\right]\\
&=\frac{1}{2}\lim_{n\to\infty}\trace\left[I_{p(x)}(\hat{\theta}_{U(x),\mc{D}_n})^{-1}\right]\trace\left[I_{q(x)}(\theta^{\star})\right]
\end{align*}
holds with high probability.
\end{proof}

It is worth noting that Theorem~\ref{appthm:theorem_1} states that the upper bound of the MLE ${\hat{\theta}}_{U(x), \mc{D}_{n}}$, $\mc{D}_{n}$ test loss can be minimized by lowering $\trace\left[I_{p(x)}(\hat{\theta}_{U(x),\mc{D}_n})^{-1}\right]$ when training marginal $p(x)$, the only tractable and controllable variable.

%% file: appendix/thm2.tex
\section{Theorem 2}
\label{appsec:theorem_2}

In this section, we introduce the motivation of gradient norm-based importance sampling. To show why this is important, we introduce the debiasing object problem for a given $\mc{D}_{n}$ under sampling probability $h(x)$ and show how to solve it in a toy example because the problem is difficult.

\subsection{Practical objective function for the dataset bias problem}
Remark that the right-hand side term of~\eqref{appeq:theorem_1} are controlled by the training and test marginals $p(x)$, and $q(x)$. Since we can only control the training dataset $\mc{D}_n$ not $p(x)$ and $q(x)$, we can design a practical objective function for the dataset bias problem by using EFI and Theorem~\ref{appthm:theorem_1} as follows:
\begin{equation}
    \label{appeq:ultimate_obj}
    \min_{h(x)\in\mc{H}} \trace \left[ \hat{I}_{h(x)}(\hat{\theta}_{h(x),\mc{D}_n})^{-1}\right],
\end{equation}
where $\hat{I}_{h(x)}(\theta)$ is an empirical Fisher information matrix. Remark that EFI is defined as:
\begin{equation}\label{appeq:empirical_fisher}
\hat{I}_{h(x)}(\theta) =     \sum_{i=1}^{n}h(x_i)\nabla_{\theta}\log\,f(y_i|x_i,\theta)\nabla^\top_{\theta}\log\,f(y_i|x_i,\theta).
\end{equation}
Here, $h(x)$ describes the sampling probability on $\mc{D}_n$, which is the only controllable term. We deal with \eqref{appeq:ultimate_obj} in the toy example because of the difficulty of the problem.

\subsection{One-dimensional Toy Example Setting}

For simplicity, we assume that $\mc{D}_{n}$ comprises sets $M$ and $m$, and the samples in each set share the same loss function and the same probability mass. The details are as follows:
\begin{itemize}
    \item For the given $a\in\bb{R}$, at the model parameter $\theta\in\bb{R}$, $\frac{1}{2}(\theta+a)^{2}$ and $\frac{1}{2}(\theta-a)^{2}$ loss function arise for all data in $M$ and $m$, respectively.
    \item ${\hat{\theta}}_{h,\mc{D}_{n}}$ denote the trained model from the arbitrary PMF $h(x)\in\mc{H}$ which has a constraint having degree of freedom 2, $(h_{M}(x), h_{m}(x))$.
    \item Concretely, each samples of $M$ and $m$ has a probability mass $h_M(x)$ and $h_m(x)$, respectively. \emph{i.e.,} $|M|\cdot h_M(x)+|m|\cdot h_m(x)=1$, where $|M|$ and $|m|$ denote the cardinality of $M$ and $m$, respectively. 
    \item Let $g_M(\theta)$ and $g_m(\theta)$ denote the gradient of each sample in $M$ and $m$ at $\theta \in \bb{R}$, respectively.
    \item Then, $|M|\cdot h_M(x)\cdot  g_M({\hat{\theta}}_{h(x),\mc{D}_{n}}) + |m|\cdot h_m(x) \cdot g_m({\hat{\theta}}_{h(x),\mc{D}_{n}}) = 0$ hold by the definition of ${\hat{\theta}}_{h(x),\mc{D}_{n}}$.
    \item In these settings, our objective can be written as finding $h^\star(x) = \argmin_{h(x)\in\mc{H}} \trace \left[ \hat{I}_h(x)(\hat{\theta}_{h(x),\mc{D}_n})^{-1}\right]$ and this is equivalent to find $(h^\star_M(x), h^\star_m(x))$.
\end{itemize}

\subsection{Statement and proof of Theorem~\ref{appthm:norm}}
In this section, we introduce the motivation for the gradient norm-based importance sampling in the toy example setting.

\begin{thm}\label{appthm:norm}
    Under the above setting, the solution of $(h^\star_M(x),h^\star_m(x)) = \argmin_{h(x)\in\mc{H}} \trace \left[ \hat{I}_{h(x)}(\hat{\theta}_{h(x),\mc{D}_n})^{-1} \right]$ is:
    \begin{equation*}
        h^{\star}_{M}(x) =\frac{|g_M({\hat{\theta}}_{U(x),\mc{D}_n})|}{Z}, \quad h^{\star}_{m}(x) =\frac{|g_m({\hat{\theta}}_{U(x),\mc{D}_n})|}{Z},
    \end{equation*}
    where $Z = |M| |g_M(\hat{\theta}_{U(x),\mc{D}_n})| +  |m| |g_m(\hat{\theta}_{U(x),\mc{D}_n})|$, and $|M|$ and $|m|$ denote the cardinality of $M$ and $m$, respectively.
\end{thm}

\begin{proof}
The trained model $\hat{\theta}_{h(x),\mc{D}_n}\in[-a,a]$ holds trivially for any $h(x)\in\mc{H}$. By the loss function definition in the toy setting, $g_{M}(\theta)=\theta+a$ and $g_{m}(\theta)=\theta-a$, $\forall$ $\theta$. Thus, $|g_{M}(\theta)|+|g_{m}(\theta)|=2a$ satisfies for $\forall$ $\theta\in[-a,a]$. Since the gradient is scalar in the toy setting, $\hat{I}_{h(x)}(\hat{\theta}_{h(x),\mc{D}_n})$ is also scalar and the same as the unique eigenvalue, that is,
\begin{equation*}
    \hat{I}_{h(x)}(\hat{\theta}_{h(x),\mc{D}_n})=|M|\cdot h_{M}(x)\cdot\{g_{M}(\hat{\theta}_{h(x),\mc{D}_n})\}^{2}+|m|\cdot h_{m}(x)\cdot\{g_{m}(\hat{\theta}_{h(x),\mc{D}_n})\}^{2} \quad \forall h(x)\in\mc{H}.
\end{equation*}
Thus, our problem is deciding $h_{M}(x)$ and $h_{m}(x)$ that maximize $|M|\cdot h_{M}(x)\cdot\{g_{M}(\hat{\theta}_{h(x),\mc{D}_n})\}^{2}+|m|\cdot h_{m}(x)\cdot\{g_{m}(\hat{\theta}_{h(x),\mc{D}_n})\}^{2}$.\\
Because of the toy setting, three constraints are held for arbitrary $\theta \in [-a, a]$ and $h(x) \in \mc{H}$.
\begin{enumerate}
    \item $|M|\cdot h_M(x)+|m|\cdot h_m(x)=1.$ (probability definition)
    \item $|M|\cdot h_M(x)\cdot g_{M}(\hat{\theta}_{h(x),\mc{D}_n})+|m|\cdot h_m(x)\cdot g_{m}(\hat{\theta}_{h(x),\mc{D}_n})=0.$\\Note that convex linear sum of the sample's gradient w.r.t. $h(x)=(h_M(x),h_m(x))$ is zero at the trained model $\hat{\theta}_{h(x),\mc{D}_n}$.
    \item $|g_{M}(\theta)|+|g_{m}(\theta)|=2a\Leftrightarrow g_{M}(\theta)-g_{m}(\theta)=2a\Leftrightarrow g_{M}(\theta)=2a+g_{m}(\theta).$\\Note that this is derived by the property of predefined loss function at $\theta\in[-a,a]$.
\end{enumerate}
2nd constraint is equivalent to $|M|\cdot h_{M}(x)\cdot (2a+g_m(\hat{\theta}_{h(x),\mc{D}_n}))+|m|\cdot h_{m}(x)\cdot g_m(\hat{\theta}_{h(x),\mc{D}_n})=0.\Leftrightarrow g_m(\hat{\theta}_{h(x),\mc{D}_n})=-2a|M|\cdot h_{M}(x)$. Because of the 3rd constraint, $g_{M}(\hat{\theta}_{h(x),\mc{D}_n})=2a(1-|M|\cdot h_{M}(x))$.
Then the objective is, maximizing 
\begin{equation}\label{appeq:objective_function}
\begin{split}
|M|\cdot h_{M}(x)\cdot \{2a(1-|M|\cdot h_{M}(x))\}^{2}&+(1-|M|\cdot h_{M}(x))\{2a|M|\cdot h_{M}(x)\}^{2}\\
&=4a^2|M|\cdot h_{M}(x)(1-|M|\cdot h_{M}(x))    
\end{split}
\end{equation}\\
\eqref{appeq:objective_function} is maximized when $|M|\cdot h_{M}(x)=\frac{1}{2}$, and it means $|m|\cdot h_{m}(x)=\frac{1}{2}$. Thus, $h^{\star}_{M}(x)=\frac{|m|}{2|M|\cdot |m|},h^{\star}_{m}(x)=\frac{|M|}{2|M|\cdot |m|}.$ This result is related with the trained model $\hat{\theta}_{U(x),\mc{D}_n}$, where $U_{M}(x)=U_{m}(x)=\frac{1}{|M|+|m|}$. At $\hat{\theta}_{U(x),\mc{D}_n}$, $|M|g_{M}(\hat{\theta}_{U(x),\mc{D}_n})+|m|g_{m}(\hat{\theta}_{U(x),\mc{D}_n})=0$ satisfies and this is equivalent to $|M|:|m|=|g_{m}(\hat{\theta}_{U(x),\mc{D}_n})|:|g_{M}(\hat{\theta}_{U(x),\mc{D}_n})|$.
\end{proof}
Thus, it is consistent with our intuition that setting the sampling probability $h$ for set M and m in proportion to $|g_{M}(\hat{\theta}_{U(x),\mc{D}_n})|$ and $|g_{m}(\hat{\theta}_{U(x),\mc{D}_n})|$ helps to minimize the trace of the inverse empirical Fisher information.

%% file: main.bbl
\begin{thebibliography}{75}
\providecommand{\natexlab}[1]{#1}
\providecommand{\url}[1]{\texttt{#1}}
\expandafter\ifx\csname urlstyle\endcsname\relax
  \providecommand{\doi}[1]{doi: #1}\else
  \providecommand{\doi}{doi: \begingroup \urlstyle{rm}\Url}\fi

\bibitem[Agarwal et~al.(2018)Agarwal, Beygelzimer, Dud{\'\i}k, Langford, and
  Wallach]{agarwal2018reductions}
Alekh Agarwal, Alina Beygelzimer, Miroslav Dud{\'\i}k, John Langford, and Hanna
  Wallach.
\newblock A reductions approach to fair classification.
\newblock In \emph{International Conference on Machine Learning}, pp.\  60--69.
  PMLR, 2018.

\bibitem[Ahmed et~al.(2020)Ahmed, Bengio, van Seijen, and
  Courville]{ahmed2020systematic}
Faruk Ahmed, Yoshua Bengio, Harm van Seijen, and Aaron Courville.
\newblock Systematic generalisation with group invariant predictions.
\newblock In \emph{International Conference on Learning Representations}, 2020.

\bibitem[Alvi et~al.(2018)Alvi, Zisserman, and Nell{\aa}ker]{alvi2018turning}
Mohsan Alvi, Andrew Zisserman, and Christoffer Nell{\aa}ker.
\newblock Turning a blind eye: Explicit removal of biases and variation from
  deep neural network embeddings.
\newblock In \emph{Proceedings of the European Conference on Computer Vision
  (ECCV)}, pp.\  0--0, 2018.

\bibitem[An et~al.(2020)An, Ying, and Zhu]{an2020resampling}
Jing An, Lexing Ying, and Yuhua Zhu.
\newblock Why resampling outperforms reweighting for correcting sampling bias
  with stochastic gradients.
\newblock In \emph{International Conference on Learning Representations}, 2020.

\bibitem[Arjovsky et~al.(2019)Arjovsky, Bottou, Gulrajani, and
  Lopez-Paz]{arjovsky2019invariant}
Martin Arjovsky, L{\'e}on Bottou, Ishaan Gulrajani, and David Lopez-Paz.
\newblock Invariant risk minimization.
\newblock \emph{arXiv preprint arXiv:1907.02893}, 2019.

\bibitem[Ash et~al.(2019)Ash, Zhang, Krishnamurthy, Langford, and
  Agarwal]{ash2019deep}
Jordan~T Ash, Chicheng Zhang, Akshay Krishnamurthy, John Langford, and Alekh
  Agarwal.
\newblock Deep batch active learning by diverse, uncertain gradient lower
  bounds.
\newblock \emph{arXiv preprint arXiv:1906.03671}, 2019.

\bibitem[Bahng et~al.(2020)Bahng, Chun, Yun, Choo, and Oh]{bahng2019learning}
Hyojin Bahng, Sanghyuk Chun, Sangdoo Yun, Jaegul Choo, and Seong~Joon Oh.
\newblock Learning de-biased representations with biased representations.
\newblock In \emph{International Conference on Machine Learning}, pp.\
  528--539. PMLR, 2020.

\bibitem[Bao et~al.(2021)Bao, Chang, and Barzilay]{bao2021predict}
Yujia Bao, Shiyu Chang, and Regina Barzilay.
\newblock Predict then interpolate: A simple algorithm to learn stable
  classifiers.
\newblock In \emph{International Conference on Machine Learning}, pp.\
  640--650. PMLR, 2021.

\bibitem[Borkan et~al.(2019)Borkan, Dixon, Sorensen, Thain, and
  Vasserman]{DBLP:journals/corr/abs-1903-04561}
Daniel Borkan, Lucas Dixon, Jeffrey Sorensen, Nithum Thain, and Lucy Vasserman.
\newblock Nuanced metrics for measuring unintended bias with real data for text
  classification.
\newblock \emph{CoRR}, abs/1903.04561, 2019.
\newblock URL \url{http://arxiv.org/abs/1903.04561}.

\bibitem[Brendel \& Bethge(2018)Brendel and Bethge]{brendel2018approximating}
Wieland Brendel and Matthias Bethge.
\newblock Approximating cnns with bag-of-local-features models works
  surprisingly well on imagenet.
\newblock In \emph{International Conference on Learning Representations}, 2018.

\bibitem[Cadene et~al.(2019)Cadene, Dancette, Cord, Parikh,
  et~al.]{cadene2019rubi}
Remi Cadene, Corentin Dancette, Matthieu Cord, Devi Parikh, et~al.
\newblock Rubi: Reducing unimodal biases for visual question answering.
\newblock In \emph{Advances in Neural Information Processing Systems}, pp.\
  839--850, 2019.

\bibitem[Chaudhari et~al.(2019)Chaudhari, Choromanska, Soatto, LeCun, Baldassi,
  Borgs, Chayes, Sagun, and Zecchina]{chaudhari2019entropy}
Pratik Chaudhari, Anna Choromanska, Stefano Soatto, Yann LeCun, Carlo Baldassi,
  Christian Borgs, Jennifer Chayes, Levent Sagun, and Riccardo Zecchina.
\newblock Entropy-sgd: Biasing gradient descent into wide valleys.
\newblock \emph{Journal of Statistical Mechanics: Theory and Experiment},
  2019\penalty0 (12):\penalty0 124018, 2019.

\bibitem[Chaudhuri et~al.(2015)Chaudhuri, Kakade, Netrapalli, and
  Sanghavi]{chaudhuri2015convergence}
Kamalika Chaudhuri, Sham Kakade, Praneeth Netrapalli, and Sujay Sanghavi.
\newblock Convergence rates of active learning for maximum likelihood
  estimation, 2015.

\bibitem[Clanuwat et~al.(2018)Clanuwat, Bober-Irizar, Kitamoto, Lamb, Yamamoto,
  and Ha]{clanuwat2018deep}
Tarin Clanuwat, Mikel Bober-Irizar, Asanobu Kitamoto, Alex Lamb, Kazuaki
  Yamamoto, and David Ha.
\newblock Deep learning for classical japanese literature, 2018.

\bibitem[Clark et~al.(2019)Clark, Yatskar, and Zettlemoyer]{clark2019don}
Christopher Clark, Mark Yatskar, and Luke Zettlemoyer.
\newblock Don’t take the easy way out: Ensemble based methods for avoiding
  known dataset biases.
\newblock In \emph{Proceedings of the 2019 Conference on Empirical Methods in
  Natural Language Processing and the 9th International Joint Conference on
  Natural Language Processing (EMNLP-IJCNLP)}, pp.\  4069--4082, 2019.

\bibitem[Cohen et~al.(2017)Cohen, Afshar, Tapson, and
  Van~Schaik]{cohen2017emnist}
Gregory Cohen, Saeed Afshar, Jonathan Tapson, and Andre Van~Schaik.
\newblock Emnist: Extending mnist to handwritten letters.
\newblock In \emph{2017 International Joint Conference on Neural Networks
  (IJCNN)}, pp.\  2921--2926. IEEE, 2017.

\bibitem[Creager et~al.(2021)Creager, Jacobsen, and
  Zemel]{creager2021environment}
Elliot Creager, J{\"o}rn-Henrik Jacobsen, and Richard Zemel.
\newblock Environment inference for invariant learning.
\newblock In \emph{International Conference on Machine Learning}, pp.\
  2189--2200. PMLR, 2021.

\bibitem[Darlow et~al.(2020)Darlow, Jastrz{\k{e}}bski, and
  Storkey]{darlow2020latent}
Luke Darlow, Stanis{\l}aw Jastrz{\k{e}}bski, and Amos Storkey.
\newblock Latent adversarial debiasing: Mitigating collider bias in deep neural
  networks.
\newblock \emph{arXiv preprint arXiv:2011.11486}, 2020.

\bibitem[Deng et~al.(2009)Deng, Dong, Socher, Li, Li, and
  Fei-Fei]{imagenet_cvpr09}
J.~Deng, W.~Dong, R.~Socher, L.-J. Li, K.~Li, and L.~Fei-Fei.
\newblock {ImageNet: A Large-Scale Hierarchical Image Database}.
\newblock In \emph{CVPR09}, 2009.

\bibitem[Geirhos et~al.(2018)Geirhos, Rubisch, Michaelis, Bethge, Wichmann, and
  Brendel]{geirhos2018imagenet}
Robert Geirhos, Patricia Rubisch, Claudio Michaelis, Matthias Bethge, Felix~A
  Wichmann, and Wieland Brendel.
\newblock Imagenet-trained cnns are biased towards texture; increasing shape
  bias improves accuracy and robustness.
\newblock In \emph{International Conference on Learning Representations}, 2018.

\bibitem[Goyal et~al.(2020)Goyal, Yang, Yang, and Deng]{goyal2020rel3d}
Ankit Goyal, Kaiyu Yang, Dawei Yang, and Jia Deng.
\newblock Rel3d: A minimally contrastive benchmark for grounding spatial
  relations in 3d.
\newblock \emph{Advances in Neural Information Processing Systems}, 33, 2020.

\bibitem[Goyal et~al.(2017)Goyal, Khot, Summers-Stay, Batra, and
  Parikh]{goyal2017making}
Yash Goyal, Tejas Khot, Douglas Summers-Stay, Dhruv Batra, and Devi Parikh.
\newblock Making the v in vqa matter: Elevating the role of image understanding
  in visual question answering.
\newblock In \emph{Proceedings of the IEEE Conference on Computer Vision and
  Pattern Recognition}, pp.\  6904--6913, 2017.

\bibitem[Hardt et~al.(2016)Hardt, Price, and Srebro]{hardt2016equality}
Moritz Hardt, Eric Price, and Nati Srebro.
\newblock Equality of opportunity in supervised learning.
\newblock \emph{Advances in neural information processing systems}, 29, 2016.

\bibitem[He et~al.(2016)He, Zhang, Ren, and Sun]{he2016deep}
Kaiming He, Xiangyu Zhang, Shaoqing Ren, and Jian Sun.
\newblock Deep residual learning for image recognition.
\newblock In \emph{Proceedings of the IEEE conference on computer vision and
  pattern recognition}, pp.\  770--778, 2016.

\bibitem[Hendrycks \& Dietterich(2019)Hendrycks and
  Dietterich]{hendrycks2019benchmarking}
Dan Hendrycks and Thomas Dietterich.
\newblock Benchmarking neural network robustness to common corruptions and
  perturbations.
\newblock \emph{arXiv preprint arXiv:1903.12261}, 2019.

\bibitem[Hsu et~al.(2020)Hsu, Shen, Jin, and Kira]{hsu2020generalized}
Yen-Chang Hsu, Yilin Shen, Hongxia Jin, and Zsolt Kira.
\newblock Generalized odin: Detecting out-of-distribution image without
  learning from out-of-distribution data.
\newblock In \emph{Proceedings of the IEEE/CVF Conference on Computer Vision
  and Pattern Recognition}, pp.\  10951--10960, 2020.

\bibitem[Huang et~al.(2021)Huang, Geng, and Li]{huang2021importance}
Rui Huang, Andrew Geng, and Yixuan Li.
\newblock On the importance of gradients for detecting distributional shifts in
  the wild.
\newblock \emph{Advances in Neural Information Processing Systems}, 34, 2021.

\bibitem[Idrissi et~al.(2022)Idrissi, Arjovsky, Pezeshki, and
  Lopez-Paz]{idrissi2022simple}
Badr~Youbi Idrissi, Martin Arjovsky, Mohammad Pezeshki, and David Lopez-Paz.
\newblock Simple data balancing achieves competitive worst-group-accuracy.
\newblock In \emph{Conference on Causal Learning and Reasoning}, pp.\
  336--351. PMLR, 2022.

\bibitem[Jastrz{\k{e}}bski et~al.(2017)Jastrz{\k{e}}bski, Kenton, Arpit,
  Ballas, Fischer, Bengio, and Storkey]{jastrzkebski2017three}
Stanis{\l}aw Jastrz{\k{e}}bski, Zachary Kenton, Devansh Arpit, Nicolas Ballas,
  Asja Fischer, Yoshua Bengio, and Amos Storkey.
\newblock Three factors influencing minima in sgd.
\newblock \emph{arXiv preprint arXiv:1711.04623}, 2017.

\bibitem[Karras et~al.(2019)Karras, Laine, and Aila]{karras2019style}
Tero Karras, Samuli Laine, and Timo Aila.
\newblock A style-based generator architecture for generative adversarial
  networks.
\newblock In \emph{Proceedings of the IEEE/CVF Conference on Computer Vision
  and Pattern Recognition}, pp.\  4401--4410, 2019.

\bibitem[Killamsetty et~al.(2020)Killamsetty, Sivasubramanian, Ramakrishnan,
  and Iyer]{killamsetty2020glister}
Krishnateja Killamsetty, Durga Sivasubramanian, Ganesh Ramakrishnan, and
  Rishabh Iyer.
\newblock Glister: Generalization based data subset selection for efficient and
  robust learning.
\newblock 2020.

\bibitem[Killamsetty et~al.(2021{\natexlab{a}})Killamsetty, Durga,
  Ramakrishnan, De, and Iyer]{killamsetty2021grad}
Krishnateja Killamsetty, S~Durga, Ganesh Ramakrishnan, Abir De, and Rishabh
  Iyer.
\newblock Grad-match: Gradient matching based data subset selection for
  efficient deep model training.
\newblock In \emph{International Conference on Machine Learning}, pp.\
  5464--5474. PMLR, 2021{\natexlab{a}}.

\bibitem[Killamsetty et~al.(2021{\natexlab{b}})Killamsetty, Zhao, Chen, and
  Iyer]{killamsetty2021retrieve}
Krishnateja Killamsetty, Xujiang Zhao, Feng Chen, and Rishabh Iyer.
\newblock Retrieve: Coreset selection for efficient and robust semi-supervised
  learning.
\newblock \emph{Advances in Neural Information Processing Systems}, 34,
  2021{\natexlab{b}}.

\bibitem[Kim et~al.(2019)Kim, Kim, Kim, Kim, and Kim]{kim2019learning}
Byungju Kim, Hyunwoo Kim, Kyungsu Kim, Sungjin Kim, and Junmo Kim.
\newblock Learning not to learn: Training deep neural networks with biased
  data.
\newblock In \emph{Proceedings of the IEEE Conference on Computer Vision and
  Pattern Recognition}, pp.\  9012--9020, 2019.

\bibitem[Kim et~al.(2021)Kim, Lee, and Choo]{kim2021biaswap}
Eungyeup Kim, Jihyeon Lee, and Jaegul Choo.
\newblock Biaswap: Removing dataset bias with bias-tailored swapping
  augmentation.
\newblock \emph{arXiv preprint arXiv:2108.10008}, 2021.

\bibitem[Kim et~al.(2022)Kim, Hwang, Ahn, Park, and Kwak]{kim2022learning}
Nayeong Kim, Sehyun Hwang, Sungsoo Ahn, Jaesik Park, and Suha Kwak.
\newblock Learning debiased classifier with biased committee.
\newblock \emph{arXiv preprint arXiv:2206.10843}, 2022.

\bibitem[Krishnakumar et~al.(2021)Krishnakumar, Prabhu, Sudhakar, and
  Hoffman]{krishnakumar2021udis}
Arvindkumar Krishnakumar, Viraj Prabhu, Sruthi Sudhakar, and Judy Hoffman.
\newblock Udis: Unsupervised discovery of bias in deep visual recognition
  models.
\newblock In \emph{British Machine Vision Conference (BMVC)}, volume~1, pp.\
  ~3, 2021.

\bibitem[Krizhevsky et~al.(2009)Krizhevsky, Hinton,
  et~al.]{krizhevsky2009learning}
Alex Krizhevsky, Geoffrey Hinton, et~al.
\newblock Learning multiple layers of features from tiny images.
\newblock 2009.

\bibitem[Lang et~al.(2021)Lang, Gandelsman, Yarom, Wald, Elidan, Hassidim,
  Freeman, Isola, Globerson, Irani, et~al.]{lang2021explaining}
Oran Lang, Yossi Gandelsman, Michal Yarom, Yoav Wald, Gal Elidan, Avinatan
  Hassidim, William~T Freeman, Phillip Isola, Amir Globerson, Michal Irani,
  et~al.
\newblock Explaining in style: Training a gan to explain a classifier in
  stylespace.
\newblock In \emph{Proceedings of the IEEE/CVF International Conference on
  Computer Vision}, pp.\  693--702, 2021.

\bibitem[Le~Bras et~al.(2020)Le~Bras, Swayamdipta, Bhagavatula, Zellers,
  Peters, Sabharwal, and Choi]{bras2020adversarial}
Ronan Le~Bras, Swabha Swayamdipta, Chandra Bhagavatula, Rowan Zellers, Matthew
  Peters, Ashish Sabharwal, and Yejin Choi.
\newblock Adversarial filters of dataset biases.
\newblock In \emph{International Conference on Machine Learning}, pp.\
  1078--1088. PMLR, 2020.

\bibitem[LeCun et~al.(2010)LeCun, Cortes, and Burges]{lecun2010mnist}
Yann LeCun, Corinna Cortes, and CJ~Burges.
\newblock Mnist handwritten digit database.
\newblock \emph{ATT Labs [Online]. Available:
  http://yann.lecun.com/exdb/mnist}, 2, 2010.

\bibitem[Lee et~al.(2021)Lee, Kim, Lee, Lee, and Choo]{lee2021learning}
Jungsoo Lee, Eungyeup Kim, Juyoung Lee, Jihyeon Lee, and Jaegul Choo.
\newblock Learning debiased representation via disentangled feature
  augmentation.
\newblock \emph{Advances in Neural Information Processing Systems}, 34, 2021.

\bibitem[Lee et~al.(2019)Lee, Lee, Shin, and Lee]{lee2019simple}
Kimin Lee, Kibok Lee, Jinwoo Shin, and Honglak Lee.
\newblock Network randomization: A simple technique for generalization in deep
  reinforcement learning.
\newblock In \emph{International Conference on Learning Representations}, 2019.

\bibitem[Lehmann \& Casella(1998)Lehmann and Casella]{LehmCase98}
Erich~L. Lehmann and George Casella.
\newblock \emph{Theory of Point Estimation}.
\newblock Springer-Verlag, New York, NY, USA, second edition, 1998.

\bibitem[Li \& Vasconcelos(2019)Li and Vasconcelos]{li2019repair}
Yi~Li and Nuno Vasconcelos.
\newblock Repair: Removing representation bias by dataset resampling.
\newblock In \emph{Proceedings of the IEEE Conference on Computer Vision and
  Pattern Recognition}, pp.\  9572--9581, 2019.

\bibitem[Li et~al.(2018)Li, Li, and Vasconcelos]{li2018resound}
Yingwei Li, Yi~Li, and Nuno Vasconcelos.
\newblock Resound: Towards action recognition without representation bias.
\newblock In \emph{Proceedings of the European Conference on Computer Vision
  (ECCV)}, pp.\  513--528, 2018.

\bibitem[Li \& Xu(2021)Li and Xu]{li2021discover}
Zhiheng Li and Chenliang Xu.
\newblock Discover the unknown biased attribute of an image classifier.
\newblock In \emph{Proceedings of the IEEE/CVF International Conference on
  Computer Vision}, pp.\  14970--14979, 2021.

\bibitem[Liu et~al.(2021)Liu, Haghgoo, Chen, Raghunathan, Koh, Sagawa, Liang,
  and Finn]{liu2021just}
Evan~Z Liu, Behzad Haghgoo, Annie~S Chen, Aditi Raghunathan, Pang~Wei Koh,
  Shiori Sagawa, Percy Liang, and Chelsea Finn.
\newblock Just train twice: Improving group robustness without training group
  information.
\newblock In \emph{International Conference on Machine Learning}, pp.\
  6781--6792. PMLR, 2021.

\bibitem[Liu et~al.(2015)Liu, Luo, Wang, and Tang]{DBLP:conf/iccv/LiuLWT15}
Ziwei Liu, Ping Luo, Xiaogang Wang, and Xiaoou Tang.
\newblock Deep learning face attributes in the wild.
\newblock In \emph{2015 {IEEE} International Conference on Computer Vision,
  {ICCV} 2015, Santiago, Chile, December 7-13, 2015}, pp.\  3730--3738. {IEEE}
  Computer Society, 2015.
\newblock \doi{10.1109/ICCV.2015.425}.
\newblock URL \url{https://doi.org/10.1109/ICCV.2015.425}.

\bibitem[McDuff et~al.(2019)McDuff, Ma, Song, and
  Kapoor]{mcduff2019characterizing}
Daniel McDuff, Shuang Ma, Yale Song, and Ashish Kapoor.
\newblock Characterizing bias in classifiers using generative models.
\newblock In \emph{Advances in Neural Information Processing Systems}, pp.\
  5404--5415, 2019.

\bibitem[McInnes et~al.(2018)McInnes, Healy, and Melville]{mcinnes2018uniform}
Leland McInnes, John Healy, and James Melville.
\newblock Umap: Uniform manifold approximation and projection for dimension
  reduction, 2018.
\newblock URL \url{http://arxiv.org/abs/1802.03426}.
\newblock cite arxiv:1802.03426Comment: Reference implementation available at
  http://github.com/lmcinnes/umap.

\bibitem[Mirzasoleiman et~al.(2020)Mirzasoleiman, Bilmes, and
  Leskovec]{mirzasoleiman2020coresets}
Baharan Mirzasoleiman, Jeff Bilmes, and Jure Leskovec.
\newblock Coresets for data-efficient training of machine learning models.
\newblock In \emph{International Conference on Machine Learning}, pp.\
  6950--6960. PMLR, 2020.

\bibitem[Nam et~al.(2020)Nam, Cha, Ahn, Lee, and Shin]{nam2020learning}
Jun~Hyun Nam, Hyuntak Cha, Sungsoo Ahn, Jaeho Lee, and Jinwoo Shin.
\newblock Learning from failure: De-biasing classifier from biased classifier.
\newblock In \emph{34th Conference on Neural Information Processing Systems
  (NeurIPS) 2020}. Neural Information Processing Systems, 2020.

\bibitem[Park et~al.(2020)Park, Zhu, Wang, Lu, Shechtman, Efros, and
  Zhang]{park2020swapping}
Taesung Park, Jun-Yan Zhu, Oliver Wang, Jingwan Lu, Eli Shechtman, Alexei~A
  Efros, and Richard Zhang.
\newblock Swapping autoencoder for deep image manipulation.
\newblock \emph{arXiv preprint arXiv:2007.00653}, 2020.

\bibitem[Pleiss et~al.(2017)Pleiss, Raghavan, Wu, Kleinberg, and
  Weinberger]{pleiss2017fairness}
Geoff Pleiss, Manish Raghavan, Felix Wu, Jon Kleinberg, and Kilian~Q
  Weinberger.
\newblock On fairness and calibration.
\newblock \emph{Advances in neural information processing systems}, 30, 2017.

\bibitem[Ramaswamy et~al.(2021)Ramaswamy, Kim, and
  Russakovsky]{ramaswamy2021fair}
Vikram~V Ramaswamy, Sunnie~SY Kim, and Olga Russakovsky.
\newblock Fair attribute classification through latent space de-biasing.
\newblock In \emph{Proceedings of the IEEE/CVF Conference on Computer Vision
  and Pattern Recognition}, pp.\  9301--9310, 2021.

\bibitem[Sagawa et~al.(2019)Sagawa, Koh, Hashimoto, and
  Liang]{sagawa2019distributionally}
Shiori Sagawa, Pang~Wei Koh, Tatsunori~B Hashimoto, and Percy Liang.
\newblock Distributionally robust neural networks for group shifts: On the
  importance of regularization for worst-case generalization.
\newblock \emph{arXiv preprint arXiv:1911.08731}, 2019.

\bibitem[Sch{\"a}fer(2016)]{schafer2016bias}
Roland Sch{\"a}fer.
\newblock On bias-free crawling and representative web corpora.
\newblock In \emph{Proceedings of the 10th web as corpus workshop}, pp.\
  99--105, 2016.

\bibitem[Seo et~al.(2022)Seo, Lee, and Han]{Seo_2022_CVPR}
Seonguk Seo, Joon-Young Lee, and Bohyung Han.
\newblock Unsupervised learning of debiased representations with
  pseudo-attributes.
\newblock In \emph{Proceedings of the IEEE/CVF Conference on Computer Vision
  and Pattern Recognition (CVPR)}, pp.\  16742--16751, June 2022.

\bibitem[Serfling(1980)]{Serf:appr:1980}
Robert~J. Serfling.
\newblock \emph{Approximation Theorems of Mathematical Statistics}.
\newblock John Wiley \& Sons, 1980.

\bibitem[Shah et~al.(2020)Shah, Tamuly, Raghunathan, Jain, and
  Netrapalli]{shah2020pitfalls}
Harshay Shah, Kaustav Tamuly, Aditi Raghunathan, Prateek Jain, and Praneeth
  Netrapalli.
\newblock The pitfalls of simplicity bias in neural networks.
\newblock \emph{Advances in Neural Information Processing Systems},
  33:\penalty0 9573--9585, 2020.

\bibitem[Shrestha et~al.(2021)Shrestha, Kafle, and
  Kanan]{shrestha2021investigation}
Robik Shrestha, Kushal Kafle, and Christopher Kanan.
\newblock An investigation of critical issues in bias mitigation techniques.
\newblock \emph{arXiv preprint arXiv:2104.00170}, 2021.

\bibitem[Singh et~al.(2020)Singh, Mahajan, Grauman, Lee, Feiszli, and
  Ghadiyaram]{singh2020don}
Krishna~Kumar Singh, Dhruv Mahajan, Kristen Grauman, Yong~Jae Lee, Matt
  Feiszli, and Deepti Ghadiyaram.
\newblock Don't judge an object by its context: Learning to overcome contextual
  bias.
\newblock In \emph{Proceedings of the IEEE/CVF Conference on Computer Vision
  and Pattern Recognition}, pp.\  11070--11078, 2020.

\bibitem[Sohoni et~al.(2020)Sohoni, Dunnmon, Angus, Gu, and
  R{\'e}]{sohoni2020no}
Nimit Sohoni, Jared Dunnmon, Geoffrey Angus, Albert Gu, and Christopher R{\'e}.
\newblock No subclass left behind: Fine-grained robustness in coarse-grained
  classification problems.
\newblock \emph{Advances in Neural Information Processing Systems},
  33:\penalty0 19339--19352, 2020.

\bibitem[Sourati et~al.(2016)Sourati, Akcakaya, Leen, Erdogmus, and
  Dy]{sourati2016asymptotic}
Jamshid Sourati, Murat Akcakaya, Todd~K. Leen, Deniz Erdogmus, and Jennifer~G.
  Dy.
\newblock Asymptotic analysis of objectives based on fisher information in
  active learning, 2016.

\bibitem[Tartaglione et~al.(2021)Tartaglione, Barbano, and
  Grangetto]{tartaglione2021end}
Enzo Tartaglione, Carlo~Alberto Barbano, and Marco Grangetto.
\newblock End: Entangling and disentangling deep representations for bias
  correction.
\newblock In \emph{Proceedings of the IEEE/CVF Conference on Computer Vision
  and Pattern Recognition}, pp.\  13508--13517, 2021.

\bibitem[Teney et~al.(2021)Teney, Abbasnejad, Lucey, and
  Hengel]{teney2021evading}
Damien Teney, Ehsan Abbasnejad, Simon Lucey, and Anton van~den Hengel.
\newblock Evading the simplicity bias: Training a diverse set of models
  discovers solutions with superior ood generalization.
\newblock \emph{arXiv preprint arXiv:2105.05612}, 2021.

\bibitem[Torralba \& Efros(2011)Torralba and Efros]{torralba2011unbiased}
Antonio Torralba and Alexei~A Efros.
\newblock Unbiased look at dataset bias.
\newblock In \emph{CVPR 2011}, pp.\  1521--1528. IEEE, 2011.

\bibitem[Wang et~al.(2018)Wang, He, Lipton, and Xing]{wang2019learning}
Haohan Wang, Zexue He, Zachary~C Lipton, and Eric~P Xing.
\newblock Learning robust representations by projecting superficial statistics
  out.
\newblock In \emph{International Conference on Learning Representations}, 2018.

\bibitem[Wasserman(2004)]{Wassermann04}
Larry Wasserman.
\newblock \emph{All of Statistics: A Concise Course in Statistical Inference}.
\newblock Springer Texts in Statistics. Springer, New York, 2004.
\newblock ISBN 978-1-4419-2322-6.
\newblock \doi{10.1007/978-0-387-21736-9}.

\bibitem[Woodworth et~al.(2017)Woodworth, Gunasekar, Ohannessian, and
  Srebro]{Woodworth2017LearningNP}
Blake~E. Woodworth, Suriya Gunasekar, Mesrob~I. Ohannessian, and Nathan Srebro.
\newblock Learning non-discriminatory predictors.
\newblock In \emph{COLT}, 2017.
\newblock URL
  \url{https://www.semanticscholar.org/paper/Learning-Non-Discriminatory-Predictors-Woodworth-Gunasekar/00cda3a1c7f716d6136f0d3c0c1fe1046e685e82}.

\bibitem[Xiao et~al.(2017)Xiao, Rasul, and Vollgraf]{xiao2017/online}
Han Xiao, Kashif Rasul, and Roland Vollgraf.
\newblock Fashion-mnist: a novel image dataset for benchmarking machine
  learning algorithms, 2017.

\bibitem[Zhang et~al.(2022{\natexlab{a}})Zhang, Lopez-Paz, and
  Bottou]{zhang2022rich}
Jianyu Zhang, David Lopez-Paz, and L{\'e}on Bottou.
\newblock Rich feature construction for the optimization-generalization
  dilemma.
\newblock \emph{arXiv preprint arXiv:2203.15516}, 2022{\natexlab{a}}.

\bibitem[Zhang et~al.(2022{\natexlab{b}})Zhang, Sohoni, Zhang, Finn, and
  R{\'e}]{zhang2022correct}
Michael Zhang, Nimit~S Sohoni, Hongyang~R Zhang, Chelsea Finn, and Christopher
  R{\'e}.
\newblock Correct-n-contrast: A contrastive approach for improving robustness
  to spurious correlations.
\newblock \emph{arXiv preprint arXiv:2203.01517}, 2022{\natexlab{b}}.

\bibitem[Zhang \& Sabuncu(2018)Zhang and Sabuncu]{zhang2018generalized}
Zhilu Zhang and Mert~R Sabuncu.
\newblock Generalized cross entropy loss for training deep neural networks with
  noisy labels.
\newblock In \emph{32nd Conference on Neural Information Processing Systems
  (NeurIPS)}, 2018.

\end{thebibliography}
